\documentclass[12pt]{article}
\usepackage{amsmath,amsthm,amssymb,bm}
\usepackage{graphicx,psfrag,epsf,algorithm,algpseudocode}
\usepackage{enumerate,multirow,array,color,natbib,url,booktabs,setspace,authblk}
\usepackage[font=small,labelfont=bf]{caption}
\usepackage[total={6.45in,8.75in}]{geometry}
\newcommand{\blind}{1}

\newtheorem{thm}{Theorem}

\newtheorem{lemma}{Lemma}

\newtheoremstyle{exampstyle}
{\topsep} 
{\topsep} 
{} 
{} 
{\bfseries} 
{.} 
{.5em} 
{} 
\graphicspath{{figs0/}}

\let\proglang=\textsf

\newcommand{\Mean}{{\mathbb{E}}}
\newcommand{\Var}{{\mbox{Var}}}
\newcommand{\Cov}{{\mbox{cov}}}

\newcommand{\diag}{{\mbox{diag}}}
\newcommand{\prob}{{\mbox{Pr}}}

\DeclareMathOperator*{\argmin}{arg\,min}
\newcommand\independent{\protect\mathpalette{\protect\independenT}{\perp}}
\def\independenT#1#2{\mathrel{\rlap{$#1#2$}\mkern2mu{#1#2}}}
\def\spacingset#1{\renewcommand{\baselinestretch}%
	{#1}\small\normalsize} \spacingset{1}

\usepackage{xr}
\begin{document}


\if1\blind
{
\title{\Large{\textbf{Dynamic
			Causal Effects Evaluation in A/B Testing with a Reinforcement Learning Framework}}}

		
\author[1]{Chengchun Shi}
\author[2]{Xiaoyu Wang}
\author[3]{Shikai Luo}
\author[4]{Hongtu Zhu}	
\author[5]{Jieping Ye}
\author[6]{Rui Song}
\affil[1]{London School of Economics and Political Science}
\affil[2]{Institute of Systems Science, Academy of Mathematics and Systems Science, Chinese Academy of Sciences}
\affil[3]{ByteDance}
\affil[4]{The Univeristy of North Carolina at Chapell Hill}
\affil[5]{University of Michigan}
\affil[6]{North Carolina State University}

\date{}
\maketitle
} \fi

\if0\blind
{
\title{\Large{\textbf{Dynamic
			Causal Effects Evaluation in A/B Testing with a Reinforcement Learning Framework}}}
\author{
}
\date{}
\maketitle
} \fi

\bigskip
\begin{abstract}
A/B testing, or online experiment is a standard  business strategy to compare a new product with an old one in pharmaceutical, technological, and traditional industries.  Major challenges arise in online experiments of two-sided marketplace platforms (e.g., Uber) 
where there is only one unit that receives a sequence of treatments over time. In those  experiments, the treatment at a given time   impacts  current outcome as well as future outcomes. The aim of this paper is to  introduce a reinforcement learning framework for carrying A/B testing in these experiments, while characterizing the long-term treatment effects. Our proposed testing procedure allows for sequential monitoring and online updating. It is generally applicable to a variety of treatment designs in different industries. In addition, we systematically investigate the theoretical properties (e.g., size and power) of our testing procedure. Finally, we apply our framework to both simulated data and a real-world data example obtained  from a technological company to illustrate its advantage over the current practice. A \proglang{Python} implementation of our test is available at \url{https://github.com/callmespring/CausalRL}.  
\end{abstract}

\vfill

\newpage
 
\baselineskip=20pt

\section{Introduction}
\label{secintroduction}
A/B testing, or online experiment is a business strategy to compare a   new product with an old one  in pharmaceutical, technological, and traditional industries (e.g., google, Amazon, or Facebook). It has became the gold standard to make data-driven decisions on a new service, feature, or product. For example, in web analytics, it is common to compare two variants of the same webpage (denote by A and B) by randomly splitting visitors into A and B and then contrasting metrics of interest (e.g., click-through rate) on each of the splits. There is a growing literature on developing A/B testing methods \citep[see e.g.,][and the references therein]{johari2015,kharitonov2015,johari2017,yang2017framework}. 
The key idea of these approaches is to apply causal inference methods to estimating the treatment effect of a new change under the assumption of the stable unit treatment value assumption \citep[SUTVA,][]{rubin1980randomization}. Please see e.g., \cite{Wager2018}, \cite{Imbens2015}, \cite{Yao2020}, \cite{Hernn2020} and the references therein. SUTVA precludes the existence of the \textit{interference} effect such that the response of each subject in the experiment depends only on their own treatment and is independent of others' treatments. Despite its ubiquitousness, however, the standard A/B testing is not directly applicable for causal inference under interference \citep{zhou2020cluster}. 

In this paper, we focus on the setting where there is only one unit (or system) in the experiment that receives a sequence of treatments over time. In many applications, the treatment at a given time can impact future outcomes, leading SUTVA being invalid. 
These studies frequently occur in the two-sided markets (intermediary economic platforms having two distinct user groups that provide each other with network benefits) that involve sequential decision making over time. As an illustration, we consider evaluating the effects of different order dispatching strategies in ride-sharing companies (e.g., Uber) for large-scale fleet management. See our real data analysis in Section \ref{sec:realdata} for details. These companies form a typical two-sided market that enables efficient interactions between passengers and drivers \citep{Rysman2009}. With the rapid development of smart mobile phones and internet of things, they have substantially transformed the transportation landscape of human beings \citep{Frenken2017,Jin2018,Hagiu2019}. Order dispatching is one of the most critical problems in online ride-sharing platforms to adapt the operation and management strategy to the dynamics in demand and supply. At a given time, an order dispatching strategy  not only affects the platform's immediate outcome (e.g., passengers' answer time, drivers' income), but also impacts the spatial distribution of drivers in the future. This in turn affects the platform's future outcome.  
The no interference assumption is thus violated. 

A fundamental question of interest that we consider here is how to develop valid A/B testing methods in the presence of interference.
Solving this fundamental question faces at least three major challenges. 
(i) The first one lies in establishing causal relationship between treatments and outcomes over time, by taking the carryover effect into consideration. Most of the existing A/B testing methods are ineffective. They fail to identify the carryover effect, leading the subsequent inference being invalid. See Section \ref{sec:toyexample} for details. (ii) The second one is that running each
experiment takes a considerable time. The company wishes to terminate the experiment as early as possible in order to save both time and budget. As such, the testing hypothesis needs to be sequentially evaluated online as the data are being collected, and the experiment shall be stopped in accordance with a pre-defined stopping rule as soon as significant results are observed.  (iii) The third one is that treatments are desired to be allocated in a manner to maximize the cumulative outcomes or to detect the alternative more efficiently. The testing procedure shall allow the treatment to be adaptively assigned. Addressing these challenges requires the development of new tools and theory for A/B testing and causal effects evaluation.

\subsection{Contributions} 
\begin{figure}[!t]
	\centering
	\includegraphics[width=14cm]{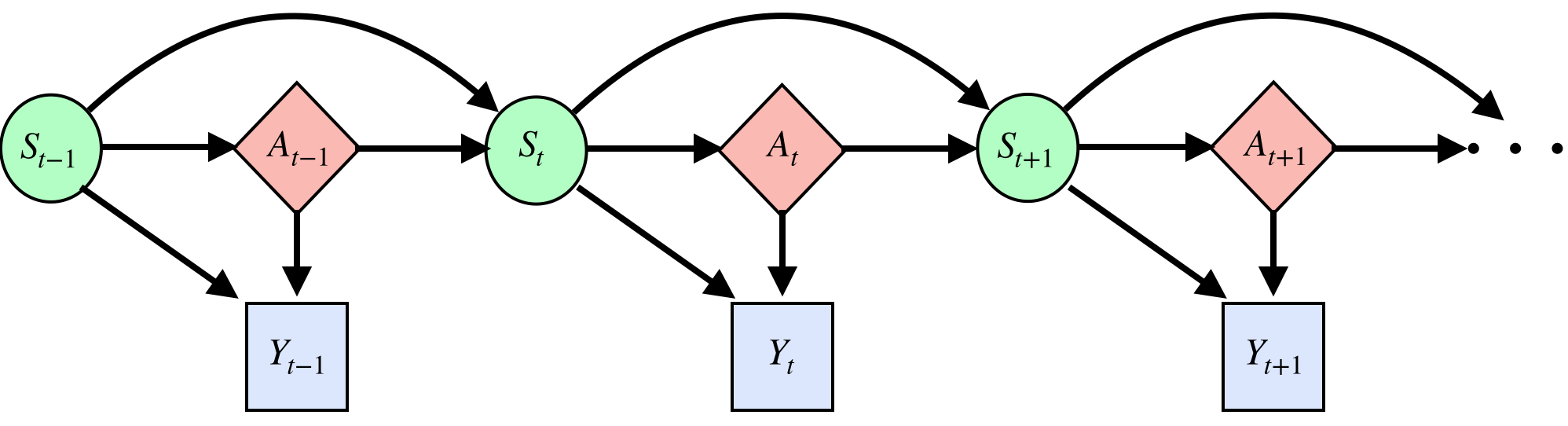}\vspace{-0.2cm}
	\caption{Causal diagram for MDP under settings where treatments depend on current states only. $(S_t,A_t,Y_t)$ represents the state-treatment-outcome triplet. Solid lines represent causal relationships. 
	}\label{fig0}
\end{figure}

We summarize our contributions as follows. First, to address the challenge mentioned in (i), we introduce a reinforcement learning \citep[RL, see e.g.,][for an overview]{Sutton2018} framework for A/B testing. RL is suitable framework to handle the carryover effects over time. In addition to the treatment-outcome pairs, it is assumed that  there is a set of time-varying state confounding variables. We model the state-treatment-outcome triplet by using the Markov decision process \citep[MDP, see e.g.][]{Puterman1994} to characterize the association between treatments and outcomes across time. Specifically, at each time point, the decision maker selects a treatment based on the observed state variables. The system responds by giving the decision maker a corresponding outcome and moving into a new state in the next time step. In this way, past treatments will have an indirect influence on future rewards through its effect on future state variables. See Figure \ref{fig0} for an illustration. In addition, the long-term treatment effects can be characterized by the  value functions (see Section \ref{secformprob} for details) that measure the discounted cumulative gain from a given initial state. Under this framework, it suffices to evaluate the difference between two value functions to compare different treatments. Our proposal gives an example of how to utilize some state-of-the-art machine learning tools, such as reinforcement learning, to address a challenging statistical inference problem for making business decisions. 

Second, to address the challenges mentioned in (ii) and (iii), we propose a novel {\color{black}sequential} testing procedure for detecting the difference between two value functions. 
Our proposed test integrates reinforcement learning and 
sequential analysis \citep[see e.g.][and the references therein]{jennison1999} 
to allows for sequential monitoring and online updating\footnote{\color{black}Our test statistic and its stopping boundary are updated as batches of new observations arrive without storing historical data.}. 
Meanwhile, our proposal contributes to each of these two areas as well. 
\begin{itemize}
	\item To the best of our knowledge, this is the first work on developing valid {\color{black}sequential} tests in the RL framework. Our work is built upon the temporal-difference learning method based on  function approximation \citep[see e.g.,][]{precup2001off,sutton2008}. In the computer science literature, convergence guarantees of temporal difference learning have been derived by \cite{sutton2008} under the setting of independent noise and by \cite{bhandari2018} for Markovian noise. However, uncertainty quantification and asymptotic distribution of the resulting value function estimators have been less studied. Such results are critical for carrying out  A/B testing. Recently, \cite{luckett2019} outlined a procedure for estimating the value under a given policy. \cite{shi2020statistical} developed a confidence interval for the value function.  However, these method do not allow for sequential monitoring or online updating. 
	
	\item Our proposal is built upon the $\alpha$-spending approach \citep{Lan1983} for sequential testing. We note that most test statistics in classical sequential analysis have the canonical joint distribution \citep[see Equation (3.1) in][]{jennison1999} and their associated stopping boundary can be recursively updated via numerical integration. However, in our setup, test statistics no longer have the  canonical joint distribution. This is due to the existence of the carryover effects in time. We discuss this in detail in Section \ref{secalphaspend}. As such, the numerical integration approach is not applicable to our setting. To resolve this issue, we propose a bootstrap-assisted procedure to determine the stopping boundary. It is much more computationally efficient than the classical wild bootstrap   algorithm \citep[see Section \ref{secalphaspend} for details]{wu1986jackknife}. The resulting test is generally applicable to a variety of treatment designs, including the Markov design, the alternating-time-interval design and the adaptive design (see Section \ref{asytest} for details).
	
\end{itemize}

Third, we systematically investigate the asymptotic properties of our testing procedure. We show that our test not only maintains the nominal type I error rate, but also has non-negligible powers against local alternatives. {\color{black}In particular, we show that when the sieve method is used for function approximation in temporal difference learning, undersmoothing is not needed to guarantee that the resulting value estimator has a tractable limiting distribution. This occurs because sieve estimators of conditional expectations are idempotent \citep{newey1998undersmoothing}. It implies that the proposed test will not be overly sensitive to the choice of the number of basis functions. To our knowledge, these results have not been established in the existing RL framework. Please see Section \ref{asytest} for details. 

Finally, our proposal addresses an important practical question in ride-sharing companies. In particular, the proposed methodology allows the company to evaluate different policies more accurately in the presence of the carryover effects. It also allows the company to terminate the online experiment earlier and to evaluate more policies within the same time frame. These policies have the potential to improve drivers' salary and meet more customer requests, providing a more efficient transportation network. Please see Section \ref{sec:realdata} for details.}


\subsection{Related work}
There is a huge literature on RL in the computer science community such that various algorithms are proposed for an agent to learn an optimal policy and interact with an environment. Recently, a few methods have been developed in the statistics literature on learning the optimal policy in mobile health applications \citep{ertefaie2014,luckett2019,hu2020personalized,liao2020batch}. In addition, there is a growing literature on adapting reinforcement learning to develop dynamic treatment regimes in precision medicine, to recommend treatment decisions based on individual patients' information \citep{Murphy2003,chak2010,qian2011performance,zhao2012,Zhang2013,song2015penalized,zhao2015,zhangyc2015,zhangyc2016,zhu2017greedy,wang2018quantile,shi2018high,shi2018maximin,mo2020learning,meng2020near}. 

Our work is closely related to the literature on off-policy evaluation, whose  objective is to estimate the value of a new policy based on data collected by a different policy. Existing literature can be cast into model-based methods, importance sampling (IS)-based and doubly-robust procedures. Model-based methods first fit an MDP model  from data and then compute the resulting value function. 
The estimated value function might suffer from a large bias due to potential misspecification of the model. Popular IS based methods include \cite{thomas2015,thomas2016data,liu2018}. 
These methods re-weight the observed rewards with the density ratio of the target and behavior policies. The value estimate might suffer from a large variance, due to the use of importance sampling. Doubly-robust methods \citep[see, e.g.,][]{jiang2016,kallus2019efficiently} learn the Q-function as well as the probability density ratio and combine these estimates properly for more robust and efficient value evaluation. However, both IS and doubly-robust methods required the treatment assignment probability (propensity score) to be bounded away from 0 and 1. As such, they are inapplicable to the alternating-time-interval design, which is the treatment allocation strategy in our real data application {\color{black}(see Section \ref{sec:realdata} for details)}. 

 
{\color{black}In addition to the literature on RL, our work is also related to a line of research on causal inference with interference. Most of the works studied the interference effect across different subjects \citep[see e.g.,][]{hudgens2008toward,pouget2019testing,li2019randomization,zhou2020cluster,reich2020review}. That is, the outcome for one subject depends on the treatment assigned to other subjects as well. To the contrary, our work focuses on the interference effect over time. We also remark that most of the aforementioned methods were primarily motivated by research questions in psychological, environmental and epidemiological studies, so their generalization to infer time dependent causal effects in two-sided markets remains unknown.}

Finally, we remark that there is a growing literature on evaluating time-varying causal effects \citep[see e.g.][]{robins1986,sobel2014causal,Murphy2018,Ning2019,rambachan2019,viviano2019synthetic,bojinov2020}. However, none of the above cited works used a RL framework to characterize the treatment effects. In particular, \cite{bojinov2020} proposed to use IS based methods to test the null hypothesis of no (average) temporal causal effects in time series experiments. Their causal estimand is different from ours since they focused on $p$ lag treatment effects,  whereas we consider the long-term effects characterized by the value function. Moreover, their method requires the propensity score to be bounded away from 0 and 1, and  thus it is not valid for our applications. In addition, these method do not allow for sequential monitoring. 

\subsection{Organization of the paper}
{\color{black}The rest of the paper is organised as follows. In Section \ref{secpotential}, we introduce a potential outcome framework to MDP and describe the causal estimand. Our testing procedure is introduced in Section \ref{sectest}. In Section \ref{secnumerical}, we demonstrate the effectiveness of our test via simulations. In Section \ref{sec:realdata}, we apply the proposed test to a data from an online ride-hailing platform to illustrate its usefulness. Finally, we conclude our paper in Section \ref{sec:dis}. }
\section{Problem formulation}\label{secpotential}
\subsection{A potential outcome framework for  MDP}\label{secformprob}
For simplicity, we assume that there are only two treatments (actions, products), coded as 0 and 1, respectively. For any $t\ge 0$, let $\bar{a}_t=(a_0,a_1,\cdots,a_t)^\top\in \{0,1\}^{t+1}$ denote a treatment history vector up to time $t$. Let $\mathbb{S}$ denote the support of state variables and $S_0$ denote the initial state variable. 
We assume $\mathbb{S}$ is a compact subset of $\mathbb{R}^d$. 
For any $(\bar{a}_{t-1},\bar{a}_{t})$, let $S_{t}^*(\bar{a}_{t-1})$ and $Y_t^*(\bar{a}_t)$ be the counterfactual state and counterfactual outcome, respectively,  that would occur at time $t$ had the agent followed the treatment history $\bar{a}_{t}$. 
The set of potential outcomes up to time $t$ is given by
\begin{eqnarray*}
	W_t^*(\bar{a}_t)=\{S_0,Y_0^*(a_0),S_1^*(a_0),\cdots,S_{t}^*(\bar{a}_{t-1}),Y_t^*(\bar{a}_t)\}.
\end{eqnarray*}
Let $W^*=\cup_{t\ge 0,\bar{a}_t\in \{0,1\}^{t+1}} W_t^*(\bar{a}_t)$ be the set of all potential outcomes.

A deterministic policy $\pi$ is a time-homogeneous function that maps the space of state variables to the set of available actions. 
Following $\pi$, the agent will assign actions according to $\pi$ at each time.  
We use $S_t^*(\pi)$ and $Y_t^*(\pi)$ to denote the associated potential state and outcome that would occur at time $t$ had the agent followed $\pi$. 
The goodness of  a policy $\pi$ is measured by its (state) value function, 
\begin{eqnarray*}
	V(\pi;s)=\sum_{t\ge 0} \gamma^t \Mean \{Y_t^*(\pi)|S_0=s\},
\end{eqnarray*}
where $0<\gamma<1$ is a discount factor that reflects the trade-off between immediate and future outcomes. The value function measures the discounted cumulative outcome that the agent would receive had they followed $\pi$. Note that our definition of the value function is slightly different from those in the existing literature \citep[see][for example]{Sutton2018}. Specifically, $V(\pi;s)$ is defined through potential outcomes rather than the observed data. 

Similarly, we define the Q function by
\begin{eqnarray*}
	Q(\pi;a,s)=\sum_{t\ge 0} \gamma^t \Mean \{Y_t^*(\pi(a))|S_0=s\}, 
\end{eqnarray*}
where $\pi(a)$ denotes a time-varying policy where the initial action equals to $a$ and all other actions are assigned according to $\pi$. 

{\color{black}The goal of A/B testing is to compare the difference between the two treatments. 
Toward that end, we focus on two nondynamic (state-agnostic) policies that assign the same treatment at each time point. We remark that this is non-traditional in RL where the goal is to build a policy that depends on the state. In Section \ref{sec:dynamic}, we discuss the extension to testing two dynamic policies.} For these two nondynamic policies, we use their value functions (denote by $V(1;\cdot)$ and $V(0;\cdot)$) to measure their long-term treatment effects. Meanwhile, our proposed method is equally applicable to the dynamic policy scenario as well. See Section \ref{sec:dis} for details. To quantitatively  compare the two policies, we introduce the Conditional Average Treatment Effect (CATE) and Average Treatment Effect (ATE) based on their value functions in the following definitions. These two definitions relate RL to causal inference. 

\medskip

\noindent \textbf{Definition 1.} Conditional on the initial state $S_0=s$, CATE is defined by the difference between two value functions,  i.e.,
$\hbox{CATE}(s)=V(1;s)-V(0;s)$.

\medskip

\noindent \textbf{Definition 2.} For a given reference distribution function $\mathbb{G}$ that has a bounded density function on $\mathbb{S}$, ATE is defined by the integrated difference between two value function, i.e., $\hbox{ATE}=\int_{s} \{V(1;s)-V(0;s)\}\mathbb{G}(ds). $


\medskip

The focus of this paper is to test the following hypotheses:  
\begin{eqnarray*}
	H_0: \tau_0=\hbox{ATE}\le 0\,\,\,\,\hbox{v.s}\,\,\,\,H_1: \tau_0=\hbox{ATE}>0.
\end{eqnarray*}
{\color{black}When $H_0$ holds, the new product is no better than the old one on average and is not of practical interest.}
\subsection{Identifiability of ATE}
One of the most important question in causal inference is the identifiability of causal effects. In this section, we present sufficient conditions that guarantee the identifiability of the value function.

We first introduce two conditions that are commonly assumed in multi-stage decision making problems \citep[see e.g.][]{Murphy2003,robins2004,Zhang2013}.  
We  need to use the notation $Z_1\independent Z_2|Z_3$ to indicate that $Z_1$ and $Z_2$ are independent conditional on $Z_3$. 
In practice, with the exception of $S_0$, the set $W^*$ cannot be observed, whereas at time $t$, we observe the state-action-outcome triplet $(S_t,A_t,Y_t)$. 
For any $t\ge 0$, let $\bar{A}_t=(A_0,A_1,\cdots,A_t)^\top$ denote the observed treatment history.

\medskip

\noindent (CA) Consistency assumption: $S_{t+1}=S_{t+1}^*(\bar{A}_{t})$ and $Y_t=Y_t^*(\bar{A}_t)$ for all $t\ge 0$, almost surely.

\noindent (SRA) Sequential randomization assumption: $A_t\independent W^*| S_{t}, \{S_j,A_j,Y_j\}_{0\le j<t}$.

\medskip

{\color{black}The CA requires that the observed state and outcome correspond to the potential state and outcome whose treatments are assigned according to the observed treatment history. 
It generalizes SUTVA to our setting, allowing the potential outcomes to depend on past treatments.} The SRA implies that  there are no unmeasured confounders and it automatically holds in online randomized experiments, in which the treatment assignment mechanism is pre-specified. In SRA, we  allows $A_t$ to depend on the observed data history $S_{t}, \{S_j,A_j,Y_j\}_{0\le j<t}$ and thus, the treatments can   be adaptively chosen.  


{\color{black}We next introduce two conditions that are unique to the reinforcement learning setting.}

\medskip

\noindent (MA)	Markov assumption:  there exists a Markov transition kernel $\mathcal{P}$ such that  for any $t\ge 0$, $\bar{a}_{t}\in \{0,1\}^{t+1}$ and $\mathcal{S}\subseteq \mathbb{R}^d$, we have 
$\prob\{S_{t+1}^*(\bar{a}_{t})\in \mathcal{S}|W_t^*(\bar{a}_t)\}=\mathcal{P}(\mathcal{S};a_t,S_t^*(\bar{a}_{t-1})).$

\noindent (CMIA)	Conditional mean independence assumption: there exists a function $r$ such that  for any $t\ge 0, \bar{a}_{t}\in \{0,1\}^{t+1}$, we have 
$\Mean \{Y_t^*(\bar{a}_t)|S_t^*(\bar{a}_{t-1}),W_{t-1}^*(\bar{a}_{t-1})\}=r(a_t,S_t^*(\bar{a}_{t-1}))$.

\medskip

We make a few remarks. First, these two conditions are central to the empirical validity of reinforcement learning (RL). Specifically, under these two conditions, one can show that there exists an optimal time-homogenous stationary policy whose value is no worse than any history-dependent policy \citep{Puterman1994}. This observation forms the foundation of most of the existing state-of-the-art RL algorithms. 

Second, when CA and SRA hold, it implies that the Markov assumption and the conditional mean independence assumption hold on the observed data as well,
\begin{eqnarray}\label{eqn:Markovobserve}
	\prob(S_{t+1}\in \mathcal{S}|A_t,S_t,\{S_j,A_j,Y_j\}_{0\le j<t})&=&\mathcal{P}(\mathcal{S};A_t,S_t),\\\label{eqn:robserve}
	\Mean(Y_t|A_t,S_t,\{S_j,A_j,Y_j\}_{0\le j<t})&=&r(A_t,S_t).
\end{eqnarray}
As such, $\mathcal{P}$ corresponds to the transition function that defines the next state distribution conditional on the current state-action pair and $r$ corresponds to the conditional expectation of the immediate reward as a function of the state-action pair. 

Assumption \eqref{eqn:Markovobserve} is commonly assumed in the existing reinforcement learning literature \citep[see e.g.,][]{ertefaie2014,luckett2019}. It is testable based on the observed data. See the goodness-of-fit test developed by \cite{shi2020does}. In practice, to ensure the Markov property is satisfied, we can construct the state by concatenating measurements over multiple decision points till the Markovian property is satisfied. 

Assumption \eqref{eqn:robserve} implies that past treatments will affect future response only through its impact on the future state variables. In other words, the state variables shall be chosen to include those that serve as important mediators between past treatments and current outcomes. By Assumption \eqref{eqn:Markovobserve}, this assumption is automatically satisfied when $Y_t$ is a deterministic function of $(S_t,A_t,S_{t+1})$ that measures the system's status at time $t+1$. The latter condition is commonly imposed in the reinforcement learning literature and is stronger than \eqref{eqn:robserve}.  




To conclude this section, we derive {\color{black}a version of Bellman equation for the Q function under the potential outcome framework. Specifically, for $a',a\in \{0,1\}$, let $Q(a';a,\cdot)$ denote the Q function where treatment $a$ is assigned at the initial decision point and treatment $a'$ is repeatedly assigned afterwards.} By definition, we have $V(a;s)=Q(a;a,s)$ for any $(a,s)$.


\begin{lemma}\label{lemma2}
	Under MA, CMIA, CA and SRA, for any $t\ge 0$, $a'\in \{0,1\}$ and any function $\varphi: \mathbb{S}\times \{0,1\}\to \mathbb{R}$, we have
	$\Mean [\{Q(a';A_t,S_t)-Y_t -\gamma Q(a';a',S_{t+1})\}\varphi(S_t,A_t)]=0$.
\end{lemma}
\vspace{-0.1cm}

Lemma \ref{lemma2} 
implies that the Q-function is estimable from the observed data. 
Specifically, an estimating equation can be constructed based on Lemma \ref{lemma2} and the Q-function can be learned by solving this estimating equation. Note that $V(a,s)=Q(a;a,s)$ and $\tau_0$ is completely determined by the value function $V$. As a result, $\tau_0=$ATE is identifiable. 

Note that the positivity assumption is not needed in Lemma \ref{lemma2}. Our procedure can thus handle the case where treatments are deterministically assigned.  This is due to MA and CMIA that assume the system dynamics are invariant across time. {\color{black}To elaborate this, note that the discounted value function is completely determined by the transition kernel $\mathcal{P}$ and the reward function $r$. These quantities can be consistently estimated under certain conditions, regardless of whether the treatments are deterministically assigned or not. Consequently, the value can be consistently estimated even when the treatment assignments are deterministic.}
We formally introduce our testing procedure in the next section.  



\section{Testing procedure}\label{sectest}
We first introduce a toy example to illustrate the limitations of existing A/B testing methods. We next present our method and prove its consistency under a variety of different treatment designs.
\subsection{Toy examples}\label{sec:toyexample}
Existing A/B testing methods can only detect short-term treatment effects, but fail to identify any long-term effects. To elaborate this, we introduce two examples below. 

\medskip

\noindent \textbf{Example 1.} $S_t=0.5\varepsilon_t$, $Y_t=S_t+\delta A_t$ for any $t\ge 1$ and $S_0=0.5\varepsilon_0$.

\noindent \textbf{Example 2.} $S_t=0.5S_{t-1}+\delta A_{t-1}+0.5\varepsilon_t$, $Y_t=S_t$ for any $t\ge 1$ and $S_0=0.5\varepsilon_0$.

\medskip

In both examples, the random errors $\{\varepsilon_t\}_{t\ge 0}$ follow independent standard normal distributions and the parameter $\delta$ describes the degree of treatment effects. When $\delta=0$, $H_0$ holds. 
Suppose $\delta>0$. Then $H_1$ holds. In Example 1, the observations are independent and there are no carryover effects at all. In this case, both the existing A/B tests and the proposed test are able to discriminate $H_1$ from $H_0$. In Example 2, however, treatments have delayed effects on the outcomes. Specifically, $Y_t$ does not depend on $A_t$, but is affected by $A_{t-1}$ through $S_t$. Existing tests will fail to detect $H_1$ as the short-term conditional average treatment effects $\Mean (Y_t|A_t=1,S_t)-\Mean (Y_t|A_t=0,S_t)=0$ in this example. As an illustration, we conduct a small experiment by assuming the decision is made once at $T=500$, and report the empirical rejection probability of the classical two-sample t-test that is commonly used in online experiments, a more complicated  test based on the double machine learning method \citep[DML,][]{chernozhukov2017double} that is widely employed for inferring causal effects, and the proposed test. It can be seen the competing methods do not have any power under Example 2. 

\begin{table}\label{tabletoy}
	\centering
	\small
	\begin{tabular}{c|c|c|c|c|c}\toprule
		\multicolumn{3}{c|}{Example 1} &  \multicolumn{3}{c}{Example 2} \\ \midrule
		t-test 0.76 & DML-based test 1 & our test 0.98 & t-test 0.04 & DML-based test 0.06 & our test 0.73 \\ \bottomrule
	\end{tabular}
	\caption{\small Powers of t-test, DML-based test and the proposed test under Examples 1 and 2, with $T=500$, $\delta=0.1$. $\{A_t\}_t$ follow i.i.d. Bernoulli distribution with success probability 0.5.}
\end{table}
\subsection{An overview of the proposal}
We present an overview of our proposal in this section. As commented before, we adopt a reinforcement learning framework to address the limitations of existing A/B testing methods and characterize the long-term treatment effects.  
First, we estimate $\tau_0$ based on a version of temporal difference learning. The idea is to apply basis function approximations to solve an estimating equation derived from Lemma \ref{lemma2}. Specifically, {\color{black}let $\mathcal{Q}=\{\Psi^\top(s) \beta_{a}:\beta_a\in \mathbb{R}^q\}$ be a large linear approximation space for $Q(a;a,s)=V(a,s)$, where $\Psi(\cdot)$ is a vector containing $q$ basis functions on $\mathbb{S}$. The dimension $q$ is allowed to grow with the number of samples $T$ to alleviate the effects of model misspecification. 
Let us suppose $Q\in \mathcal{Q}$ for a moment. Set the function $\psi(s,a)$ in Lemma \ref{lemma2} to $\Psi(s) \mathbb{I}(a=a')$ for $a'=0,1$, there exists some $\bm{\beta}^*=(\beta_0^{*\top},\beta_1^{*\top})^\top$ such that
\begin{eqnarray*}
	\Mean [\{\Psi^\top(S_t)\beta_{a}^*-Y_t -\gamma \Psi^\top(S_{t+1})\beta_{a}^*\}\Psi(S_t) \mathbb{I}(A_t=a)]=0,\,\,\,\,\forall a\in \{0,1\}, 
\end{eqnarray*}
where $\mathbb{I}(\cdot)$ denotes the indicator function. 
The above equations can be rewritten as $\Mean (\bm{\Sigma}_t \bm{\beta}^*)=\Mean \bm{\eta}_t$,  where $\bm{\Sigma}_t$ is a block diagonal matrix given by 
\begin{eqnarray*}
	\bm{\Sigma}_t=\left[\begin{array}{cc}
		\Psi(S_t)\mathbb{I}(A_t=0)\{\Psi(S_t)-\gamma \Psi(S_{t+1})\}^\top & \\ 
	& \Psi(S_t)\mathbb{I}(A_t=1)\{\Psi(S_t)-\gamma \Psi(S_{t+1})\}^\top
	\end{array}\right]\\ \textrm{and}\,\,\,\, \bm{\eta}_t=
\{\Psi(S_t)^\top \mathbb{I}(A_t=0) Y_t, \Psi(S_t)^\top \mathbb{I}(A_t=1) Y_t\}^\top.
\end{eqnarray*}
Let $\widehat{\bm{\Sigma}}(t)=t^{-1}\sum_{j<t} \bm{\Sigma}_j$ and   $\widehat{\bm{\eta}}(t)=t^{-1}\sum_{j<t} \bm{\eta}_j$. It follows that $\Mean \{\widehat{\bm{\Sigma}}(t)\beta^*\}=\Mean \{\widehat{\bm{\eta}}(t)\}$. This motivates us to estimate $\bm{\beta}^*$ by $$\widehat{\bm{\beta}}(t)=\{\widehat{\beta}_{0}^{\top}(t),\widehat{\beta}_{1}^{\top}(t)\}^\top=\widehat{\bm{\Sigma}}^{-1}(t)\widehat{\bm{\eta}}(t).$$ 
ATE can thus be estimated by the plug-in estimator $\widehat{\tau}(t)=\int_{s} \Psi^\top(s) \{\widehat{\beta}_{1}(t)- \widehat{\beta}_{0}(t) \}\mathbb{G}(ds)$. {\color{black}We remark that there is no guarantee that $\widehat{\bm{\Sigma}}(t)$ is always invertible. However, its population limit, $\bm{\Sigma}(t)$ is invertible for any $t$ (see Lemma \ref{lemmaA1} in the supplementary article). Consequently, for sufficiently large $t$, $\widehat{\bm{\Sigma}}(t)$ is invertible with large probability. In cases where $\widehat{\bm{\Sigma}}(t)$ is not invertible, we may add a ridge penalty to compute the resulting estimator. See Appendix D.4 of \cite{shi2020statistical} for details.}

Second, we use $ \widehat{\tau}(t)$ to construct  our test statistic at time $t$. Let 
\begin{eqnarray}\label{eqn:U}
\bm{U}=\left\{-\int_{s\in \mathbb{S}} \Psi(s)^\top \mathbb{G}(ds), \int_{s\in \mathbb{S}} \Psi(s)^\top \mathbb{G}(ds)\right\}^\top.
\end{eqnarray}
It follows that $\widehat{\tau}(t)=\bm{U}\widehat{\bm{\beta}}(t)$. We will show that $\sqrt{t}\{\widehat{\bm{\beta}}(t)-\bm{\beta}^*\}$ is multivariate normal. This implies that $\sqrt{t}\{\widehat{\tau}(t) -\tau_0\}$ is asymptotically normal. 
Its variance can be consistently estimated by 
$$\widehat{\sigma}^2(t)=\bm{U}^\top \widehat{\bm{\Sigma}}^{-1}(t) \widehat{\bm{\Omega}}(t )\{\widehat{\bm{\Sigma}}^{-1}(t)\}^\top \bm{U},$$ 
as $t$ grows to infinity, where $\widehat{\bm{\Sigma}}^{-1}(t) \widehat{\bm{\Omega}}(t) \{\widehat{\bm{\Sigma}}^{-1}(t)\}^\top$ 
is the sandwich estimator for the variance of $\sqrt{t}\{\widehat{\bm{\beta}}(t)-\bm{\beta}^*\}$, and that
\begin{eqnarray*}
\widehat{\bm{\Omega}}(t)=\frac{1}{t} \sum_{j=0}^{t-1} \left\{\begin{array}{ll}
\Psi(S_j)(1-A_j)\widehat{\varepsilon}_{j,0}\\
\Psi(S_j)A_j\widehat{\varepsilon}_{j,1}
\end{array}\right\}
 \left\{\begin{array}{ll}
\Psi(S_j)(1-A_j)\widehat{\varepsilon}_{j,0}\\
\Psi(S_j)A_j\widehat{\varepsilon}_{j,1}
\end{array}\right\}^\top,
\end{eqnarray*}
where $\widehat{\varepsilon}_{j,a}$ is the temporal difference error $Y_j+\gamma \Psi^\top(S_{j+1})\widehat{\beta}_{a}-\Psi^\top(S_j)\widehat{\beta}_{a}$ whose conditional expectation given $(A_j=a,S_j)$ is zero asymptotically (see Lemma \ref{lemma2}).} 
This yields our test statistic $\sqrt{t}\widehat{\tau}(t)/\widehat{\sigma}(t)$, at time $t$. 
For a given significance level $\alpha>0$, we reject $H_0$ when $\sqrt{t} \widehat{\tau}(t)/\widehat{\sigma}(t)>z_{\alpha}$,  where $z_{\alpha}$ is the upper $\alpha$-th quantile of a standard normal distribution.


Third, we integrate the $\alpha$-spending approach with bootstrap to sequentially implement our test (see Section \ref{secalphaspend}). The idea is to generate bootstrap samples that mimic the distribution of our test statistics, to specify the stopping boundary at each interim stage. 
Suppose that the interim analyses are conducted at time points $T_1<\cdots<T_K=T$. {\color{black}We focus on the setting where both $K$ and $\{T_k\}_k$ are pre-determined, as in our application (see Section \ref{sec:realdata} for details).}
To simplify the presentation, for each $1\le k<K$, we assume $T_k/T\to c_k $ for some constants $0<c_1<c_2<\cdots<c_{K-1}<1$. To better understand our algorithm, we investigate the limiting distribution of our test statistics at these interim stages in the next section. 

{\color{black}Finally, we remark that in the current setup, we assume the dimension of the state is fixed whereas the number of basis functions diverges to infinity at a rate that is slower than $T$. In Appendix \ref{secexthighd}, we extend our proposal to settings with high-dimensional state information. In that case, we recommend to include a rich class of basis functions to ensure that the Q-function can be well-approximated. The number of basis functions is allowed to be much larger than $T$. To handle high-dimensionality, we first adopt the Dantzig selector \citep{candes2007dantzig} which directly penalizes the Bellman equation to compute an initial estimator. We next develop a decorrelated estimator to reduce the bias of the initial estimator and outline the corresponding testing statistic. We also remark that for simplicity, we use the same Q-function model at each interim stage. This works when $\{T_k\}_k$ are of the same order of magnitude, which is the case in our real data application where $T_1=T_K/2$. Alternatively, one could allow $q$ to grow with $k$. The testing procedure can be similarly derived.}

\subsection{Asymptotic properties under different treatment designs}\label{asytest}
We consider three treatment allocation designs that can be handled by our procedure as follows:

\noindent \textbf{D1}. Markov design: $\prob(A_t=1|S_t, \{S_j,A_j,Y_j\}_{0\le j<t})=b^{(0)}(S_t)$ for some function $b^{(0)}(\cdot)$ uniformly bounded away from $0$ and $1$. 

\noindent \textbf{D2}. Alternating-time-interval design: $A_{2j}=0$,  $A_{2j+1}=1$ for all $j\ge 0$.

\noindent \textbf{D3}. Adaptive design: For $T_{k}\le t<T_{k+1}$ for some $k\ge 0$, $\prob(A_t=1|S_t, \{S_j,A_j,Y_j\}_{0\le j<t})=b^{(k)}(S_t)$ for some $b^{(k)}(\cdot)$ that depends on $\{S_j,A_j,Y_j\}_{0\le j<T_k}$ and is uniformly bounded away from $0$ and $1$ almost surely. We set $T_0=0$. 

Here, D2 is a deterministic design and is widely used in industry (see our real data example and this technical report\footnote{\url{https://eng.lyft.com/experimentation-in-a-ridesharing-marketplace-b39db027a66e}}). D1 and D3 are random designs. D1 is commonly assumed in the literature on reinforcement learning \citep{Sutton2018}. D3 is widely employed in the contextual bandit setting to balance the trade-off between exploration and exploitation. These three settings cover a variety of scenarios in practice. 

In D3, we require $b^{(k)}$ to be strictly bounded between 0 and 1. Suppose an $\epsilon$-greedy policy is used, i.e. $b^{(k)}(s)=\epsilon/2+(1-\epsilon) \widehat{\pi}^{(k)}(s)$,  where $\widehat{\pi}^{(k)}$ denotes some estimated optimal policy. It follows that $\epsilon/2 \le b^{(k)}(s)\le 1-\epsilon/2$ for any $s$. Such a requirement is automatically satisfied. Meanwhile, other adaptive strategies are equally applicable \citep[see e.g.,][]{zhang2007asymptotic,hu2015unified,metelkina2017information}. 

For any behaviour policy $b$ in D1-D3, define $S_t^*(\bar{b}_{t-1})$ and $Y_t^*(\bar{b}_t)$ as the potential outcomes at time $t$, where $\bar{b}_t$ denotes the action history assigned according to $b$. When $b$ is a random policy as in D1 or D3, definitions of these potential outcomes are more complicated than those under a deterministic policy 
(see Appendix \ref{secporandom} for details). 
When $b$ is a stationary policy, it follows from MA that $\{S_{t+1}^*(\bar{b}_{t})\}_{t\ge -1}$ forms a time-homogeneous Markov chain. When $b$ follows the alternating-time-interval design, both  $\{S_{2t}^*(\bar{b}_{2t-1})\}_{t\ge 0}$ and $\{S_{2t+1}^*(\bar{b}_{2t})\}_{t\ge 0}$ form time-homogeneous Markov chains. 

To study the asymptotic properties of our test, we need to introduce   assumptions C1-C3 and move them and their corresponding detailed discussions to Appendix \ref{sectechcond}.  
In C1, we require the above mentioned Markov chains to be geometrically ergodic. Geometric ergodicity is weaker than the uniform ergodicity condition imposed in the existing reinforcement learning literature \citep[see e.g.][]{bhandari2018,zou2019}. 
In C2, we impose conditions on the set of basis functions $\Psi(\cdot)$ such that $\Psi(\cdot)$ is chosen to yield a good approximation for the Q function. {\color{black}It is worth mentioning that we only require the approximation error to decay at a rate of $o(T^{-1/4})$ instead of $o(T^{-1/2})$. In other words, ``undersmoothing" is not required and the value estimator has a well-tabulated limiting distribution even when the bias of the Q-estimator decays at a rate that is slower than $O(T^{-1/2})$. This result has a number of importation implications. First, it suggests the proposed test will not be overly sensitive to the choice of the number of basis functions. Such a theoretical finding is consistent with our empirical observations in Section \ref{sec:sensiana}. Second, the number of basis functions could be potentially selected by minimizing the prediction loss of the Q-estimator via cross validation.}
We also present examples of basis functions that satisfy C2 in Appendix \ref{subsecC2}. 
In C3, we impose some mild conditions on the action value temporal-difference error, requiring their variances to be non-degenerate.


Let $\{Z_1,\cdots,Z_K\}$ denote the sequence of our test statistics,  where $Z_k=\sqrt{T_k}\widehat{\tau}(T_k)/\widehat{\sigma}(T_k)$. In the following, we study their joint asymptotic distributions. We also present an estimator of their covariance matrix that is consistent under all designs. 

\vspace{-0.2cm}
\begin{thm}[Limiting distributions]\label{thm1}
	Assume C1-C3,   MA, CMIA, CA, and SRA  hold. Assume all immediate rewards are uniformly bounded variables, the density function of $S_0$ is uniformly bounded on $\mathbb{S}$ and $q$ satisfies $q=o(\sqrt{T}/\log T)$. Then under either D1, D2 or D3, we have
	\begin{itemize}
		\vspace{-0.25cm}
		\item  $\{Z_k\}_{1\le k\le K}$ are jointly asymptotically normal;
		\vspace{-0.25cm}
		\item  their asymptotic means are non-positive under $H_0$;
		\vspace{-0.25cm}
		\item their covariance matrix can be consistently estimated by some $\widehat{\bm{\Xi}}$,  whose $(k_1,k_2)$-th element $\widehat{\Xi}_{k_1,k_2}$ equals
		\begin{eqnarray*}
			\sqrt{\frac{T_{k_1}}{T_{k_2}}} \frac{\bm{U}^\top \widehat{\bm{\Sigma}}^{-1}(T_{k_1}) \widehat{\bm{\Omega}}(T_{k_1}) \{\widehat{\bm{\Sigma}}^{-1}(T_{k_2}) \}^\top\bm{U}}{\widehat{\sigma}(T_{k_1})\widehat{\sigma}(T_{k_2})}.
		\end{eqnarray*} 
	\end{itemize}
\end{thm}
This theorem forms the basis of our sequential testing procedure, which we elaborate in the next section. 

\subsection{Sequential monitoring and online updating}\label{secalphaspend}
To sequentially monitor our test, we need to specify the stopping boundary $\{b_k\}_{1\le k\le K}$ such that the experiment is terminated and $H_0$ is rejected when $Z_k>b_k$ for some $k$. 

First, we use the $\alpha$ spending function approach to guarantee the validity of our test. It  requires to specify a monotonically increasing function $\alpha(\cdot)$  that satisfies $\alpha(0)=0$ and $\alpha(T)=\alpha$. 
Some popular choices of the $\alpha$ spending function include\vspace{-0.2cm}
\begin{eqnarray}\label{spend}
	\alpha_1(t)=2-2\Phi\{\Phi^{-1}(1-\alpha/2) \sqrt{T/t} \}\,\,\,\,\hbox{and}\,\,\,\,
	\alpha_2(t)=\alpha (t/T)^{\theta}\,\,\,\,\hbox{for}~~\theta>0,
\end{eqnarray}
where $\Phi(\cdot)$ denotes the normal cumulative distribution function. Adopting the $\alpha$ spending approach, we require $b_k$'s to satisfy 
\begin{eqnarray}\label{eqnbkrequire}
	\prob(\cup_{j=1}^k\{Z_j> b_j\})=\alpha(T_k)+o(1),  \,\,\,\,\,\,\,\,\forall 1\le k\le K. 
\end{eqnarray}
Suppose there exist a sequence of information levels $\{\mathcal{I}_k\}_{1\le k\le K}$ such that 
\begin{eqnarray}\label{eqn:markovproperty}
\Cov(Z_{k_1},Z_{k_2})=\sqrt{\mathcal{I}_{k_1}/\mathcal{I}_{k_2}}+o(1),
\end{eqnarray}
for all $1\le k_1\le k_2$. Then the sequence $\{Z_k\}_{1\le k\le K}$ satisfies the Markov property. The stopping boundary can be efficiently computed based on the numerical integration method detailed in Section 19.2 of \cite{jennison1999}. However, in our setup, condition \eqref{eqn:markovproperty} might not hold when adaptive design is used. As commented in the introduction, this is due to the existence of carryover effects in time. Specifically, when treatment effects are adaptively generated, the behavior policy at difference stages are likely to vary. Due to the carryover effects in time, the state vectors at difference stages have different distribution functions. {\color{black}As such, the asymptotic distribution of the test statistic at each interim stage depends on the behavior policy. Consequently,} the covariance $\Cov(Z_{k_1},Z_{k_2})$ is a very complicated function of $k_1$ and $k_2$ (see e.g., the form of $\widehat{\Xi}_{k_1,k_2}$ in Theorem \ref{thm1}) that can not be represented by \eqref{eqn:markovproperty}. Consequently, the numerical integration method is not applicable. 

{\color{black}Next, we outline a method based on the wild bootstrap \citep{wu1986jackknife}. Then we discuss its limitation and present our proposal, a scalable bootstrap algorithm to determine the stopping boundary. The idea is to generate bootstrap samples $\{\widehat{Z}^{\textrm{MB}}(t)\}_t$  that have asymptotically the same joint distribution as $\{\sqrt{t}\widehat{\sigma}^{-1}(t)(\widehat{\tau}(t)-\tau_0)\}_t$. {\color{black}By the requirement on $\{b_k\}_k$ in \eqref{eqnbkrequire}, we obtain $$\prob\left\{Z_k>b_k|\max_{1\le j<k}(Z_j- b_j)\le 0\right\}=\frac{\alpha(T_k)-\alpha(T_{k-1})}{1-\alpha(T_{k-1})}+o(1).$$
	To implement the test, 
	we thus recursively calculate the threshold $\widehat{b}_k$ as follows,
	\begin{eqnarray}\label{eqn:boundary}
		\prob^*\left\{\widehat{Z}^{\textrm{MB}}(t_k)>\widehat{b}_k|\max_{1\le j< k} (\widehat{Z}^{\textrm{MB}}(t_j)-\widehat{b}_j)\le 0\right\}=\frac{\alpha(T_k)-\alpha(T_{k-1})}{1-\alpha(T_{k-1})},
	\end{eqnarray}
	where $\prob^*$ denotes the probability conditional on the data, and reject $H_0$ when $Z_k^*>\widehat{b}_k$ for some $k$. In practice, the above conditional probability can be approximated via Monte carlo simulations. This forms the basis of the bootstrap algorithm.}

Specifically, let $\{\zeta_t\}_{t\ge 0}$ be a sequence of i.i.d. {\color{black}mean-zero, unit variance} random variables independent of the observed data. Define
\begin{eqnarray}\label{eqn:betaMB}
	\widehat{\bm{\beta}}^{\textrm{MB}}(t)=\widehat{\bm{\Sigma}}^{-1}(t) \left[\frac{1}{t}\sum_{j<t}  \zeta_j \left\{\begin{array}{c}
		\Psi(S_j)(1-A_j)\widehat{\varepsilon}_{j,0} \\
		\Psi(S_j)A_j\widehat{\varepsilon}_{j,1} 
	\end{array}\right\}\right],
\end{eqnarray}
where $\widehat{\varepsilon}_{t,a}$ is the temporal difference error defined. Based on $\widehat{\bm{\beta}}^{\textrm{MB}}(t)$, one can define the bootstrap sample $\widehat{Z}^{\textrm{MB}}(t)=\sqrt{t}\widehat{\sigma}^{-1}(t)\bm{U}^\top \widehat{\bm{\beta}}^{\textrm{MB}}(t)$. Based on the definition of $\widehat{\sigma}(t)$, it is immediate to see that each $\widehat{Z}^{\textrm{MB}}(t)$ follows a standard normal distribution conditional on the data. 

We remark that although the wild bootstrap method is developed under the i.i.d. settings, it is valid under our setup as well. This is due to that under CMIA, $\widehat{\bm{\beta}}(t)-\bm{\beta}^*$ forms a martingale sequence with respect to the filtration $\{(S_j,A_j,Y_j):j<t\}$. It guarantees that the covariance matrices of $\widehat{\bm{\beta}}^{\textrm{MB}}(t)$ and $\widehat{\bm{\beta}}(t)$ are asymptotically equivalent. As such, the bootstrap approximation is valid.  

However, calculating $\widehat{\bm{\beta}}^{\textrm{MB}}(T_k)$ requires $O(T_k)$ operations. The time complexity of the resulting bootstrap algorithm is $O(BT_k)$ up to the $k$-th interim stage,  where $B$ is the total number of bootstrap samples. This can be time consuming when $\{T_k-T_{k-1}\}_{k=1}^K$ are large.  
To facilitate the computation, we observe that in the calculation of $\widehat{\bm{\beta}}^{\textrm{MB}}$, the random noise $\zeta_t$ is generated upon the arrival of each observation. This is unnecessary as we aim to 
approximate the distribution of $\widehat{\bm{\beta}}(\cdot)$ only at finitely many time points $T_1,T_2,\cdots,T_K$.  
}

\begin{algorithm}[!t]
	\caption{The testing procedure}\label{alg1}
	\begin{algorithmic}
		\item \textbf{Input:} number of basis functions $q$, number of bootstrap samples $B$, an $\alpha$ spending function $\alpha(\cdot)$.
		\item \textbf{Initialize: }{\color{black}$T_0=0$, $\mathcal{I}=\{1,2,\cdots,B\}$. \textbf{Set}  $\widehat{\bm{\Omega}}$,
			$\widehat{\bm{\Omega}}^*$, $\widehat{\bm{\Sigma}}_0,\widehat{\bm{\Sigma}}_1$ to zero matrcies, and $\widehat{\bm{\eta}}$,  $\widehat{S}_1,\cdots,\widehat{S}_B$ to zero vectors.} 
		\item \textbf{Compute} $\bm{U}$ according to \eqref{eqn:U}, 
		using either Monte Carlo methods or numerical integration, where $0_q$ denotes a zero vector of length $q$. 
		\item \textbf{For} $k=1$ to $K$:
		\item \qquad {\color{black}\textbf{Step 1.} Online update of ATE.}
		\item \qquad \textbf{For} $t=T_{k-1}$ to $T_k-1$:
		\item \qquad\qquad $\widehat{\bm{\Sigma}}_a=(1-t^{-1}) \widehat{\bm{\Sigma}}_a+t^{-1}\Psi(S_t)\mathbb{I}(A_t=a) \{ \Psi(S_t)-\gamma \Psi(S_{t+1})\mathbb{I}(A_{t+1}=a) \}^\top$, $a=0,1$;
		\item \qquad\qquad $\widehat{\bm{\eta}}_a=(1-t^{-1})\widehat{\bm{\eta}}_a+t^{-1}\Psi(S_t)\mathbb{I}(A_t=a) Y_t$.
		\item \qquad \textbf{Set} $\widehat{\beta}_{a}=\widehat{\bm{\Sigma}}_a^{-1} \widehat{\bm{\eta}}_a$ for $a\in \{0,1\}$ {\color{black}and $\widehat{\tau}=\bm{U}^\top \widehat{\bm{\beta}}$. } 
		\item \qquad {\color{black}\textbf{Step 2.} Online update of the variance estimator.}
		\item \qquad {\color{black}\textbf{Initialize} $\widehat{\bm{\Omega}}^*$ to a zero matrix.}
		\item \qquad \textbf{For} {$t=T_{k-1}$ to $T_k-1$}:
		\item \qquad \qquad $\widehat{\varepsilon}_{t,a}=Y_t+\gamma \Psi^\top(S_{t+1})\widehat{\beta}_{a}-\Psi^\top(S_t)\widehat{\beta}_{a}$ for $a=0,1$;
		\item \qquad \qquad $\widehat{\bm{\Omega}}^*=\widehat{\bm{\Omega}}^*+\{\Psi(S_t)^\top (1-A_t) \widehat{\varepsilon}_{t,0},\Psi(S_t)^\top A_t \widehat{\varepsilon}_{t,1}\}^\top  \{\Psi(S_t)^\top (1-A_t) \widehat{\varepsilon}_{t,0},\Psi(S_t)^\top A_t \widehat{\varepsilon}_{t,1}\}$.
		\item \qquad \textbf{Set}  $\widehat{\bm{\Sigma}}$ to a block diagonal matrix by aligning $\widehat{\bm{\Sigma}}_0$ and $\widehat{\bm{\Sigma}}_1$ along the diagonal of $\widehat{\bm{\Sigma}}$;
		\item \qquad \textbf{Set} $\widehat{\bm{\Omega}}=T_{k}^{-1} (T_{k-1} \widehat{\bm{\Omega}}+\widehat{\bm{\Omega}}^*)$ and the variance estimator $\widehat{\sigma}^2= \bm{U}^\top\widehat{\bm{\Sigma}}^{-1}\widehat{\bm{\Omega}} \{\widehat{\bm{\Sigma}}^{-1}\}^\top \bm{U}$. 
		\item \qquad {\color{black} \textbf{Step 3.} Bootstrap test statistic.} 
		\item \qquad \textbf{For} $b=1$ to $B$:	
		\item \qquad \qquad Generate $e_k^{(b)}\sim N(0,I_{4q})$; \item \qquad \qquad $\widehat{S}_b=\widehat{S}_b+\widehat{\bm{\Omega}}^{*1/2}e_k^{(b)}$; \item \qquad \qquad$\widehat{Z}_b^*=T_k^{-1/2} \widehat{\sigma}^{-1} \bm{U}^{\top} \widehat{\bm{\Sigma}}^{-1} \widehat{S}_b$;
		\item \qquad\textbf{Set} $z$ to be the upper $\{\alpha(t)-|\mathcal{I}^c|/B\}/(1-|\mathcal{I}^c|/B)$-th percentile of $\{\widehat{Z}_b^*\}_{b\in \mathcal{I}}$.
		\item \qquad\textbf{Update} $\mathcal{I}$ as $\mathcal{I}\leftarrow \{ b\in \mathcal{I}: \widehat{Z}_b^*\le z \}$;
		\item \qquad{\color{black}\textbf{Step 4.} Reject or not?} 
		\item \qquad\textbf{Reject} the null if $\sqrt{T_k}\widehat{\sigma}^{-1}\widehat{\tau}>z$.
	\end{algorithmic}	
\end{algorithm}	

Finally, we present our bootstrap algorithm to determine $\{b_k\}_{1\le k\le K}$, based on Theorem \ref{thm1}. Let $\{e_k\}_{1\le k\le K}$ be a sequence of i.i.d $N(0,I_{4q})$ random vectors, where 
$I_{J}$ stands for a $J\times J$ identity matrix for any $J$. Let $\widehat{\bm{\Omega}}(T_0)$ be a zero matrix. At the $k$-th stage, we compute the bootstrap sample
\begin{eqnarray*}
	\widehat{Z}_k^*=\frac{\bm{U}^{\top}\widehat{\bm{\Sigma}}^{-1}(T_k)}{\sqrt{T_k}\widehat{\sigma}(T_k)} \sum_{j=1}^k  \{T_j\widehat{\bm{\Omega}}(T_j)-T_{j-1}  \widehat{\bm{\Omega}}(T_{j-1})\}^{1/2} e_j.
\end{eqnarray*}

A key observation is that, conditional on the observed dataset, the covariance of $\widehat{Z}_{k_1}^*$ and $\widehat{Z}_{k_2}^*$ equals
\begin{eqnarray*}
	\frac{\bm{U}^{\top} \widehat{\bm{\Sigma}}^{-1}(T_{k_1})}{\sqrt{T_{k_1} T_{k_2}} 
		\widehat{\sigma}(T_{k_1}) \widehat{\sigma}(T_{k_2})}  [\sum_{j=1}^{k_1} \{T_j\widehat{\bm{\Omega}}(T_j)-T_{j-1}  \widehat{\bm{\Omega}}(T_{j-1})\}  ] \{\widehat{\bm{\Sigma}}^{-1}(T_{k_2})\}^{-1}\bm{U}=
	\widehat{\Xi}_{k_1,k_2}.
\end{eqnarray*}
By Theorem \ref{thm1}, the covariance matrices of $\{Z_k\}_k$ and $\{Z_k^*\}_k$ are asymptotically equivalent. In addition, the limiting distributions of $\{Z_k\}_k$ and $\{Z_k^*\}_k$ are multivariate normal with zero means. As such, the joint distribution of $\{Z_k\}_{1\le k\le K}$ can be well approximated by that of $\{Z_k^*\}_{1\le k\le K}$ conditional on the data. The rejection boundary can thus be computed in a similar fashion as in \eqref{eqn:boundary}. 
\begin{thm}[Type-I error]\label{thm2}
	Suppose that the conditions of Theorem \ref{thm1} hold and $\alpha(\cdot)$ is continuous. Then the proposed thresholds satisfy
	$\prob(\bigcup_{j=1}^k\{Z_j> \widehat{b}_j\})\le \alpha(T_k)+o(1)$, 
	for all $1\le k\le K$  under $H_0$. The equality 
	holds when $\tau_0=0$. 
\end{thm}

Theorem \ref{thm2} implies that the type-I error rate of the proposed test is well controlled. 
When $\mathrm{ATE}= 0$,
the equality in Theorem \ref{thm2} holds. The rejection probability achieves the nominal level under $H_0$.
We next investigate the power property of our test. 
\begin{thm}[Power]\label{thm3}
	Suppose that  the conditions of  Theorem \ref{thm2} hold. Assume $\tau_0\gg T^{-1/2}$, then 
	$\prob(\bigcup_{j=1}^k\{Z_j> \widehat{b}_j\})\to 1$. 
	Assume $\tau_0=T^{-1/2} h$ for some $h>0$. Then 
	$\lim_{T\to \infty}[\prob(\cup_{j=1}^k\{Z_j> \widehat{b}_j\})-\alpha(T_k)]>0$.
\end{thm}
Combining Theorems \ref{thm2} and \ref{thm3} yields the consistency of our test. The second assertion in Theorem \ref{thm3} implies that our test has non-negligible powers against local alternatives converging to $H_0$ at the $T^{-1/2}$ rate. When the signal decays at a slower rate, the power of our test approaches 1. 

Since we use a linear basis function to approximate the Q-function, the regression coefficients $\widehat{\bm{\beta}}(t)$ as well as their covariance estimator can be online updated as batches of observations arrive at the end of each interim stage. As such, our test can be implemented online. We summarize our procedure in Algorithm \ref{alg1}. Recall that $q$ is the number of basis functions. As the $k$th interim stage, the time complexities of Steps 1-3 in Algorithm \ref{alg1} are dominated by $O\{q^2(T_k-T_{k-1})+q^3\}$, $O\{q^2(T_k-T_{k-1})+q^3\}$ and $O(Bq^2+q^3)$, respectively. As such, 
the time complexity of Algorithm \ref{alg1} is dominated by $O(BKq^2+Tq^2+Kq^3)$. In contrast, one can show that the classical wild bootstrap algorithm would take at least $\Omega(BTq^2+Kq^3)$ number of flops and is much more computationally intensive when $T\gg K$, which is case in phase 3 clinical trials and our real data application. 

{\color{black}To conclude this section, we remark that a few bootstrap algorithms have been developed in the RL literature for policy evaluation. Specifically, \cite{hanna2017bootstrapping} and \cite{hao2021bootstrapping} proposed to use bootstrap for uncertainty quantification in off-policy evaluation. These algorithms require the number of trajectories to diverge to infinity to be consistent and are thus not applicable to our setting where there is only one trajectory in the experiment. In addition, they are developed in offline settings and do not allow online updating. \cite{ramprasad2021online} developed a bootstrap algorithm for policy evaluation in online settings. Their algorithm generates bootstrap samples upon the arrival of each observation and is thus more computationally intensive than the proposed algorithm. 
}

\section{Simulation study}\label{secnumerical}
\subsection{Settings and implementation}\label{sec:setting}
Simulated data of states and rewards was generated as follows, 
\begin{align*}
	S_{1,t}&=(2 A_{t-1}-1) S_{1,(t-1)}/2 +S_{2,(t-1)}/4 +\delta A_{t-1} +\varepsilon_{1,t}, \\
	S_{2,t}&=(2 A_{t-1}-1) S_{2,(t-1)}/2 +S_{1,(t-1)}/4+\delta A_{t-1} + \varepsilon_{2,t}, \,\,\,\,
	Y_{t}=1+( S_{1,t}+ S_{2,t})/2 + \varepsilon_{3,t}, 
\end{align*}
where the random errors $\{\varepsilon_{j,t}\}_{j=1,2, 0\le t\le T}$ are i.i.d $N(0,0.5^{2})$ and  $\{\varepsilon_{3,t}\}_{0\le t\le T}$ are i.i.d $N(0,0.3^2)$. Let $S_t=(S_{1,t}, S_{2,t})^\top$ denote the state at time $t$. Under this model, treatments have delayed effects on the outcomes, as in Example 2. The parameter $\delta$ characterizes the degree of such carryover effects. When $\delta=0$, $\tau_0=0$ and $H_0$ holds. When $\delta>0$,  $H_1$ holds. Moreover, $\tau_0$ increases as $\delta$ increases. 

We set $K=5$ and $(T_1,T_2,T_3,T_4,T_5)=(300,375,450,525,600)$.  The discounted factor $\gamma$ is set to $0.6$ and $\mathbb{G}$ is chosen as the initial state distribution.
We consider three behavior policies, according to the designs D1-D3, respectively. For the behavior policy in D1, we set $b^{(0)}(s)=0.5$ for any $s\in \mathbb{S}$. For the behavior policy in D3, we use an $\epsilon$-greedy policy and set 
$b^{(k)}(s)=\epsilon/2+(1-\epsilon) 
\mathbb{I}(\Psi(s)^\top (\widehat{\beta}_{1}(T_{k})-\widehat{\beta}_{0}(T_{k}))>0),$
with $\epsilon=0.1$, 
for any $k\ge 1$ and $s\in \mathbb{S}$. 

For each design, we further consider five choices of $\delta$, corresponding to $0, 0.05, 0.1, 0.15$ and $0.2$. 
The significance level $\alpha$ is set to $0.05$ in all cases. To implement our test, we choose two $\alpha$-spending functions, corresponding to $\alpha_1(\cdot)$ and $\alpha_2(\cdot)$ given in \eqref{spend}. The hyperparameter $\theta$ in $\alpha_2(\cdot)$ is set to $3$. The number of bootstrap sample is set to $1000$. In addition, we consider the following polynomial basis function,
$\Psi(s)=\Psi(s_1, s_2) = (1, s_1, s_1^2, \cdots, s_1^J, s_2, s_2^2, \cdots, s_2^J)^\top$,
with $J=4$. 

\begin{figure}[!t]
	\centering
	\begin{tabular}{ccc}
		\hspace{-0.5cm}\includegraphics[scale=0.43]{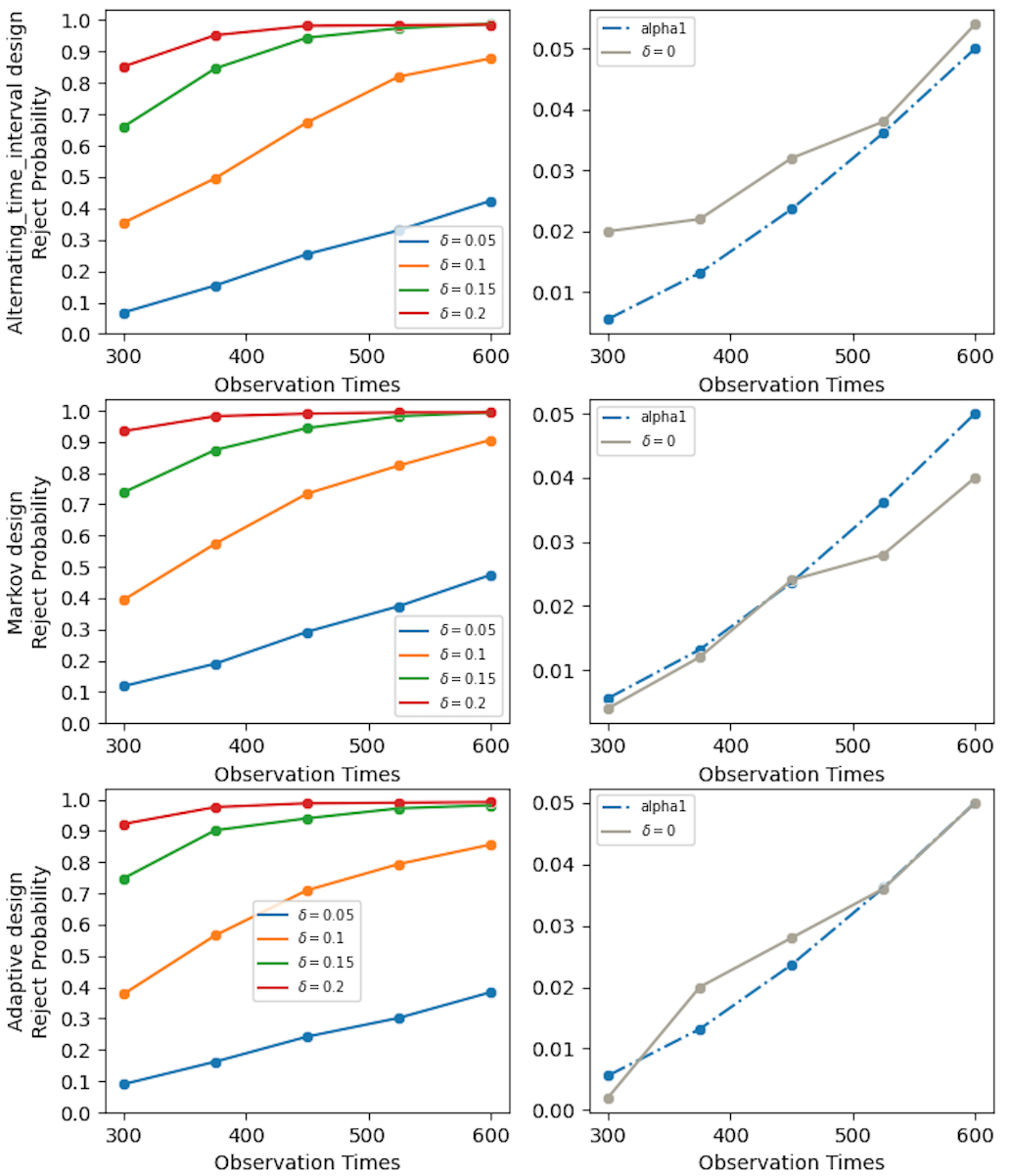} &
		&
		\hspace{-0.5cm}\includegraphics[scale=0.43]{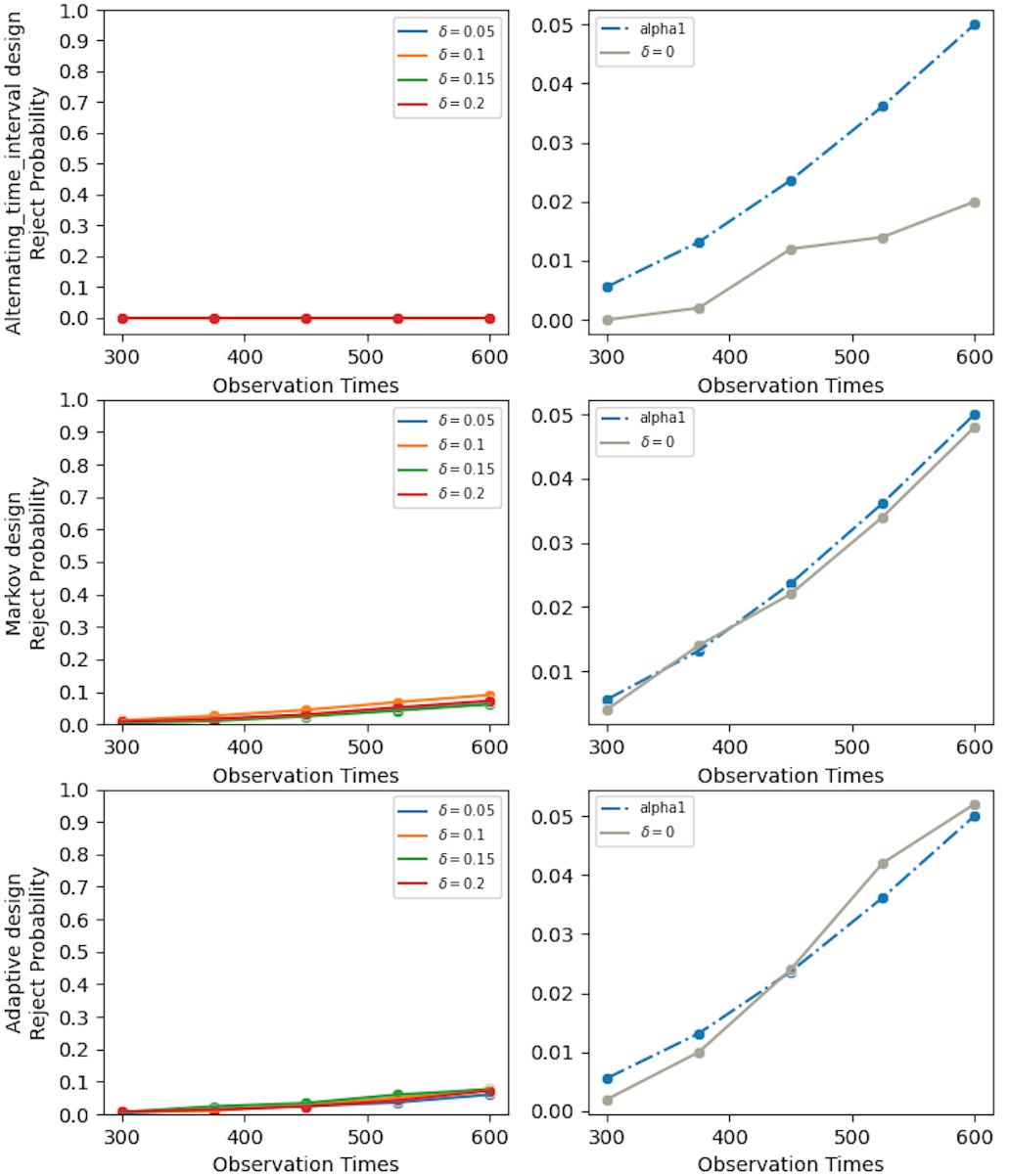} \\
		\small (a) The proposed test under $H_1$ and $H_0$ & & \small (b) Two-sample t-test under $H_1$ and $H_0$\\
		\small (from left plots to right plots) & & \small (from left plots to right plots)
	\end{tabular}
	\caption{Empirical rejection probabilities of our test and the two-sample t-test with  $\alpha(\cdot)=\alpha_1(\cdot)$. Settings correspond to the alternating-time-interval, adaptive and Markov design, from top plots to bottom plots.}
	\label{fig1}
\end{figure}

\begin{figure}[!t]
	\centering
	\begin{tabular}{ccc}
		\hspace{-0.5cm}\includegraphics[scale=0.43]{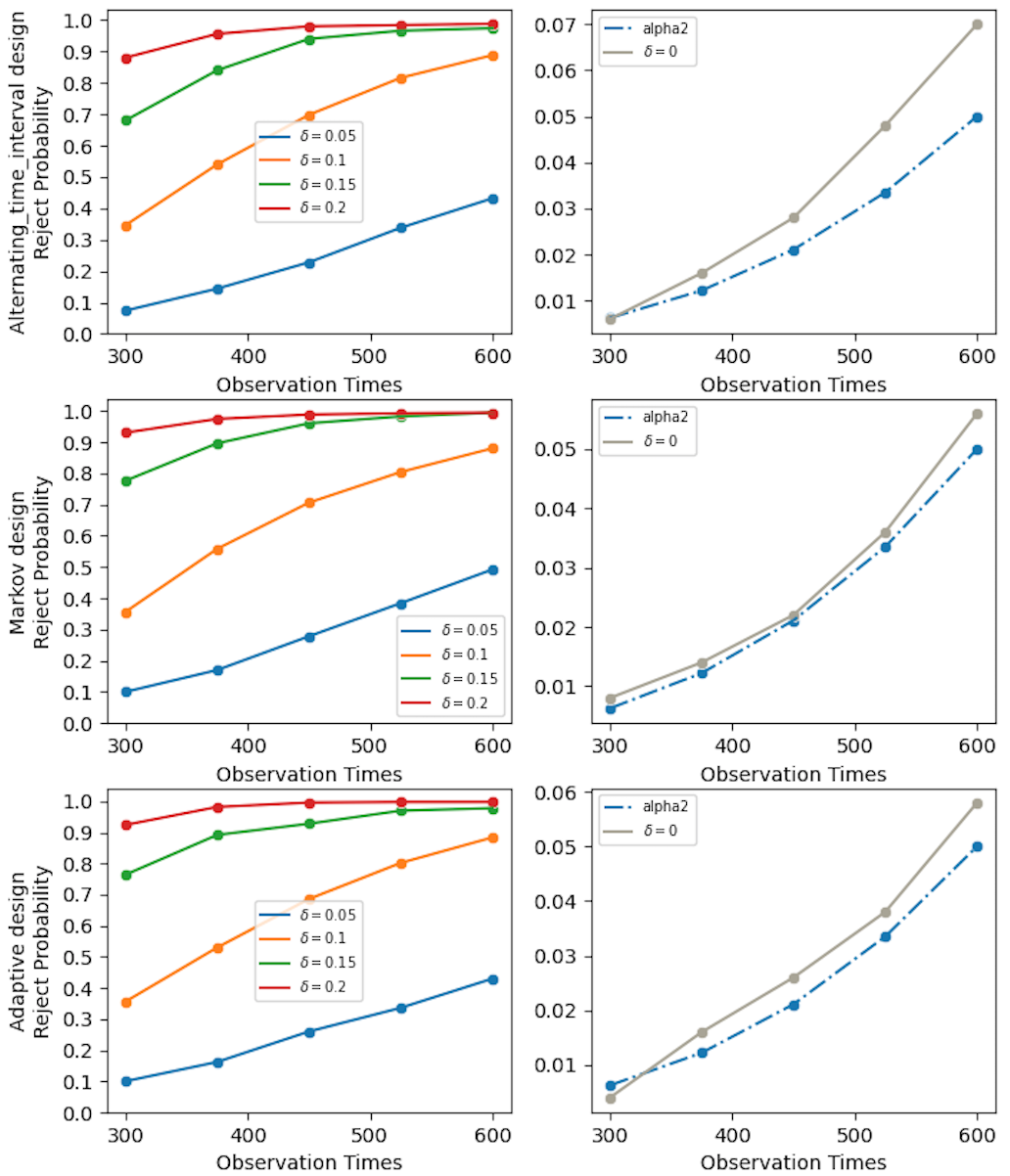} &
		&
		\hspace{-0.5cm}\includegraphics[scale=0.43]{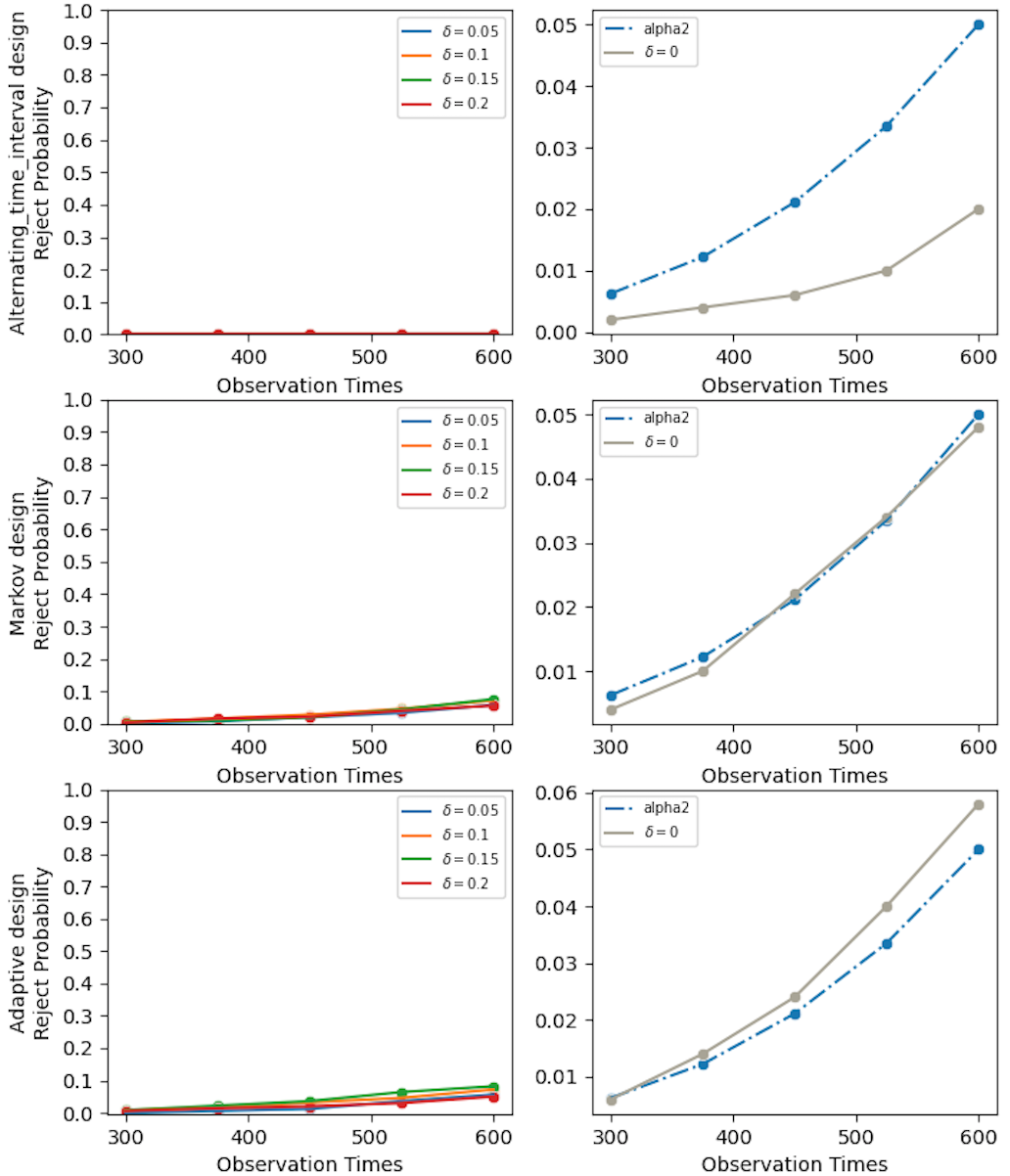} \\ 
		\small (a) The proposed test under $H_1$ and $H_0$ & \hspace{-1cm} & \small (b) Two-sample t-test under $H_1$ and $H_0$\\
		\small (from left plots to right plots) & & \small (from left plots to right plots)
	\end{tabular}
	\caption{Empirical rejection probabilities of our test and the two-sample t-test with  $\alpha(\cdot)=\alpha_2(\cdot)$. Settings correspond to the alternating-time-interval, adaptive and Markov design, from top plots to bottom plots.}
	\label{fig2}
\end{figure}

All experiments run on a macbook pro with a dual-core 2.7 GHz processor. 
Implementing a single test takes one second. 
Figures \ref{fig1}(a) and \ref{fig2}(a) 
depict the empirical rejection probabilities of our test statistics at different interim stages under $H_0$ and $H_1$ with different combinations of $\delta$, $\alpha(\cdot)$ and the designs. These rejection probabilities are aggregated over 500 simulations. We also plot $\alpha_1(\cdot)$ and $\alpha_2(\cdot)$ under $H_0$. Based on the results, it can be seen that under $H_0$, the Type-I error rate of our test is well-controlled and close to the nominal level at each interim stage in most cases. Under $H_1$, the power of our test increases as $\delta$ increases, showing the consistency of our test procedure.

\subsection{Comparison with baseline methods}\label{sec:compare}
\begin{figure}[!t]
	\centering
	\begin{tabular}{ccc}
		\includegraphics[scale=0.4]{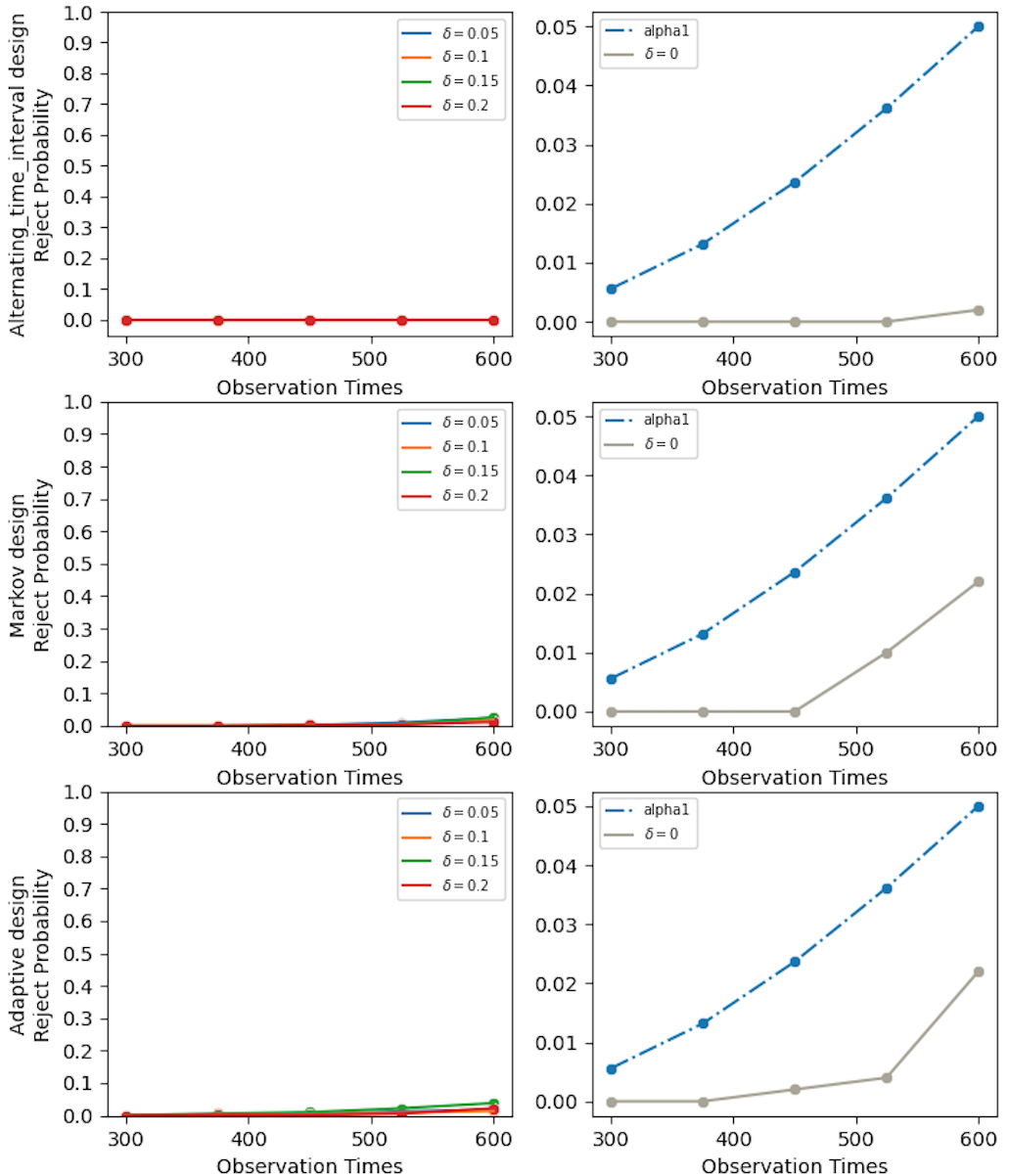}
	\end{tabular}
	\caption{Empirical rejection probabilities of the modified version of the O'Brien \& Fleming sequential test developed by \cite{kharitonov2015}. The left panels depicts the empirical type-I error and the right panels depicts the empirical power. Settings correspond to the alternating-time-interval, adaptive and Markov design, from top plots to bottom plots.}
	\label{fig1.5}
\end{figure}

To further evaluate our method, we first compare it with the classical two-sample t-test and a modified version of modified versions of the O'Brien \& Fleming sequential test developed by \cite{kharitonov2015}. We remark that the current practice of policy evaluation in most two-sided marketplace platforms is to employ classical two-sample t-test. Specifically, for each $T_k$, we apply the t-test to the data $\{A_t,Y_t\}_{0\le t\le T_k}$ and plot the corresponding empirical rejection probabilities in Figures \ref{fig1}(b) and \ref{fig2}(b). Figure \ref{fig1.5} depicts the empirical rejection probabilities of the modified version of the O’Brien \& Fleming sequential test. We remark that such a test requires equal sample size $T_1 = T_k - T_{k-1}$ for $k=2,\cdots,K$ and is not directly applicable to our setting with unequal sample size. To apply such a test, we modify the decision time and set $(T_1, T_2, T_3, T_4, T_5) = (120, 240, 360, 480, 600)$. 
As shown in these figures, all these tests fail to detect any carryover effects and do not have power at all. 

\begin{figure}[!t]
	\centering
	\small
	\begin{tabular}{cc}
		\hspace{-2em}\includegraphics[scale=0.37]{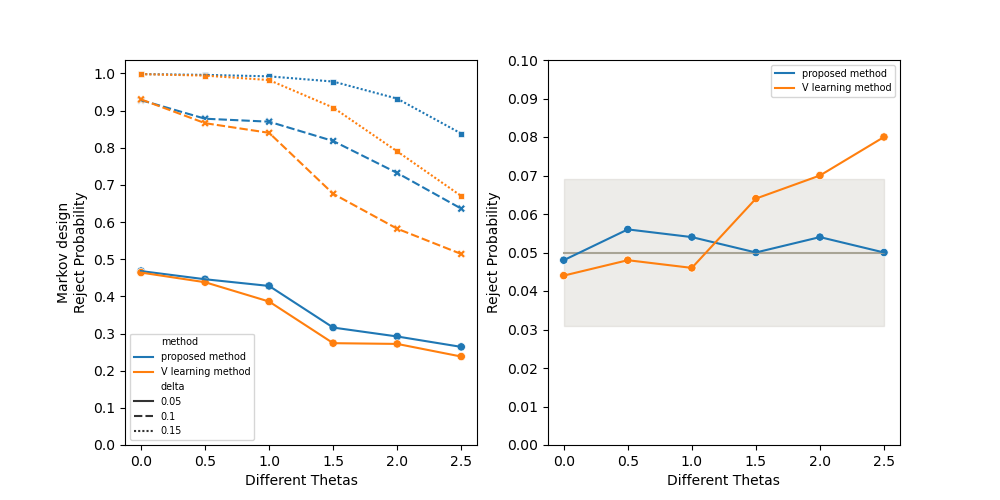} &	\hspace{-3em}\includegraphics[height=4.7cm,width=9cm]{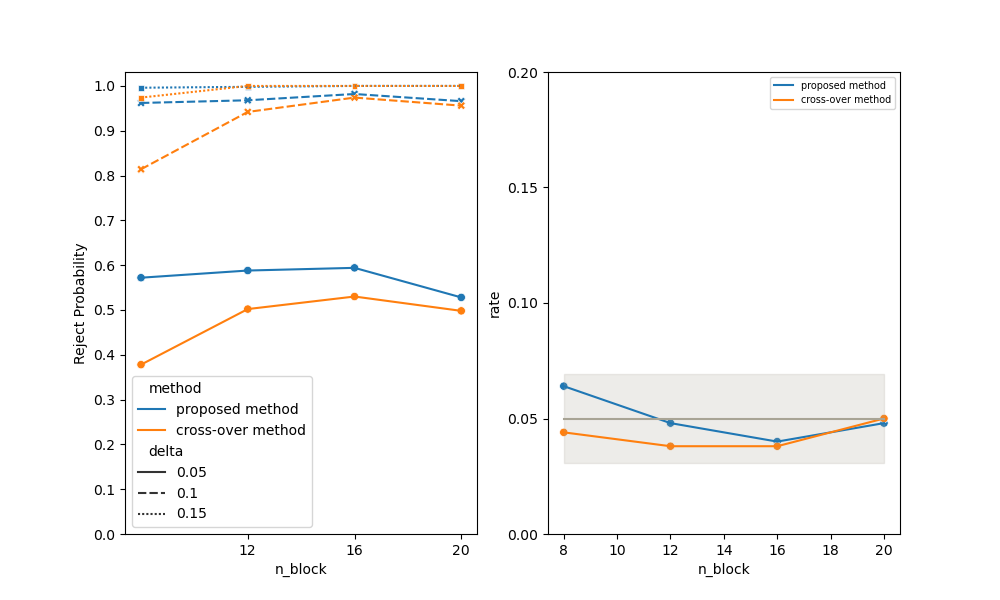} \\
		(a) The proposed test and the test & \hspace{-2em}(b) The project test and the t-test derived\\
		based on V-learning under $H_1$ and $H_0$ & \hspace{-2em}based on analysis of crossover trials under\\
		(from left plots to right plots) &\hspace{-2em} $H_1$ and $H_0$ (from left plots to right plots)
	\end{tabular}
	\caption{(a) Empirical rejection probabilities of the proposed test and the test based on V-learning. (b) Empirical rejection probabilities of the proposed test and the test derived based on analysis of crossover trials. The shaded area corresponds to the interval $[0.05-1.96\textrm{MCE},0.05+1.96\textrm{MCE}]$ where MCE denotes the Monte Carlo error $\sqrt{0.05\times 0.95/500}$.
	}\label{figa}
\end{figure}

{\color{black}We next compare the proposed test with the test based on the V-learning method developed by \citet{luckett2019}. As we have commented, V-learning does not allow sequential testing. So we focus on settings where the decision is made once at $T=600$. In addition, V-learning requires the propensity score to be bounded away from 0 and 1. To meet the positivity assumption, we generate the actions according to the Markov design where $\prob(A_t=1|S_t)=\textrm{sigmoid}(\theta S_{1,t}+\theta S_{2,t})$. 
Both tests require to specify the discounted factor $\gamma$. We fix $\gamma=0.8$. 
Results are reported in Figure \ref{figa}(a), aggregated over 500 simulations. It can be seen for large $\theta$, the test based on V-learning cannot control the type-I error and has smaller power than our test when $\delta$ is large. This is because V-learning uses inverse propensity score weighting. In cases where $\theta$ is large, the propensity score can be close to zero or one for some sample values, making the resulting test statistic unstable. 
	
Finally, we compare the proposed test with a t-test based on analysis of crossover trials 
\citep[see e.g.,][]{jones1989design}. 
We remark that such a test requires the data to be generated from crossover designs and cannot be applied under D1, D2 or D3. In addition, most crossover trials require to recruit multiple subjects/patients to estimate the carryover effect. The resulting tests are not directly applicable to our setting where only one subject receives a sequence of treatments over time. 
In Appendix \ref{secaddsimu}, we develop a t-test for the carryover effect under our setting, based on  analysis of $2\times 2$ crossover trials. For simplicity, we focus on settings where the decision is made once at $T=600$. In Figure \ref{figa}(b), we report the empirical rejection probabilities of such a test and the proposed test under several crossover designs with different number of blocks. Please refer to Appendix \ref{secaddsimu} for more details about the design and the test. It can be seen that the proposed test is more powerful in most cases. 
}

\subsection{Sensitivity analysis}\label{sec:sensiana}
\begin{figure}[!t]
	\centering
	\begin{tabular}{ccc}
		\hspace{-0.5cm}\includegraphics[scale=0.4]{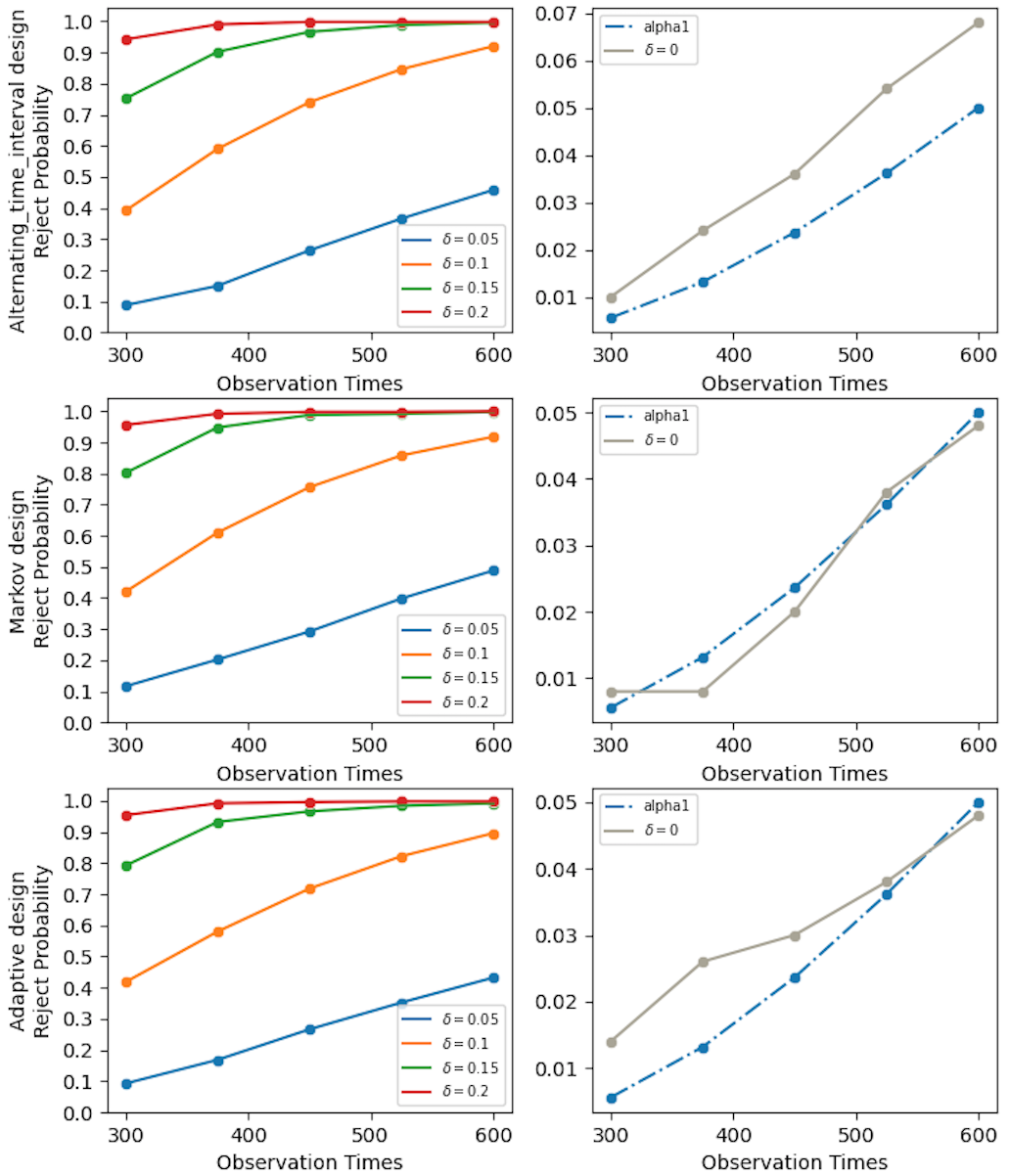} &
		&
		\hspace{-0.5cm}\includegraphics[scale=0.4]{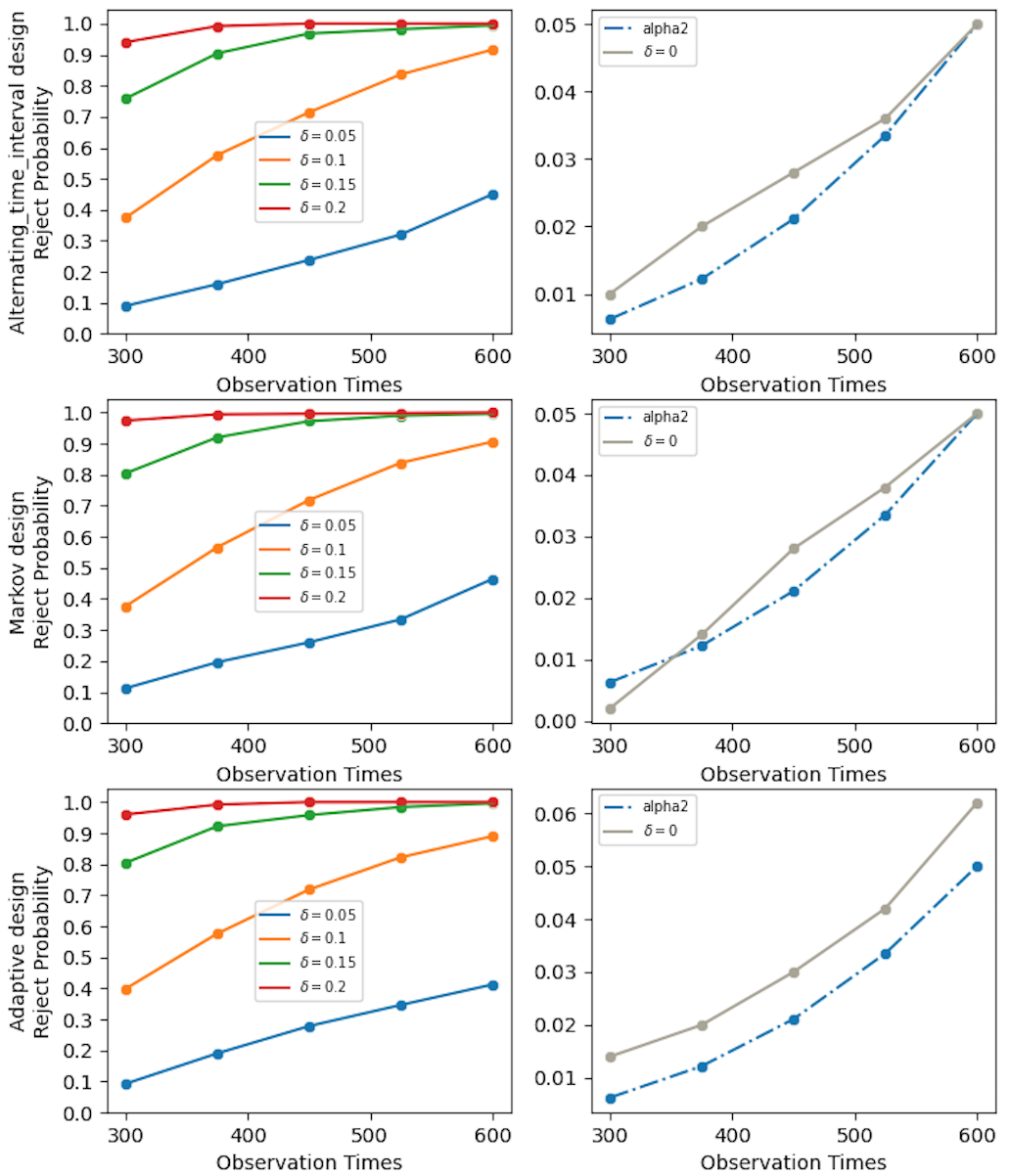} \\
		\small (a) The proposed test under $H_1$ and $H_0$ (from & & \small (b) The proposed test under $H_1$ and $H_0$ (from\\
		\small  left plots to right plots). $J=3$, $\alpha(\cdot)=\alpha_1(\cdot)$. & & \small left plots to right plots). $J=3$, $\alpha(\cdot)=\alpha_2(\cdot)$.  \\
		\hspace{-0.5cm}\includegraphics[scale=0.4]{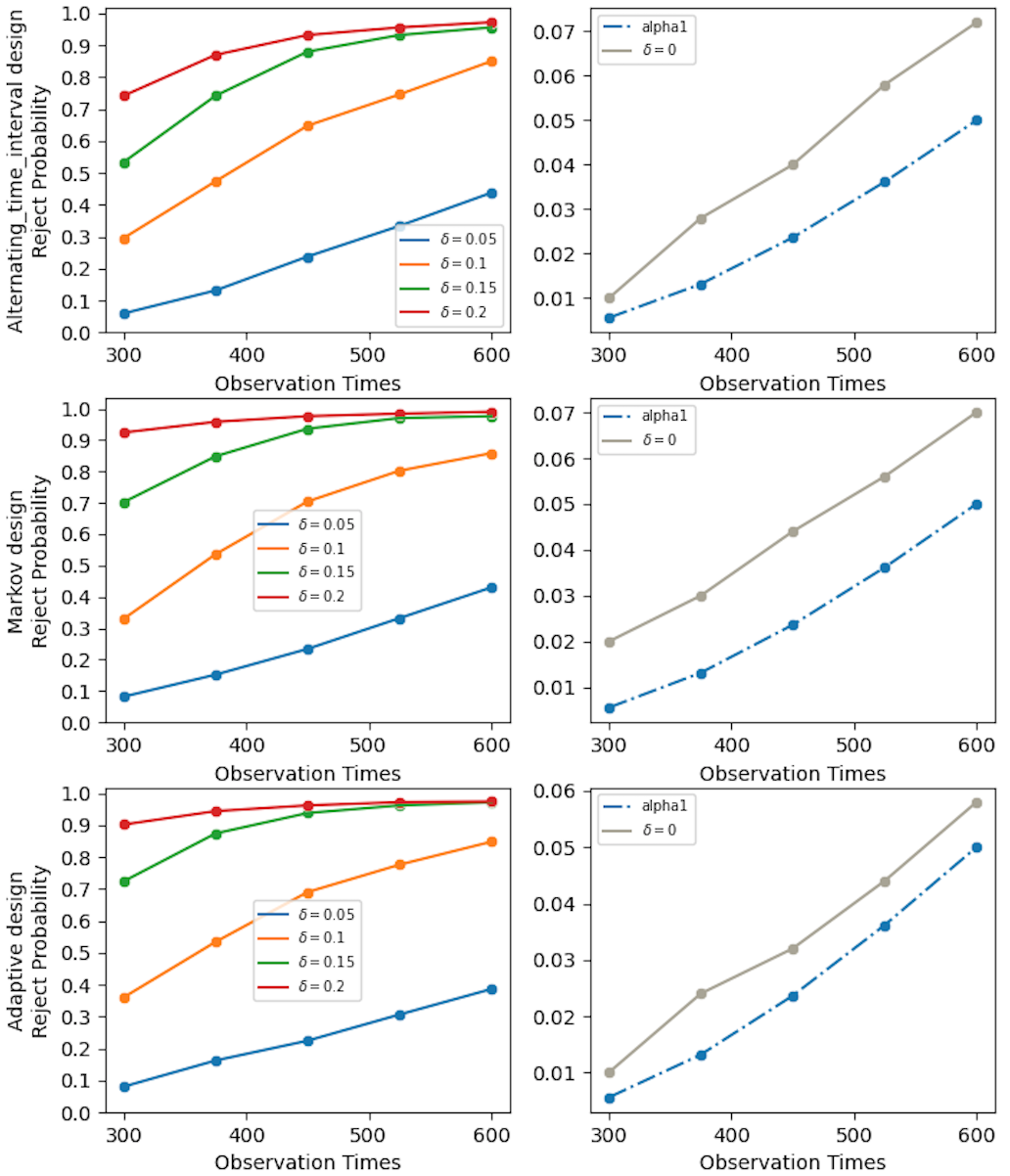} &
		&
		\hspace{-0.5cm}\includegraphics[scale=0.4]{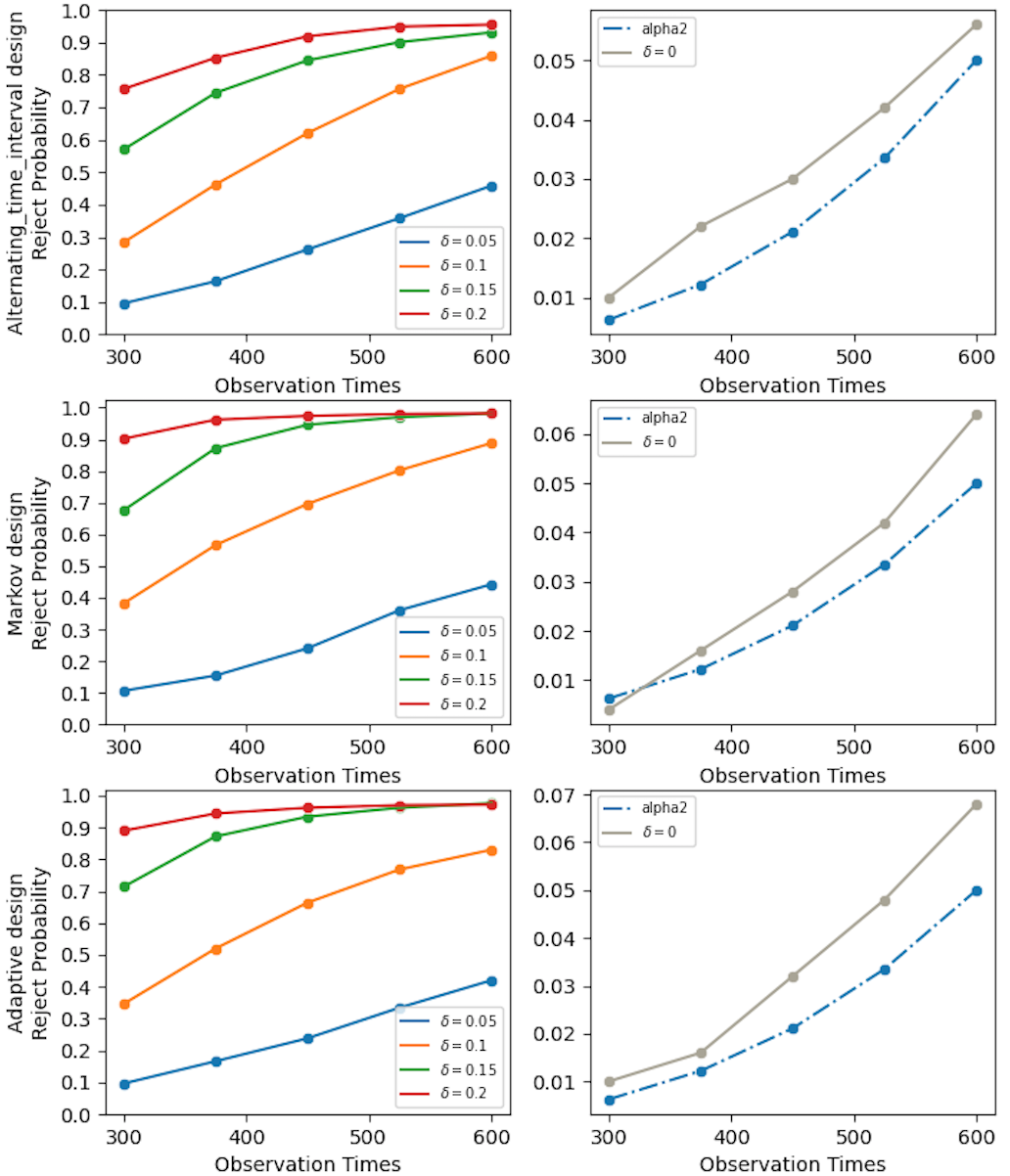} \\
		\small (c) The proposed test under $H_1$ and $H_0$ (from & & \small (d) The proposed test under $H_1$ and $H_0$ (from \\
		\small left plots to right plots). $J=5$, $\alpha(\cdot)=\alpha_1(\cdot)$. & & \small left plots to right plots).  $J=5$, $\alpha(\cdot)=\alpha_2(\cdot)$.
	\end{tabular}
	\caption{Empirical rejection probabilities of our test. Settings correspond to the alternating-time-interval, adaptive and Markov design, from top plots to bottom plots.}
	\label{fig3}
\end{figure}

In Section \ref{sec:setting}, we set the number of polynomial basis function $J$ to 4. We also tried some other values of $J$ by setting $J$ to 3 and 5. Results are reported in Figure \ref{fig3}. It can be seen that the resulting tests have very similar performance and is not sensitive to the choice of $J$. {\color{black}In Appendix \ref{secaddsimu}, we fixed $J$ to 4 and tried some other values of $\gamma\in$ (0.1, 0.3, 0.5, 0.9). Results are reported in Figure \ref{figS3}. It can be seen that our test controls the type-I error in most cases. In addition, its power increases with $\gamma$. This is consistent with the following observation: $\gamma$ characterizes the balance between the short-term and long-term treatment effects. Under the current setup, there is no short-term treatment effects. The value difference increases with $\gamma$. It is thus expected that our test has better power properties for large values of $\gamma$.}

\section{Real data application}\label{sec:realdata}
We apply the proposed test to a real dataset from a large-scale ride-sharing platform. The purpose of this study is to compare the performance of a newly developed order dispatching strategy with a standard control strategy used in the platform.  For a given order, the new strategy will dispatch it to a nearby driver that has not yet finished their previous ride request, but almost. In comparison, the standard control assigns orders to drivers that have completed their ride requests. {\color{black}The new strategy is expected to reduce the chance that the customer will cancel an order in regions with only a few available drivers. It is expected to meet more call orders and increase drivers' income on average.}

The experiment is conducted at a given city from December 3rd to December 16th. Dispatch strategies are executed based on alternating half-hourly time intervals. 
We also apply our test to a data from an A/A experiment (which compares the baseline strategy against itself), 
conducted from November 12th to November 25th. Note that it is conducted at a different time period from the A/B experiment. {\color{black}The A/A experiment is employed as a sanity check for the validity of the proposed test. We expect that our test will not reject $H_0$ when applied to this dataset, since the two strategies used are essentially the same.}

Both experiments last for two weeks. Thirty-minutes is defined as one time unit. 
{\color{black}We set $K=8$ and $T_k=48\times (k+6)$ for $k=1,\dots,8$. That is, the first interim analysis is performed at the end of the first week, followed by seven more at the end of each day during the second week.}
{\color{black}We discuss more about the experimental design in Section \ref{sec:expdesign}. We choose the overall drivers' income in each time unit as the response}. The new strategy is expected to reduce the answer time of passengers and increase drivers' income. Three time-varying variables are used to construct the state. The first two correspond to the number of requests (demand) and drivers' online time (supply) during each 30-minutes time interval. These factors are known to have large impact on drivers' income. The last one is the supply and demand equilibrium metric. This variable characterizes the degree that supply meets the demand and serves as an important mediator between past treatments and future outcomes. 

\begin{figure}[!t]
	\centering
	\includegraphics[width=8cm,height=3.5cm]{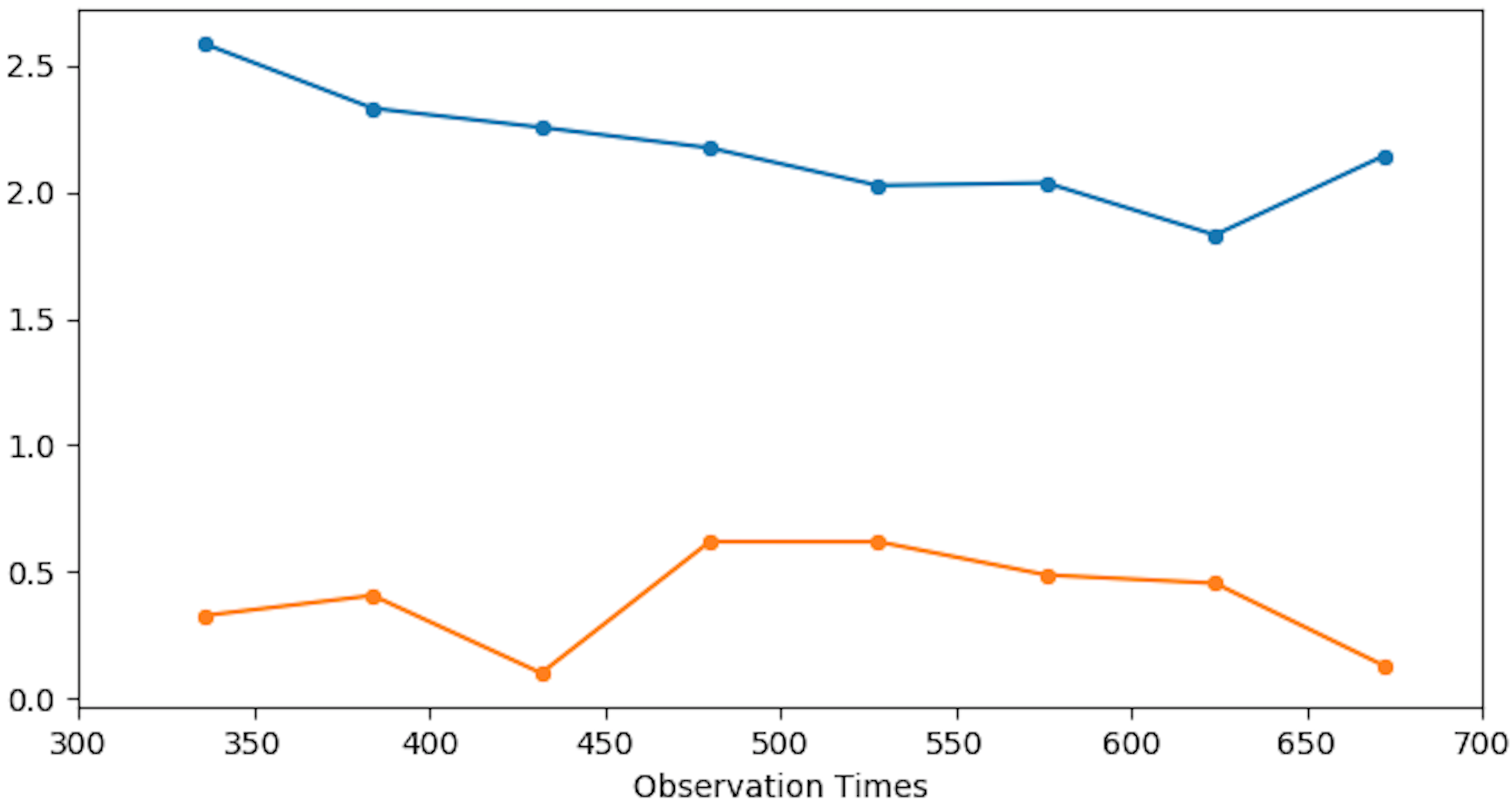}
	\includegraphics[width=8cm,height=3.5cm]{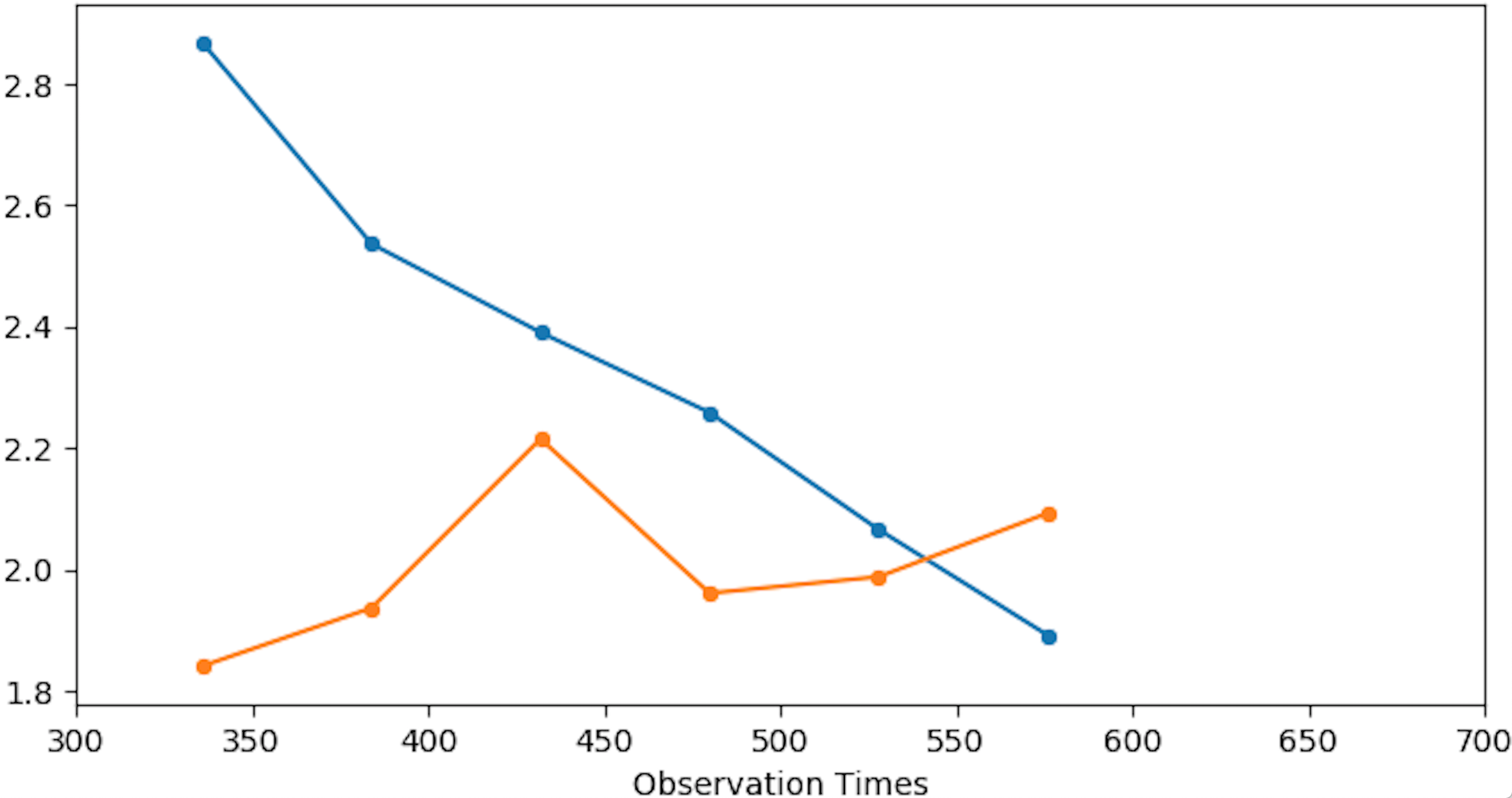}
	\caption{Our test statistic (the orange line) and the rejection boundary (the blue line) in the A/A (left plot) and A/B (right plot) experiments.}\label{fig4}
\end{figure}
To implement our test, we  set $\gamma=0.6$, $B=1000$ and use a fourth-degree polynomial basis for $\Psi(\cdot)$, as in simulations. 
We use $\alpha_1(\cdot)$ as the spending function for interim analysis and set $\alpha=0.05$. 
The test statistic and its corresponding
rejection boundary at each interim stage are plotted in Figure \ref{fig4}. 
It can be seen that our test is able to conclude, 
at the end of the 12th day, that the new order dispatch strategy can significantly increase drivers' income. When applied to the data from the A/A experiment, we fail to reject $H_0$, as expected. {\color{black}We remark that early termination of the A/B experiment is beneficial to both the platform and the society. First, take this particular experiment as an example, we find that the new strategy reduces the answer time of orders by 2\%, leading to almost 2\% increment of drivers' income. If we were to wait until Day 14, drivers would lose 2\% income and customers would have to wait longer on two days. The benefits are considerable by taking the total number of drivers and customers in the city into account. In addition, the platform can benefit a lot from the increase in the driver income, as they take a fixed proportion of the driving fee from all completed trips. Second, the platform needs to conduct a lot of A/B experiments to investigate various policies. A reduction in the experiment duration facilitates the process, allowing the platform to evaluate more policies within the same time frame. These policies have the potential to further improve the driver income and the customer satisfaction, providing safer, quicker and more convenient transportation.}

For comparison, we also apply the two-sample t-test to the data collected from the A/B experiment. The corresponding p-value is 0.18. This result is consistent with our findings. Specifically, the treatment effect at a given time affects the distribution of drivers in the future, inducing interference in time. As shown in the toy example (see Section \ref{sec:toyexample}), the t-test cannot detect such carryover effects, leading to a low power. Our procedure, according to Theorem 2, has enough powers to discriminate $H_1$ from $H_0$.

\section{Discussion}\label{sec:dis}
We discuss extensions of the proposed method to evaluate dynamic policies, the experimental
design in our data application and the off-policy evaluation problem in this section. In
Appendix \ref{app:moredis} of the supplementary article, we discuss extensions of our proposal to high-dimensional models, the methodological difference between our proposal and the V-learning
method, and the literature on crossover trials.
\subsection{Dynamic policies}\label{sec:dynamic}
{\color{black}In this paper, we focus on comparing the long-term treatment effects between two nondynamic policies. The proposed method can be easily extended to handle dynamic policies as well. Specifically, consider two time-homogeneous policies $\pi_1$ and $\pi_2$ where each $\pi_j(s)$ measures the treatment assignment probability $\prob(A_t=1|S_t=s)$. Note that the integrated value difference function $\tau_0$ can be represented by
\begin{eqnarray*}
	\int_s \{V(\pi_1;s)-V(\pi_2;s)\}\mathbb{G}(ds)=\int_s [\{Q(\pi_1;1,s)-Q(\pi_1;0,s)\}\pi_1(s)\\
	-\{Q(\pi_2;1,s)-Q(\pi_2;0,s)\}\pi_2(s)+Q(\pi_1;0,s)-Q(\pi_2;0,s)]\mathbb{G}(ds).
\end{eqnarray*}
The Q-estimators can be similarly computed via temporal difference learning. More specifically, for a given policy $\pi$, let 
\begin{eqnarray}\label{eqn:Qtpi}
	\widehat{Q}_t(\pi;a,s)=\Psi^{\top} (s)\widehat{\bm{\Sigma}}_{\pi}^{-1}(t) \left\{\frac{1}{t}\sum_{j<t} \left(\begin{array}{c}
		\Psi(S_j)A_j Y_j\\
		\Psi(S_j)(1-A_j) Y_j
	\end{array}
	\right)\right\},
\end{eqnarray}
be the Q-estimator given the data $\{(S_j,A_j,Y_j)\}_{j<t}$ where $\widehat{\bm{\Sigma}}_{\pi}(t)=t^{-1}\sum_{j<t} \bm{\Sigma}_j$ where $\bm{\Sigma}_j$ is defined by
\begin{eqnarray*}
	\left[\begin{array}{lr}
		\Psi(S_j)(1-A_j) \{\Psi(S_j)-\gamma \Psi(S_{j+1})(1-\pi(S_{j+1})) \}^\top  & -\gamma \Psi(S_j)(1-A_j)\Psi^\top(S_{j+1})\pi(S_{j+1})\\
		-\gamma \Psi(S_j)A_j\Psi^\top(S_{j+1})\pi(S_{j+1}) &\Psi(S_j)A_j \{\Psi(S_j)-\gamma \Psi(S_{j+1})(1-\pi(S_{j+1})) \}^\top
	\end{array}
	\right].
\end{eqnarray*}
We can plug-in the Q-estimator in \eqref{eqn:Qtpi} to estimate $\tau_0$. The corresponding variance estimator and the resulting test statistic can be similarly derived. A bootstrap procedure can be similarly developed as in Section \ref{secalphaspend} for sequential testing. We omit the details for brevity.}

{\color{black}\subsection{Experimental design}\label{sec:expdesign}
In our real data application, the design of experiment is determined by the company and we are in the position to analyse the data collected based on such a design. It is important and interesting to design experiments to identity the treatment effect efficiently, but it is beyond the scope of the current paper. 

In addition, it is worth mentioning that the 30-minute-interval design is adopted by the company to optimize the performance of the resulting A/B test. 
To elaborate, let us consider a few toy examples and compare the 30-minute-interval design with an alternating-day design where we switch back and forth between the two policies every day.  

\smallskip

\noindent \textbf{Example 3}. Suppose $Y_t=\delta A_t+\varepsilon_t$ for some constant $\delta>0$ and some zero-mean stationary AR(1) process $\{\varepsilon_t\}_{t\ge 0}$. Here, $Y_t$ and $A_t$ denote the collected response and the assigned action in the $t$th 30-minute interval, respectively. 
There is no carryover effects in this example and the difference in average response between the two groups can be used as the treatment effect estimator. Under the alternating-time-interval design, the difference is taken between adjacent observations. This effectively reduces the variance of the resulting estimator. 
Specifically, suppose $A_0=0$ and the number of observations is equal to $T=48 D$ where $D$ denotes the number of days the experiment lasts. The treatment effect estimator takes the following form,
\begin{eqnarray}\label{eqn:ATE1}
	\widehat{\textrm{ATE}}_1=\frac{2}{T}\sum_{t=0}^{T-1} (-1)^{t+1}R_t.
\end{eqnarray}
Its asymptotic variance of $\widehat{\textrm{ATE}}_1$ equals
\begin{eqnarray*}
	\lim_{T\to\infty}\Var(\sqrt{T}\widehat{\textrm{ATE}}_1)=\lim_{T\to \infty} \frac{4}{T} \{T-2\rho (T-1)+2\rho^2 (T-2)-\cdots\}\\=4 \sum_{t\ge 0} (-\rho)^t-4 \sum_{t\ge 1} (-\rho)^t=\frac{4-4\rho}{1+\rho},
\end{eqnarray*}
where $\rho$ denotes the autocorrelation coefficient. 

Under the alternating day design, the estimator takes the following form,
\begin{eqnarray}\label{eqn:ATE2}
	\widehat{\textrm{ATE}}_2=\frac{2}{T}\sum_{d=1}^{D} \sum_{i=1}^{48} (-1)^d R_{48(d-1)+i-1}.
\end{eqnarray}
It asymptotic variance can be approximated by
\begin{eqnarray*}
	\lim_{T\to\infty}\Var(\sqrt{T}\widehat{\textrm{ATE}}_2)\approx \frac{4}{T}D \Var\left(\sum_{i=1}^{48} R_{i-1}\right)=\frac{4D}{T} (48+2\rho \times 47+2\rho^2 46+\cdots)\\\approx \frac{4D\times 48}{T} \frac{1+\rho}{1-\rho}=\frac{4+4\rho}{1-\rho}. 
\end{eqnarray*}
Based on the above calculation, when $\rho>0$, the asymptotic variance of $\widehat{\textrm{ATE}}_2$ could be much larger than that of $\widehat{\textrm{ATE}}_1$. For instance, when $\rho=0.5$, $\Var(\sqrt{T}\widehat{\textrm{ATE}}_2)$ is approximately 9 times as large as that of $\Var(\sqrt{T}\widehat{\textrm{ATE}}_1)$.

\noindent \textbf{Example 4}: Suppose $Y_{d,t}=\delta A_{d,t}+\eta_d+\varepsilon_{d,t}$ for $1\le d\le D$ and $0\le t<48$ where $\{\varepsilon_{d,t}\}_{d,t}$ are i.i.d. measurement errors and $\{\eta_d\}_d$ are i.i.d. random effects that vary across days. According to \eqref{eqn:ATE2}, the variance of the treatment effect estimator under the alternating-day design depends on that of the random effect, whose accurate estimation is very challenging in cases where $d$ is small (e.g., 14). This would inflat the Type-I error of the resulting test. On the contrary, according to \eqref{eqn:ATE1}, the variance of the treatment effect estimator under the alternating-time-interval design relies only on that of the measurement error, as the random effect cancels each other. 

We remark that in the above two examples, we focus on settings without carryover effects to better illustrate the advantage of the 30-minute-interval design. In cases where the carryover effects exist and RL methods are applied to A/B testing, it would be appropriate to adopt such a design as well, so as to ensure the resulting test has good size and power properties. 

Finally, under the current design, the interim analyses are conducted at the end of the first week as well as the end of each day during the second week. In general, there is a trade-off between the number of interim stages and the power of the test. In particular, the more interim stages, the more likely the test can detect the alternative early. However, it will lead to a less powerful test, which is the price we pay for early termination. This is consistent with findings in classical sequential analysis \citep{jennison1999}.

{\color{black}
\subsection{Off-policy evaluation}\label{sec:opelatent}
In this paper, we focus on causal effects evaluation in online experiments where the treatment generating mechanism is pre-determined. Under these settings, there are no unmeasured confounders that confound the action-outcome or the action-next state relationship. Another equally important problem is study off-policy evaluation in our application. Unmeasured confounding is a serious issue in the observational dataset. This is because the behavior policy usually involves human interventions to balance supply and demand when severe weather or some large live events occur. However, live events and extreme weather are not recorded, leading to a confounded dataset. In the RL literature, a few methods have been developed to handle latent confounders \citep[see e.g.,][]{namkoong2020off,wang2020provably,bennett2021off,liao2021instrumental}. 
Some of these methods can be potentially applied to our setting for policy evaluation. For instance, suppose there exists some auxiliary variables in our application that mediate the treatment effect conditionally independent of the unmeasured confounders given the treatments. Then we can apply the front-door adjustment formula to consistently infer the target policy's value. A similar idea is proposed in \cite{wang2020provably} for policy optimization. To summarize, it is practically interesting to investigate the off-policy evaluation problem in our application. However, this is beyond the scope of the current paper. We leave it for future research. 
}

\appendix

\baselineskip=19pt
\bibliographystyle{apalike}
\bibliography{CausalRL}

\newpage
\section{More on simulations}\label{secaddsimu}
We first report the rejection probabilities of the proposed test with different choices of the discounted factor in Figure \ref{figS3}. We fix the number of basis functions $J=4$ and the $\alpha$-spending function $\alpha(\cdot)=\alpha_1(\cdot)$. It can be seen that the proposed test controls the type-I error in most cases. Its power increases with the discounted factor $\gamma$. 
\begin{figure}[!t]
	\centering
	\begin{tabular}{ccc}
		\hspace{-0.5cm}\includegraphics[scale=0.4]{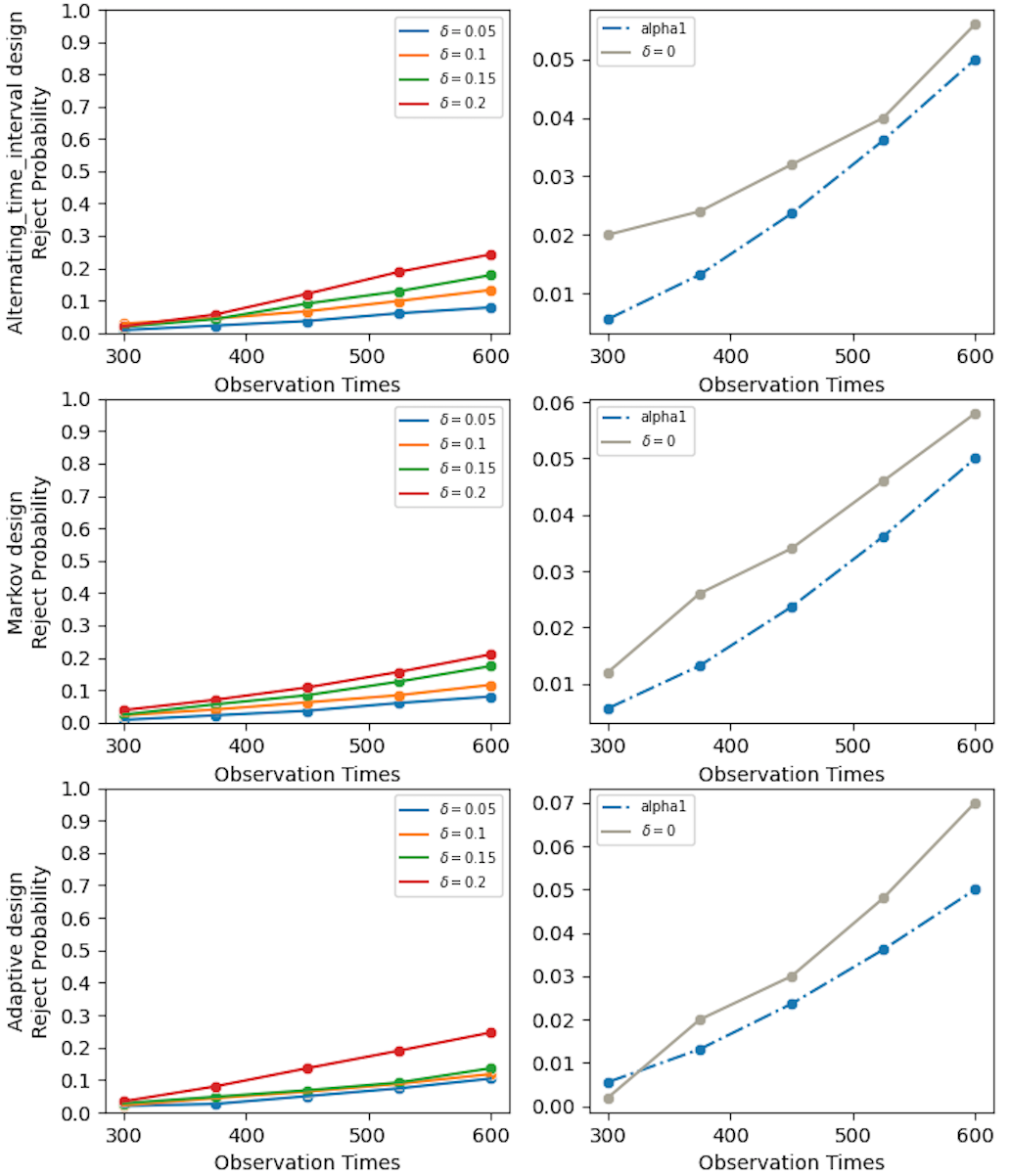} &
		&
		\hspace{-0.5cm}\includegraphics[scale=0.4]{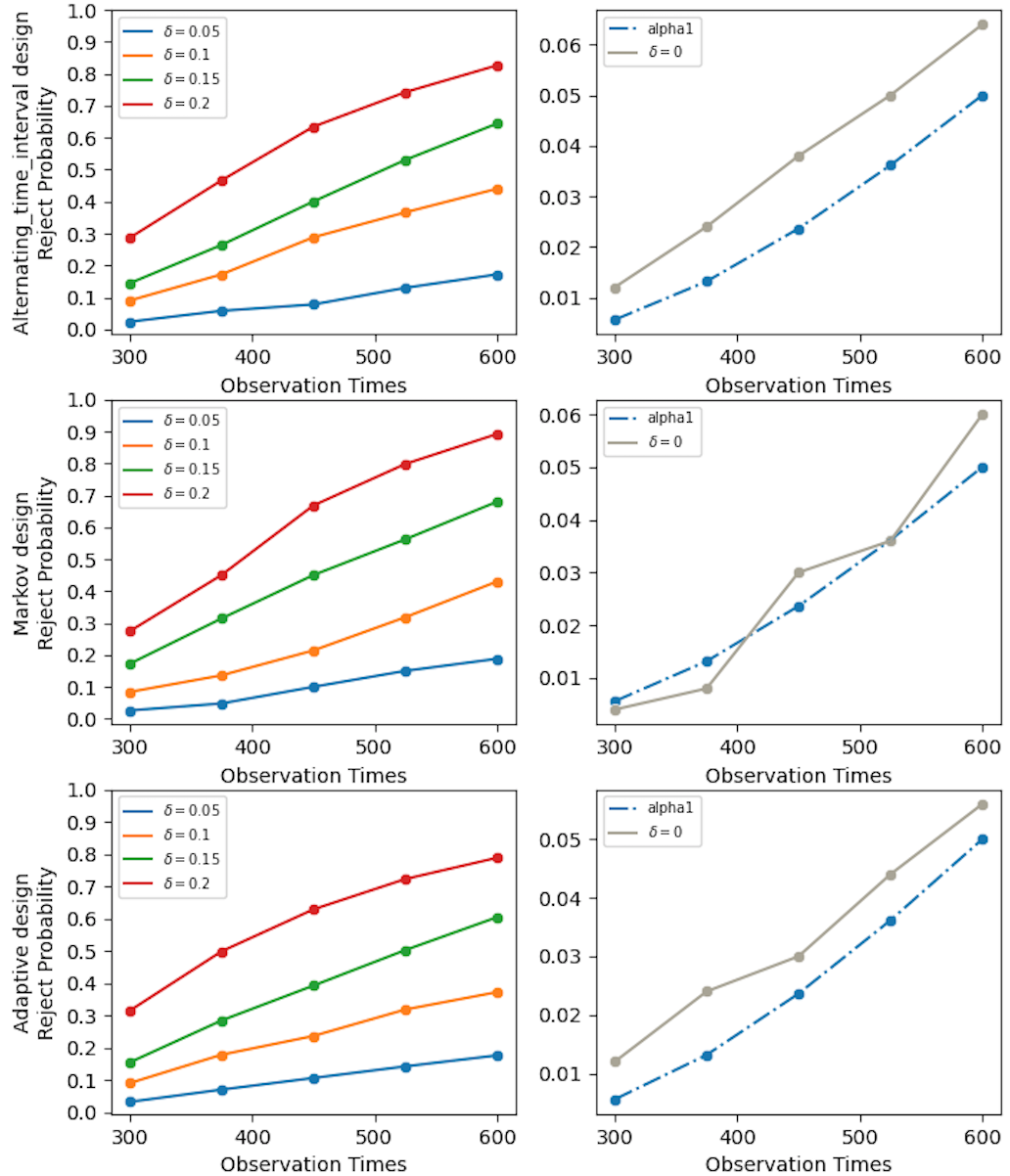} \\
		\small (a) The proposed test under $H_1$ and $H_0$ (from & & \small (b) The proposed test under $H_1$ and $H_0$ (from\\
		\small  left plots to right plots). $\gamma=0.1$. & & \small left plots to right plots). $\gamma=0.3$.  \\
		\hspace{-0.5cm}\includegraphics[scale=0.4]{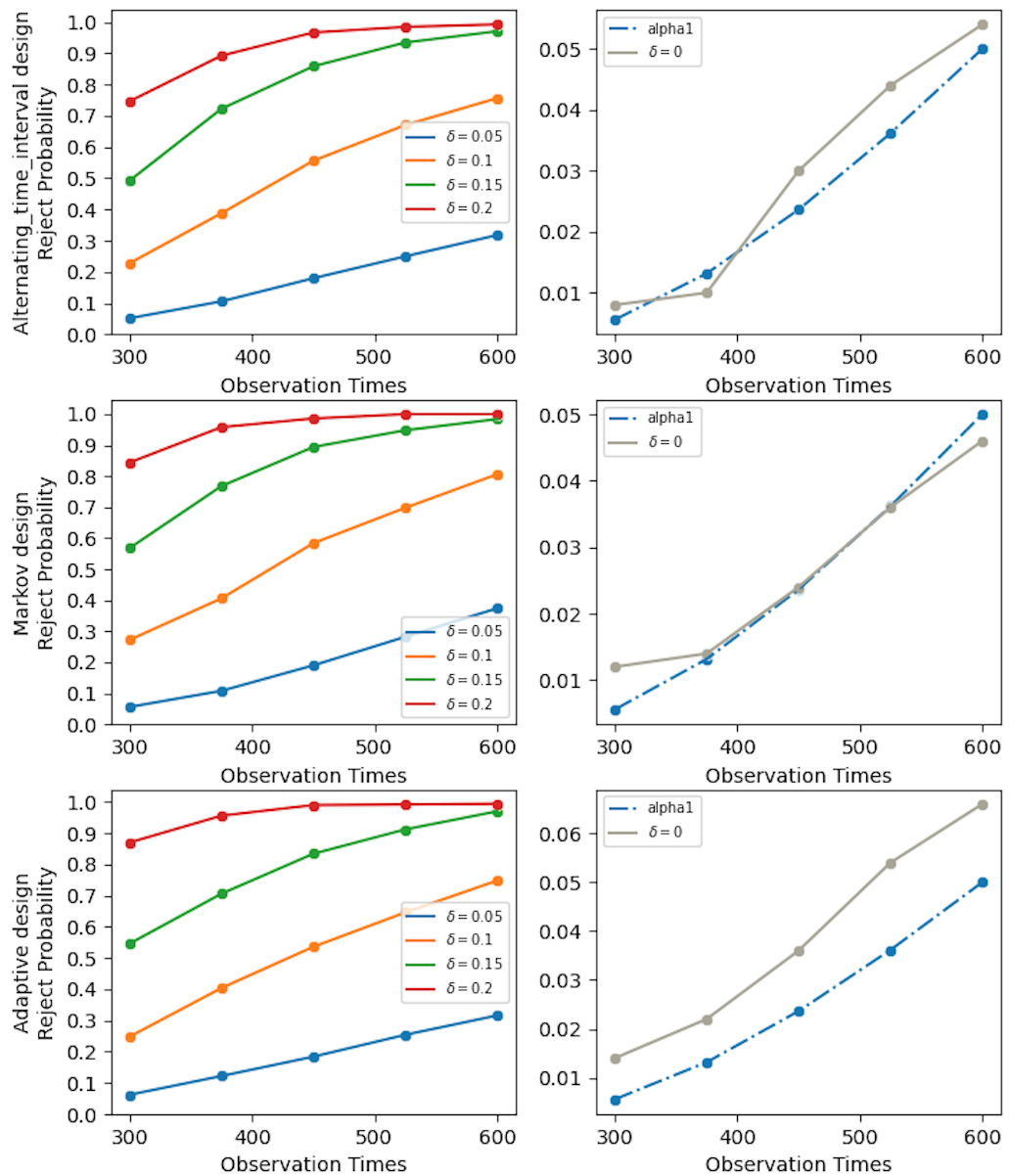} & & 	\hspace{-0.5cm}\includegraphics[scale=0.4]{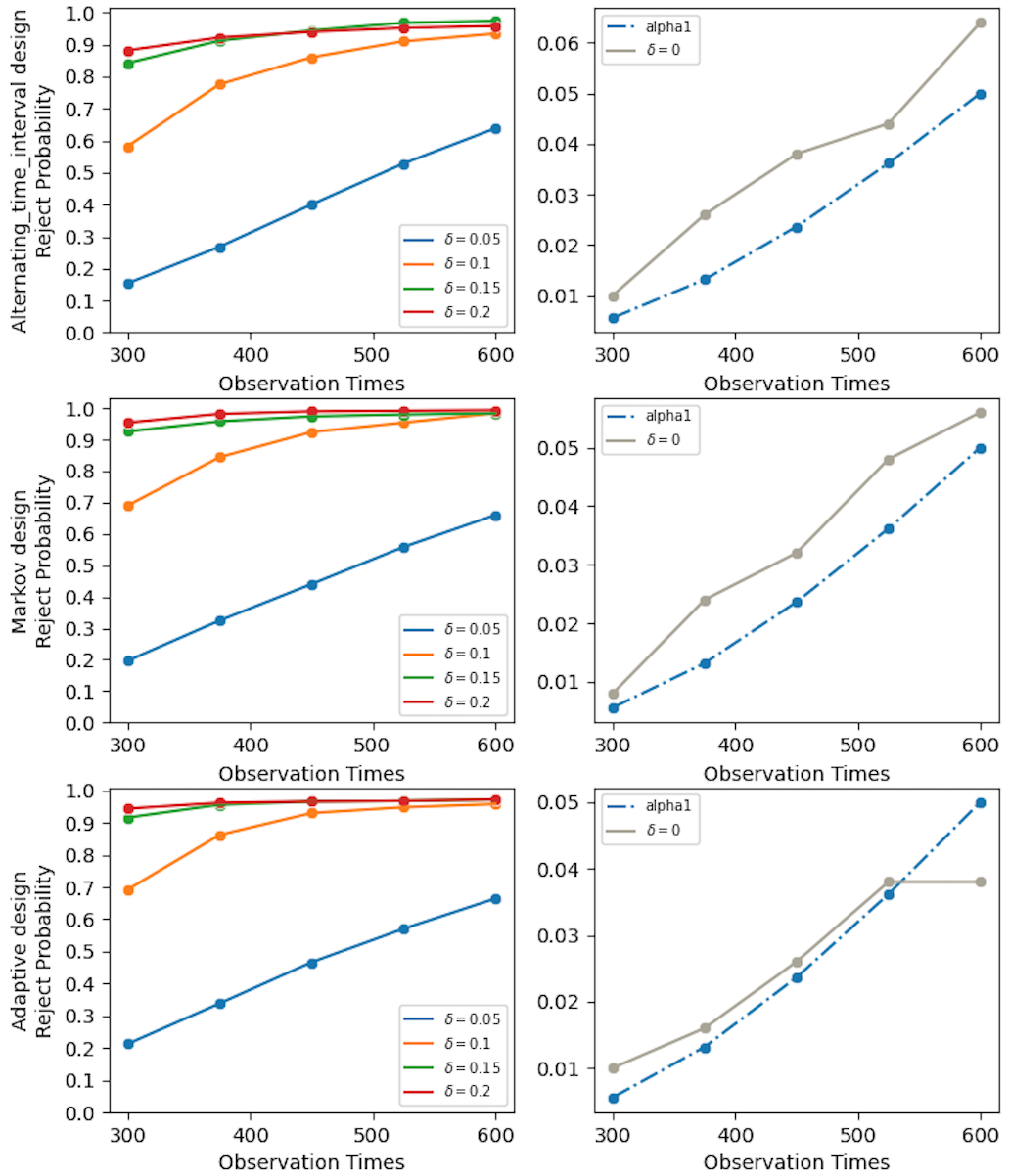} \\
		\small (c) The proposed test under $H_1$ and $H_0$ (from & & (d) The proposed test under $H_1$ and $H_0$ (from\\
		\small left plots to right plots). $\gamma=0.5$. & &  left plots to right plots). $\gamma=0.9$. 
	\end{tabular}
	\caption{Empirical rejection probabilities of our test with $J=4$ and $\alpha(\cdot)=\alpha_1(\cdot)$. Settings correspond to the alternating-time-interval, adaptive and Markov design, from top plots to bottom plots.}
	\label{figS3}
\end{figure}

We next propose a t-test for the carryover effect based on the analysis of $2\times 2$ crossover trials \citep[see e.g.,][Chapter 2]{jones1989design}. The main idea is to divide the entire experiment into a sequence of non-overlapping blocks with equal size. We require the number of blocks to be divisible by 2. Let $2n$ denote the number of blocks. We randomly allocate them with equal probability on the $(2j-1)$th block for any $j=1,2,\cdots,n$. If the $(2j-1)$th block receives one treatment, then the $2j$th block will receive the other treatment. Let $\bar{Y}_j$ denote the average response of the $j$th block. Let $\bar{A}_j=1$ if the $j$th block receives the new treatment and $\bar{A}_j=0$ otherwise.  We propose to estimate the carryover effect by $\widehat{\tau}=(\widehat{\tau}_{1,0}-\widehat{\tau}_{0,1})/2$ where
\begin{eqnarray*}
	\widehat{\tau}_{1,0}=\frac{\sum_{j=1}^{n} \bar{A}_{2j}(\bar{Y}_{2j}-\bar{Y}_{2j-1})}{\sum_{j=1}^n \bar{A}_{2j}}\,\,\hbox{and}\,\,\widehat{\tau}_{1,0}=\frac{\sum_{j=1}^{n} (1-\bar{A}_{2j})(\bar{Y}_{2j}-\bar{Y}_{2j-1})}{n-\sum_{j=1}^n \bar{A}_{2j}}.
\end{eqnarray*}  
Its standard error can be estimated by $\widehat{\sigma}^2m/4$ where $m=1/(\sum_{j} \bar{A}_{2j})+1/(n-\sum_{j} \bar{A}_{2j})$ and
\begin{eqnarray*}
	\widehat{\sigma}^2=\frac{\sum_{j=1}^n \{\bar{A}_{2j} (\bar{Y}_{2j}-\bar{Y}_{2j-1}-\widehat{\tau}_{1,0})^2+(1-\bar{A}_{2j}) (\bar{Y}_{2j}-\bar{Y}_{2j-1}-\widehat{\tau}_{0,1})^2\} }{2n-2}.
\end{eqnarray*}
We reject the null hypothesis when $\widehat{\tau}/\sqrt{\widehat{\sigma}^2 m/4}$ exceeds the upper $\alpha$th quantile of a t-distribution with 2n-2 degrees of freedom. 
\section{Some more discussions}\label{app:moredis}
\subsection{Extensions to high-dimensional models}\label{secexthighd}
In this section, we extend our proposal to settings where the dimension of the state is allowed to diverge with the sample size. We recommend to include a rich class of basis functions to ensure that the Q-function can be well-approximated. Specifically, we assume $Q(a;a,s)=\Psi^\top(s)\bm{\beta}_a^*$ for any $a$ and $s$ where the dimension $q$ is allowed to be much larger than $T$. In this case, the matrix $\widehat{\bm{\Sigma}}(t)$ might not be invertible and the estimator $\widehat{\bm{\beta}}(t)$ cannot be directly obtained by solving the Bellman equation, as in Section \ref{sectest}. 

To handle high-dimensionality, we first adopt the Dantzig selector \citep{candes2007dantzig} to compute an initial estimator $\widetilde{\bm{\beta}}(t)$, which directly penalizes the Bellman equation. We next develop a decorrelated estimator $\widehat{\bm{\beta}}(t)$ by debiasing the initial estimator. This decorrelated estimation step is to reduce the bias of $\widetilde{\bm{\beta}}(t)$. It ensures the entry of $\widehat{\bm{\beta}}(t)$, is $\sqrt{t}$-consistent and asymptotically normal. 

Specifically, for any $t$, we propose to compute $\widetilde{\bm{\beta}}(t)=\{\widetilde{\bm{\beta}}_0(t)^\top, \widetilde{\bm{\beta}}_1(t)^\top \}^\top$ by solving 
\begin{eqnarray*}
	\widetilde{\bm{\beta}}(t)=\argmin_{\bm{\beta}(t)\in \Lambda(t)} \|\bm{\beta}(t)\|_1,
\end{eqnarray*}
where
\begin{eqnarray*}
	\Lambda(t)=\left\{ \bm{\beta}(t):\left\|\widehat{\bm{\Sigma}}(t)\bm{\beta}(t)-\widehat{\bm{\eta}}(t)\right\|_{\infty} \le \lambda(t) \right\},
\end{eqnarray*}
for a sequence of tuning parameters $\{\lambda(t)\}_t$ such that $\lambda(t)\to 0$ as $t\to \infty$. 

Next, for simplicity, suppose $\{S_t\}_t$ has a limiting distribution $\Pi$. For any $1\le j\le q$, we observe that $\beta_{a,j}^*$, the $j$th element of $\bm{\beta}_a^*$, satisfies the following equation:
\begin{eqnarray}\label{eqn:doublerobust}
	\begin{split}
		\Mean_{S\sim \Pi} \mathbb{I}(A=a)\{\Psi_j(S)-\bm{\theta}_j^{*\top} \Psi_{-j}(S)\}\{Y+\gamma \Psi_j(S') \beta_{a,j}^*+ \gamma  \Psi_{-j}(S')^\top \bm{\beta}_{a,-j}^*\\-\Psi_j(S) \beta_{a,j}^*- \Psi_{-j}(S)^\top \bm{\beta}_{a,-j}^*\}=0,
	\end{split}	
\end{eqnarray}
for a state-action-outcome-next state tuple $(S,A,Y,S')$, 
where $\Psi_j(s)$ denotes the $j$th element of $\Psi(s)$, $\bm{\beta}_{a,-j}^*$ and $\Psi_{-j}(s)$ denote the subvector of $\bm{\beta}_{a}^*$ and $\Psi(s)$ obtained by removing their $j$th element, respectively, and 
\begin{eqnarray*}
	\bm{\theta}_{a,j}^*=[\Mean_{S\sim \Pi} \mathbb{I}(A=a) \Psi_{-j}(S)\{\Psi_{-j}(S)-\gamma \Psi_{-j}(S')\}^\top]^{-1} \Mean_{S\sim \Pi} \mathbb{I}(A=a)\Psi_{-j}(S) \Psi_j(S).
\end{eqnarray*}
We remark that Equation \eqref{eqn:doublerobust} is doubly-robust. It holds as long as either $\bm{\theta}_{a,j}^{*\top}$ or $\bm{\beta}_{a,-j}^*$ is correctly specified. 

To estimate $\beta_{a,j}^*$, we compute another Dantzig-type estimator $\widehat{\bm{\theta}}_{a,j}(t)$ of $\bm{\theta}_{a,j}^*$,
\begin{eqnarray*}
	\widehat{\bm{\theta}}_{a,j}(t)=\argmin_{\bm{\theta}(t)\in \Lambda_a^*(t)} \|\bm{\theta}(t)\|_1,
\end{eqnarray*}
where
\begin{eqnarray*}
	\Lambda_a^*(t)=\left\{ \bm{\theta}(t):\frac{1}{t}\left\|\sum_{k<t}\Psi_{-j}(S_k)\mathbb{I}(A_k=a)[\{\Psi_{-j}(S_k)-\gamma \Psi_{-j}(S_{k+1})\}^\top \bm{\theta}(t)-\Psi_j(S_k)]\right\|_{\infty} \le \lambda_{a,j}^*(t) \right\},
\end{eqnarray*}
for a sequence of tuning parameters $\{\lambda_{a,j}^*(t)\}_t$ such that $\lambda_{a,j}^*(t)\to 0$ as $t\to \infty$. 

Finally, based on \eqref{eqn:doublerobust}, we compute $\widehat{\beta}_{a,j}$ by construct the following estimating equation,
\begin{eqnarray*}
	\sum_{k<t} \mathbb{I}(A_k=a)\{\Psi_j(S_k)-\widehat{\bm{\theta}}_{a,j}(t)^\top \Psi_{-j}(S_k)\}\{Y_k+\gamma \Psi_j(S_{k+1}) \widehat{\beta}_{a,j}(t)+ \gamma  \Psi_{-j}(S_{k+1})^\top \widetilde{\bm{\beta}}_{a,-j}^*(t)\\-\Psi_j(S_k) \widehat{\beta}_{a,j}(t)-  \Psi_{-j}(S_k)^\top \widetilde{\bm{\beta}}_{a,-j}^*(t)\}=0,
\end{eqnarray*}
where $\widetilde{\bm{\beta}}_{a,-j}^*(t)$ denotes the subvector of the initial estimator $\widetilde{\bm{\beta}}_{a}(t)$ obtained by removing its $j$th element. 

The doubly-robustness property ensures the asymptotic normality of $\widehat{\beta}_{a,j}(t)$ in cases when neither the initial estimator nor $\widehat{\bm{\theta}}_{a,j}(t)$ converges at a parametric rate \citep[see e.g., Theorem 1,][]{shi2021testing}. Specifically, under certain mild conditions, when the tuning parameters $\lambda(t)$ and $\lambda_{a,j}^*(t)$ are set to be $C t^{-1/2} \sqrt{\log q}$ for some sufficiently large constant $C>0$, we can show that $\widetilde{\bm{\beta}}(t)$ and $\widehat{\bm{\theta}}_{a,j}(t)$ converge at a rate of $t^{-1/2}\sqrt{s\log q}$ and $t^{-1/2}\sqrt{s_{a,j}\log q}$, respectively, where $s$ and $s_{a,j}$ denotes the number of nonzero elements in $\bm{\beta}^*$ and $\bm{\theta}_{a,j}^*$ \citep[see e.g., Theorem 6.1,][]{shi2018high}. Then using similar arguments in the proof of Theorem 1 of \cite{shi2021testing}, we can show that $\widehat{\beta}_{a,j}(t)$ has a tractable limiting distribution when $\max(s,s_{a,j})\log q=o(\sqrt{t})$. 

Let $\widehat{\bm{\beta}}_{a}(t)$ denote the estimator for $\bm{\beta}_a^*$ based on $\{\widehat{\beta}_{a,j}(t)\}_j$. Its asymptotic variance can be consistently estimated by $\widehat{\bm{\Theta}}_a(t)$ whose $(j_1,j_2)$th element is given by
\begin{eqnarray*}
	\left[\sum_{k<t} \mathbb{I}(A_k=a) \{\Psi_{j_1}(S_k)-\gamma \Psi_{j_1}(S_{k+1})\} \{\Psi_{j_1}(S_k)-\widehat{\bm{\theta}}_{a,j_1}(t)^\top \Psi_{-j_1}(S_k)\}\right]^{-1}\\
	\times \left[\sum_{k<t} \mathbb{I}(A_k=a) \{\Psi_{j_2}(S_k)-\gamma \Psi_{j_2}(S_{k+1})\} \{\Psi_{j_2}(S_k)-\widehat{\bm{\theta}}_{a,j_2}(t)^\top \Psi_{-j_2}(S_k)\}\right]^{-1}\\
	\times \left[\sum_{k<t}\mathbb{I}(A_k=a) \{\Psi_{j_1}(S_k)-\widehat{\bm{\theta}}_{a,j_1}(t)^\top \Psi_{-j_1}(S_k)\} \{\Psi_{j_2}(S_k)-\widehat{\bm{\theta}}_{a,j_2}(t)^\top \Psi_{-j_2}(S_k)\} \widehat{\varepsilon}_{k,a}^2\right],
\end{eqnarray*}
where $\widehat{\varepsilon}_{k,a}$ corresponds to the estimated Bellman residual $Y_k+\gamma \Psi(S_{k+1})^\top \widetilde{\bm{\beta}}_a(t)-\Psi(S_k)^\top\widetilde{\bm{\beta}}_a(t)$. Let $\widehat{\bm{\Theta}}(t)=\diag[\widehat{\bm{\Theta}}_0(t),\widehat{\bm{\Theta}}_1(t)]$. At time $t$, we reject the null if $U^\top \widehat{\bm{\beta}}(t)>  z_{\alpha} U^\top \widehat{\bm{\Theta}}(t) U$. A bootstrap-assisted procedure can be similarly developed for sequential monitoring, as in Section \ref{secalphaspend}. We omit the details for brevity. 

\subsection{Comparison with \citet{luckett2019}}\label{sec:disVlearning}
Below, we summarize the methodological difference between our proposal and the V-learning method. First, we remark that V-learning focuses on the problem of policy optimization. That is, how to learn an optimal policy based on the observed dataset. This problem is different from policy evaluation, which is the focus of our paper. For a given randomized policy, \cite{luckett2019} outlined a procedure to learn its value in Section 2. In Theorem 4.2 of \cite{luckett2019}, they proved the asymptotic normality of the value estimator. In Theorem 4.3 of \cite{luckett2019}, they provided a consistent variance estimator when the policy being considered is an estimated optimal policy. Based on these arguments, one can develop a procedure to infer the value difference between two policies. However, its numerical performance remains unknown to us, as no numerical results were available in \cite{luckett2019} that report the coverage probability of the resulting confidence interval.

Second neither sequential monitoring nor online updating (update the test without storing the historical data) is being considered in \cite{luckett2019}. To the contrary, the focus of our paper is to design a test procedure that allows for sequential monitoring and online updating, as motivated by our application. 

Third, the proposed test relaxes the positivity assumption that is needed in the V-learning paper. This makes the proposed test applicable to the  alternating-time-interval design, which is the experimental design adopted by the company and other ride-hailing companies. In contrast, V-learning is not applicable to such a design. In addition, due to the use of inverse propensity score weighting, V-learning has inflated type-I errors and is less powerful than our test under some settings, as reflected in our simulation studies. 

Fourth, in Appendix \ref{secexthighd}, we further extend our proposal to handle high-dimensional models, develop a decorrelated estimator to reduce the bias of the initial Q-estimator and outline the corresponding test statistic. In \cite{luckett2019}, they used a different penalized objective function. The validity of their method requires the tuning parameter to decay to zero at a rate that is faster than the usual parametric rate. When this condition is violated (e.g., when cross-validation is applied for tuning parameter selection), their estimator cannot be directly applied to construct the test statistic due to its large bias resulting from the use of penalty functions \citep[see e.g.,][]{shi2019linear}. Our proposal relaxes this constraint by adopting the debiasing idea that is commonly used for statistical inference of low-dimensional parameters in high-dimensional generalized linear models \citep{ning2017general}.

Finally, we also remark that in theory, \cite{luckett2019} does not consider increasing the model complexity (e.g., the number of basis functions in the sieve estimator). A fundamental question studied in our paper is that how many number of basis functions shall the we choose. A naive solution is to adopt a sufficiently rich set of basis functions such that the Q-function could be well-approximated at a rate that is faster than the parametric rate. In other words, ``undersmoothing" is employed to guarantee the value estimator has a tractable limiting distribution. Results in our paper suggest that ``undersmoothing" is not required and the value estimator is asymptotically unbiased even when the bias of the Q-estimator decays at a rate that is slower than the parametric rate. 


\subsection{Comparison with the literature on crossover trials}\label{sec:crossover}
In crossover trials, each experimental unit receives a sequence of experimental treatments 
\citep[see e.g.,][for an overview]{jones1989design}. Unlike classical two-sample t-test, the resulting test is able to detect the carryover effect in time and can be potentially used in A/B testing for comparing the long-term treatment effect. The proposed test differs from these tests in that it imposes certain structural assumptions (e.g., Markovianity) on the time series data to identify the carryover effect. These assumptions allow our test to be applicable to a large variety of experimental designs. They also enable us to estimate the carryover effect more efficient, resulting in tests with better power properties, as shown in Section \ref{sec:compare}. 
}

\section{Potential outcomes under a random policy}
\label{secporandom}
We focus on the class of Markov policies that are functions of current state variables only. Following \cite{luckett2019}, we define $\{\xi_{b}^t(\cdot)\}_{t\ge 0}$ as a sequence of independent, binary-valued stochastic processes that satisfy $\prob\{\xi_{b}^t(s)=a\}=b_t(a|s)$  for any $t\ge 0$, $s\in \mathbb{S}$ and $a\in \{0,1\}$, where $b_t$ denotes the policy used at time $t$. The potential outcomes $Y_t^*(\bar{b}_t)$ and $S_{t+1}^*(\bar{b}_t)$ can thus be recursively defined as
\begin{eqnarray*}
Y_t^*(\bar{b}_t)&=&\sum_{\bar{a}_t\in \{0,1\}^{t+1}} Y_t^*(\bar{a}_t) \mathbb{I}(\xi_b^t(S_t^*(\bar{a}_{t-1}))=\bar{a}_t),\\
S_{t+1}^*(\bar{b}_t)&=&\sum_{\bar{a}_t\in \{0,1\}^{t+1}} S_{t+1}^*(\bar{a}_t) \mathbb{I}(\xi_b^t(S_t^*(\bar{a}_{t-1}))=\bar{a}_t),
\end{eqnarray*} 
for $t=0,1,\cdots$.

\section{Technical conditions}\label{sectechcond}
To simplify the presentation, we assume all state variables are continuous. 

\subsection{Condition C1}\label{subsecC1}

C1 Suppose (i) holds. Assume (ii) holds under D1, (iii) holds under D2 and (ii), (iv) hold under D3. \\ 
(i) The transition kernel $\mathcal{P}$ is absolutely continuous and satisfies $\mathcal{P}(ds;a,s')=p(s;a,s')ds$ for some transition density function $p$. In addition, assume $p$ is uniformly bounded away from $0$ and $\infty$. \\
(ii) The Markov chain $\{S_{t}^*(\bar{b}_{t-1})\}_{t\ge 0}$ formed under the behaviour policy is geometrically ergodic, i.e., there exists some function $M$ on $\mathbb{S}$, some constant $0\le \rho<1$ and some probability density function $\Pi$ such that $\int_{s\in \mathbb{S}} M(s)\Pi(ds)<+\infty$ and
\begin{eqnarray*}
\left\|\prob(S_{t}^*(\bar{b}_{t-1})\in \mathcal{S}|S_0=s)-\Pi(\mathcal{S})\right\|_{\textrm{TV}}\le M(s)\rho^t, \,\,\,\,\,\,\,\,\forall t\ge 0,s \in \mathbb{S},\mathcal{S}\subseteq \mathbb{S},
\end{eqnarray*}
where $\|\cdot\|_{\textrm{TV}}$ denotes the total variation norm.\\
(iii) The Markov chains $\{S_{2t}^*(\bar{b}_{2t})\}_{t\ge 0}$ and $\{S_{2t+1}^*(\bar{b}_{2t+1})\}_{t\ge 0}$ are geometrically ergodic.\\
(iv) For any $k=1,\cdots,K-1$, the following events occur with probability tending to $1$: the Markov chain $\{S_t^*(\bar{b}_{t-1}^{(k)})\}_{t\ge 0}$ is geometrically ergodic; $\sup_{a\in \{0,1\},s\in \mathbb{S}} |b^{(k)}(a|s)-b^*(a|s)|\stackrel{P}{\to} 0$ for some $b^*(\cdot)$; the stationary distribution of $\{S_t^*(\bar{b}_{t-1}^{(k)})\}_{t\ge 0}$ will converge to some $\Pi^*$ in total variation. 

\textbf{Remark}: By C1(ii), $\Pi$ is the stationary distribution of $\{S_t^*(\bar{b}_{t-1})\}_{t\ge 0}$. It follows that $$\Pi(\mathcal{S})=\sum_{a\in \{0,1\}}\int_{s\in \mathbb{S}} \mathcal{P}(\mathcal{S};a,s) b(a|s) \Pi(ds),$$ 
for any $\mathcal{S}\subseteq \mathbb{S}$. By C1(i), we obtain
\begin{eqnarray}\label{remarkeq1}
\Pi(\mathcal{S})=\sum_{a\in \{0,1\}}\int_{s\in \mathbb{S}} \int_{s'\in \mathcal{S}} b(a|s) p(s';a,s) ds' \Pi(ds) 
=\int_{s'\in \mathcal{S}} \underbrace{\sum_{a\in \{0,1\}}\int_{s\in \mathbb{S}}b(a|s) p(s';a,s)\Pi(ds)}_{\mu(s')} ds'.
\end{eqnarray}
This implies that $\mu(\cdot)$ is the density function of $\Pi$. Since $p$ is uniformly bounded away from $0$ and $\infty$, so is $\mu$. 

Under C1(iv), for any $k\in \{1,\cdots,K-1\}$, there exist some $M^{(k)}(\cdot)$, $\Pi^{(k)}(\cdot)$ and $\rho^{(k)}$ that satisfy $\int_{s\in \mathbb{S}} M^{(k)}(s)\Pi^{(k)}(ds)<+\infty$ and
\begin{eqnarray}\label{remarkeq2}
\left\|\prob(S_{t}^*(\bar{b}_{t-1}^{(k)})\in \mathcal{S}|S_0=s)-\Pi^{(k)}(\mathcal{S})\right\|_{\textrm{TV}}\le M^{(k)}(s)\{\rho^{(k)}\}^t, \,\,\,\,\,\,\,\,\forall t\ge 0,s \in \mathbb{S},\mathcal{S}\subseteq \mathbb{S},
\end{eqnarray}
with probability tending to $1$. Since $b^{(k)}$ is a function of the observe data history, so are $M^{(k)}(\cdot)$, $\Pi^{(k)}(\cdot)$ and $\rho^{(k)}$.

Suppose an $\epsilon$-greedy policy is used, i.e. $b^{(k)}(a|s)=\epsilon/2+(1-\epsilon) \widehat{\pi}^{(k)}(a|s)$ where $\widehat{\pi}^{(k)}$ denotes some estimated optimal policy. Then the condition $\sup_{a\in\{0,1\},s\in \mathbb{S}} |b^{(k)}(a|s)-b^*(a|s)|\stackrel{P}{\to} 0$ requires $\widehat{\pi}^{(k)}$ to converge. The total variation distance between the one-step transition kernel under $\bar{b}^{(k)}$ and that under $b^*$ can be bounded by
\begin{eqnarray*}
\sup_{s}|\prob(S_1^*(b^{(k)})\in \mathcal{S}|S_0=s)-\prob(S_1^*(b^*)\in \mathcal{S}|S_0=s)|\le \sup_{a,s} |b^{(k)}(a|s)-b^*(a|s)|  \sup_{s,s',a} p(s';a,s),
\end{eqnarray*}
and converges to zero in probability. When the markov chain $\{S_t^*(\bar{b}_{t-1}^{(k)})\}_{t\ge 0}$ is uniformly ergodic, it follows from Theorems 2 and 3 of \cite{Rabta2018} that $\|\Pi^{(k)}-\Pi^*\|_{\textrm{TV}}\to 0$ where $\Pi^*$ corresponds to the stationary distribution of $\{S_t^*(\bar{b}_{t-1}^*)\}$. The last condition in C1(iv) is thus satisfied. 

\subsection{Condition C2}\label{subsecC2}
For any $T_1\le t\le T_K$, we first introduce the marginalized density ratio as follows,
\begin{eqnarray*}
\omega_t(a;A,S)=(1-\gamma)\frac{\mathbb{I}(A=a)\sum_{j\ge 0}\gamma^j p_j(a;S)}{t^{-1}\sum_{j=0}^{t-1} \prob(A_j=A|S_j=S,\{(A_l,S_l)\}_{l<j})p_j(b;S)},
\end{eqnarray*}
where $p_t(a;\cdot)$ denotes the probability density function of $S_t^*(a)$. Under D1 and D2, $p_j(b;\cdot)$ denotes the probability density function of $S_j$. Let $T_0=0$. Under D3, for any $T_k\le j<T_{k+1}$, $p_j(b;\cdot)$ denotes the conditional probability density function of $S_j$ given $\{(A_l,S_l)\}_{l< T_k}$.  
The numerator corresponds to the $\gamma$-discounted visitation probability of the state-action pair assuming the system assigns Treatment $a$ at any time. The numerator corresponds to the distribution function of a randomly sampled state-action pair from the set $\{(S_j,A_j)_{0\le j<t}\}$. Such a marginalized density ratio plays an important role in breaking the curse of horizon for policy evaluation \citep{liu2018,kallus2019efficiently}. We next introduce Condition C2. 

\smallskip

\noindent C2(i) Assume there exist some $\bm{\beta}^*$ and $\{\theta_{a,t}^*\}_{a,t}$ such that
\begin{eqnarray*}
\sup_{\substack{a\in \{0,1\}, s\in \mathbb{S} }} |Q(a;a,s)-\Psi^\top(s) \beta_{a}^*|=o(T^{-1/4}),
\sup_{\substack{a\in \{0,1\}, s\in \mathbb{S}, T_1\le t<T_K }} |\omega_t(a;a,s)-\Psi^\top(s) \theta_{a,t}^*|=o(T^{-1/4}).
\end{eqnarray*}
(ii) Assume there exists some constant $\bar{c}^*\ge 1$ such that
\begin{eqnarray}\label{condanother}
(\bar{c}^*)^{-1}\le \lambda_{\min}\left\{\int_{s\in \mathbb{S}} \Psi(s)\Psi^\top(s) ds\right\}\le \lambda_{\max}\left\{\int_{s\in \mathbb{S}} \Psi(s)\Psi^\top(s)ds\right\}\le \bar{c}^*,
\end{eqnarray}
and $\sup_{s} \|\Psi(s)\|_2=O(\sqrt{q})$. \\
(iii) Assume $\liminf_q \|\int_{s\in \mathbb{S}} \Psi(s)\mathbb{G}(ds)\|_2>0 $. 

\textbf{Remark}: {\color{black}We do not require the approximation error to decay at a rate of $o(T^{-1/2})$, as commented in the main text. }For any $a\in \{0,1\}$, suppose $Q(a;a,s)$ and $\omega_t(a;a,s)$ are $p$-smooth as functions of $s$ \citep[see e.g.][for the definition of $p$-smoothness]{Stone1982}. When tensor product B-splines or wavelet basis functions \citep[see Section 6 of][for an overview of these bases]{Chen2015} are used for $\Psi(\cdot)$, the resulting approximation error will be of the order $O(q^{-p/d})$. 
%
See Section 2.2 of \cite{Huang1998} for details. It follows that Condition C2(i) automatically holds when the number of basis functions $q$ satisfies $q\gg T^{d/(4p)}$. 

Condition C2(ii) is satisfied when tensor product B-splines or wavelet basis is used. For B-spline basis, the assertion in \eqref{condanother} follows from the arguments used in the proof of Theorem 3.3, \cite{Burman1989}. For wavelet basis, the assertion in \eqref{condanother} follows from the arguments used in the proof of Theorem 5.1, \cite{Chen2015}. For both bases, the number of nonzero elements in $\Psi(\cdot)$ is bounded by some constant. Moreover, each basis function is uniformly bounded by $O(\sqrt{q})$. The condition $\sup_{s} \|\Psi(s)\|_2=O(\sqrt{q})$ thus holds. For any $q$-dimensional vector $\nu$ of unit $\ell_2$ norm, we have
\begin{eqnarray*}
\left|\nu^\top \int_{\mathbb{S}}\Psi(s)\mathbb{G}(s)\right|^2= \nu^\top  \left\{\int_{\mathbb{S}}\Psi(s)\mathbb{G}(s)\right\}\left\{\int_{\mathbb{S}}\Psi(s)\mathbb{G}(s)\right\}^\top \nu\le \nu^\top \int_{\mathbb{S}}\Psi(s)\Psi^\top(s)\mathbb{G}(s)\nu=O(1),
\end{eqnarray*}
where the first inequality is due to Cauchy-Schwarz inequality and the last equality is due to \eqref{condanother} and the fact that $\mathbb{G}$ has a bounded density function. This further implies that
\begin{eqnarray}\label{condother}
\left|\int_{\mathbb{S}}\Psi(s)\mathbb{G}(s)\right|_2=\sup_{\nu:\|\nu\|_2=1} \left|\nu^\top \int_{\mathbb{S}}\Psi(s)\mathbb{G}(s)\right|=O(1).
\end{eqnarray}

Condition C2(iii) automatically holds for tensor product B-splines basis. Notice that $\bm{1}^\top \Psi(s)=q^{1/2}$ for any $s\in \mathbb{S}$ where $\bm{1}$ denotes a vector of ones. It follows from Cauchy-Schwarz inequality that
\begin{eqnarray*}
\sqrt{q}\left\|\int_{s\in \mathbb{S}} \Psi(s)\mathbb{G}(ds)\right\|_2\ge  \left\|\int_{s\in \mathbb{S}} \bm{1}^\top \Psi(s)\mathbb{G}(ds)\right\|_2=\sqrt{q}. 
\end{eqnarray*}
C2(iii) is thus satisfied. 

\subsection{Condition C3}
\noindent C3 Assume $\inf_{a\in \{0,1\},s\in \mathbb{S}}\Var\{\varepsilon^*(a)|S_0=s\}>0$ where  $\varepsilon^*(a)=Y_0^*(a)+\gamma Q(a;a, S_1^*(a))-Q(a;a,S_0)$.

\section{Technical proofs}
\subsection{Proof of Lemma \ref{lemma2}}
To prove Lemma \ref{lemma2}, we state the following lemma.
\begin{lemma}\label{lemma1}
Under MA and CMIA, $Q(a';a,s)=r(a,s)+\gamma \int_{s'} Q(a';a',s')\mathcal{P}(ds';a,s)$ for any $(s,a)$.
\end{lemma}
\textit{Proof of Lemma \ref{lemma1}:} For any $a,a'\in \{0,1\}$, define the potential outcome $Y_t^*(a',a)$ and $S_t^*(a',a)$ as the reward and state variables that would occur at time $t$ had the agent assigned Treatment $a$ at the initial time point and Treatment $a'$ afterwards. 

Let $\mathcal{P}_{a'}^t(\mathbb{S},a,s)=\prob\{S_t^*(a',a)\in \mathbb{S}|S_0=s\}$ for any $\mathbb{S}\subseteq \mathbb{S},a,a'\in \{0,1\},s\in \mathbb{S}$ and $t\ge 0$. We break the proof into two parts. In Part 1, we show Lemma \ref{lemma1} holds when the following is satisfied:
\begin{eqnarray}\label{prooflemma1eq2}
\prob\{S_{t+1}^*(a',a)\in \mathbb{S}|S_1^*(a)=s, S_0
\}=\mathcal{P}_{a'}^t(\mathbb{S},a',s),\\\nonumber
\end{eqnarray}
In Part2, we show \eqref{prooflemma1eq2} holds.

\textbf{Part 1:} Under CMIA, we have
\begin{eqnarray}\label{prooflemma1eq0}
\begin{split}
	\Mean \{Y_t^*(a',a)|S_0=s\}=\Mean [\Mean \{Y_t^*(a',a)|S_t^*(a',a), S_0=s\}|S_0=s]\\
	=\Mean \{r(\pi(S_t^*(a',a)),S_t^*(a',a))|S_0=s\}.
\end{split}
\end{eqnarray}
It follows that
\begin{eqnarray}\label{prooflemma1eq1}
Q(a';a,s)= \sum_{t\ge 0} \gamma^t\Mean \{r(\pi(S_t^*(a',a)),S_t^*(a',a))|S_0=s\}.
\end{eqnarray}
Similar to \eqref{prooflemma1eq0}, we can show
\begin{eqnarray*}
\Mean \{Y_{t+1}^*(a',a)|S_0=s\}=\Mean  \{r(\pi(S_{t+1}^*(a',a)),S_{t+1}^*(a',a))|S_0=s\}\\
=\Mean[\Mean  \{r(\pi(S_{t+1}^*(a',a)),S_{t+1}^*(a',a))|S_1^*(a), S_0=s\}|S_0=s],
\end{eqnarray*}
and hence
\begin{eqnarray*}
\sum_{t\ge 0} \gamma^t \Mean \{Y_{t+1}^*(a',a)|S_0=s\}=\Mean\left[\sum_{t\ge 0}\gamma^t\left.\Mean  \{r(\pi(S_{t+1}^*(a',a)),S_{t+1}^*(a',a))\right|S_1^*(a), S_0=s\}|S_0=s\right].
\end{eqnarray*}
By \eqref{prooflemma1eq2}, the conditional distribution of 
$S_{t+1}^*(a',a)$ given $S_1^*(a)=s$ and $S_0$ are the same as the conditional distribution of $S_t^*(a',a)$ given $S_0=s$. It follows that from \eqref{prooflemma1eq1} that
\begin{eqnarray*}
\sum_{t\ge 0} \gamma^t \Mean \{Y_{t+1}^*(a',a)|S_0=s\}=\Mean \{Q(a';a,S_1^*(a))|S_0=s\}.
\end{eqnarray*}
This together with the definition of Q function and CMIA yields
\begin{eqnarray}\label{prooflemma1eq1.5}
Q(a';a,s)=r(a,s)+\gamma \left[\sum_{t\ge 0} \gamma^t \Mean \{Y_{t+1}^*(a',a)|S_0=s\}\right]=r(a,s)+\gamma \Mean \{Q(a';a,S_1^*(a))|S_0=s\}.
\end{eqnarray}
Under MA, we have
\begin{eqnarray*}
\Mean \{Q(a';a,S_1^*(a))|S_0=s\}=\int_{s'\in \mathbb{S}} Q(a';a,s') \mathcal{P}(ds';a,s).
\end{eqnarray*}
Combining this together with \eqref{prooflemma1eq1.5} yields the desired result. 

\textbf{Part 2:} We use induction to prove \eqref{prooflemma1eq2}. When $t=0$, it trivially holds. 

Suppose \eqref{prooflemma1eq2} holds for $t=k$. In the following, we show \eqref{prooflemma1eq2} holds for $t=k+1$. Under MA, we have
\begin{eqnarray*}
\prob\{S_{k+2}^*(a',a)\in \mathbb{S}|S_1^*(a)=s, S_0
\}=\Mean [\prob\{S_{k+2}^*(a',a)\in \mathbb{S}|S_{k+1}^*(a',a),S_1^*(a)=s, S_0
\}|S_1^*(a)=s, S_0]\\
=\Mean [\mathcal{P}(\mathbb{S};a',S_{k+1}^*(a',a))|S_1^*(a)=s, S_0].
\end{eqnarray*}
Since we have shown \eqref{prooflemma1eq2} holds for $t=k$, it follows that
\begin{eqnarray*}
\prob\{S_{k+2}^*(a',a)\in \mathbb{S}|S_1^*(a)=s, S_0
\}=\int_{s'\in \mathbb{S}} \mathcal{P}(\mathbb{S};a',s') \mathcal{P}_{a'}^{k}(ds',a',s).
\end{eqnarray*}
Similarly, we can show
\begin{eqnarray*}
\mathcal{P}_{a'}^{k+1}(\mathbb{S},a',s)=\prob\{S_{k+1}^*(a',a')\in \mathbb{S}|S_0=s
\}=\int_{s'\in \mathbb{S}} \mathcal{P}(\mathbb{S};a',s') \mathcal{P}_{a'}^{k}(ds',a',s).
\end{eqnarray*}
The proof is hence completed. 

\textit{Proof of Lemma \ref{lemma2}}: 
By CA, it is equivalent to show
\begin{eqnarray*}
\Mean \{Q(a';A_t,S_t^*(\bar{A}_{t-1}))-Y_t^*(\bar{A}_t)-\gamma Q(a';a',S_{t+1}^*(\bar{A}_t))\}\varphi(A_t,S_t^*(\bar{A}_{t-1}))=0.
\end{eqnarray*}
Let $\mathbb{S}_{0}$ denote the support of $S_0$. For any $s_0\in \mathbb{S}_0$, it suffices to show
\begin{eqnarray*}
\Mean \{Q(a';A_t,S_t^*(\bar{A}_{t-1}))-Y_t^*(\bar{A}_t)-\gamma Q(a';a',S_{t+1}^*(\bar{A}_t))\varphi(A_t,S_t^*(\bar{A}_{t-1})) |S_0=s_0\}=0.
\end{eqnarray*}
This is equivalent to show
\begin{eqnarray*}
\Mean \{Q(a';A_t,S_t^*(\bar{A}_{t-1}))-Y_t^*(\bar{A}_t)-\gamma Q(a';a',S_{t+1}^*(\bar{A}_t))\varphi(A_t,S_t^*(\bar{A}_{t-1}))\mathbb{I}(A_0=a_0)\} |S_0=s_0]=0,
\end{eqnarray*}
for any $s_0\in \mathbb{S}_0$, $a_0\in \{0,1\}$. 

Let $\mathcal{A}_0(s_0)=\{ a\in \{0,1\}: \hbox{Pr}(A_0=a|S_0=s)>0\}$. It suffices to show for any $s_0\in \mathbb{S}_0,a_0\in \mathcal{A}_0(s_0)$,
\begin{eqnarray*}
\Mean \{Q(a';A_t,S_t^*(\bar{A}_{t-1}))-Y_t^*(\bar{A}_t)-\gamma Q(a';a',S_{t+1}^*(\bar{A}_t))\varphi(A_t,S_t^*(\bar{A}_{t-1}))\mathbb{I}(A_0=a_0) |S_0=s_0\}=0,
\end{eqnarray*}
or equivalently,
\begin{eqnarray}\label{prooflemma2eq1}
\Mean \{Q(a';A_t,S_t^*(\bar{A}_{t-1}))-Y_t^*(\bar{A}_t)-\gamma Q(a';a',S_{t+1}^*(\bar{A}_t))\varphi(A_t,S_t^*(\bar{A}_{t-1}))|S_0=s_0,A_0=a_0\}=0.
\end{eqnarray}
Let $\bar{s}_j=(s_0,s_1,\cdots,s_j)^\top$, $\bar{y}_j=(y_0,y_1,\cdots,y_j)^\top$, $\bar{S}_j=(S_0,S_1,\cdots,S_j)^\top$ and $\bar{Y}_j=(Y_0,Y_1,\cdots,Y_j)^\top$. We can recursively define the sets $\mathcal{Y}_j(\bar{s}_j,\bar{a}_j,\bar{y}_{j-1})$, $\mathbb{S}_{j+1}(\bar{s}_j,\bar{a}_j,\bar{y}_{j})$, $\mathcal{A}_{j+1}(\bar{s}_{j+1},\bar{a}_j,\bar{y}_j)$ to be the supports of $Y_j,S_{j+1},A_{j+1}$ conditional on $(\bar{S}_j=\bar{s}_j,\bar{A}_j=\bar{a}_j,\bar{Y}_{j-1}=\bar{y}_{j-1})$, $(\bar{S}_j=\bar{s}_j,\bar{A}_j=\bar{a}_j,\bar{Y}_{j}=\bar{y}_{j})$, $(\bar{S}_{j+1}=\bar{s}_{j+1}, \bar{A}_j=\bar{a}_j, \bar{Y}_j=\bar{y}_j)$ respectively, for $j\ge 0$. Similar to \eqref{prooflemma2eq1}, it suffices to show
\begin{eqnarray*}
\Mean \{Q(a';A_t,S_t^*(\bar{A}_{t-1}))-Y_t^*(\bar{A}_t)-\gamma Q(a';a',S_{t+1}^*(\bar{A}_t))\varphi(A_t,S_t^*(\bar{A}_{t-1}))|\bar{S}_t=\bar{s}_t,\bar{A}_t=\bar{a}_t,\bar{Y}_{t-1}=\bar{y}_{t-1}\}=0,
\end{eqnarray*}
for any $s_0\in \mathbb{S}_0,a_0\in \mathcal{A}_0(s_0),y_0\in \mathcal{Y}_0(s_0,a_0),\cdots,s_t\in \mathbb{S}_{t}(\bar{s}_{t-1}, \bar{a}_{t-1}, \bar{y}_{t-1}), a_t\in \mathcal{A}_t(\bar{s}_t,\bar{a}_{t-1},\bar{y}_{t-1})$. This is equivalent to show
\begin{eqnarray}\label{prooflemma2eq2}
\Mean \{Q(a';a_t,S_t^*(\bar{a}_{t-1}))-Y_t^*(\bar{a}_t)-\gamma Q(a';a',S_{t+1}^*(\bar{a}_t))|\bar{S}_t=\bar{s}_t,\bar{A}_t=\bar{a}_t,\bar{Y}_{t-1}=\bar{y}_{t-1}\}=0.
\end{eqnarray}
By construction, we have $\prob(A_t=a_t|\bar{S}_t=\bar{s}_t,\bar{Y}_{t-1}=\bar{y}_{t-1},\bar{A}_{t-1}=\bar{a}_{t-1})>0$. Under SRA, the left-hand-side (LHS) of \eqref{prooflemma2eq2} equals
\begin{eqnarray}\label{prooflemma2eq3}
\Mean \{Q(a';a_t,S_t^*(\bar{a}_{t-1}))-Y_t^*(\bar{a}_t)-\gamma Q(a';a',S_{t+1}^*(\bar{a}_t))|\bar{S}_t=\bar{s}_t,\bar{A}_{t-1}=\bar{a}_{t-1},\bar{Y}_{t-1}=\bar{y}_{t-1}\}.
\end{eqnarray}
Notice that the conditioning event is the same as $\{S_t^*(\bar{a}_{t-1})=s_t, Y_{t-1}^*(\bar{a}_{t-1})=y_{t-1}, \bar{S}_{t-1}=\bar{s}_{t-1},   \bar{A}_{t-1}=\bar{a}_{t-1},\bar{Y}_{t-2}=\bar{y}_{t-2} \}$. Under SRA, \eqref{prooflemma2eq3} equals
\begin{eqnarray*}
\Mean \{Q(a';a_t,S_t^*(\bar{a}_{t-1}))-Y_t^*(\bar{a}_t)-\gamma Q(a';a',S_{t+1}^*(\bar{a}_t))|S_t^*(\bar{a}_{t-1})=s_t, Y_{t-1}^*(\bar{a}_{t-1})=y_{t-1},\\\bar{S}_{t-1}=\bar{s}_{t-1},\bar{A}_{t-2}=\bar{a}_{t-2},\bar{Y}_{t-2}=\bar{y}_{t-2}\}.
\end{eqnarray*}
By recuisvely applying SRA, we can show the left-hand-side (LHS) of \eqref{prooflemma2eq2} equals
\begin{eqnarray*}
\Mean [Q(a';a_t,S_t^*(\bar{a}_{t-1}))-Y_t^*(\bar{a}_t)-\gamma Q(a';a',S_{t+1}^*(\bar{a}_t))| \{S_{j}^*(\bar{a}_{j-1})=s_j\}_{1\le j\le t}, \{Y_{j}^*(\bar{a}_{j})=y_j\}_{1\le j\le t-1}].
\end{eqnarray*}
This is equal to zero by MA, CMIA and Lemma \ref{lemma1}. The proof is hence completed.

\subsection{Proof of Theorem \ref{thm1}}
\subsubsection{Proof under D1}
We begin by providing an outline of the proof. The key to our proof is to show
\begin{eqnarray}\label{proofthm1step0}
\int_{s\in \mathbb{S}} \{\widehat{Q}_t(a;a,s)-Q(a;a,s)\} \mathbb{G}(ds)=\frac{1}{(1-\gamma) t} \sum_{j=0}^{t-1} \omega_t(a;A_j,S_j)\varepsilon_{j,a}+o_p(t^{-1/2}),
\end{eqnarray}
for any $T_1\le t\le T_k$, where $\widehat{Q}_t$ denotes the estimated Q-function at time $t$, $\varepsilon_{j,a}=Y_j+\gamma Q(a;a,S_{j+1})-Q(a;A_j,S_j)$ is the temporal difference error. It follows from \eqref{proofthm1step0} that the proposed value estimator achieves the efficiency limit for policy evaluation \citep{kallus2019efficiently}. In addition, we have
\begin{eqnarray}\label{proofthm1step0.5}
\sqrt{t}\{\widehat{\tau}(t)-\tau_0\}=\frac{1}{(1-\gamma)t}\sum_{j=0}^{t-1}\{\omega_t(1;A_j,S_j)\varepsilon_{j,1}-\omega_t(0;A_j,S_j)\varepsilon_{j,0} \}+o_p(t^{-1/2}). 
\end{eqnarray}
We show \eqref{proofthm1step0} holds in Part 1 of our proof below. 

In Part 2, we show the first term on the right-hand-side (RHS) of \eqref{proofthm1step0.5} is equal to
\begin{eqnarray*}
\frac{1}{t}\sum_{j=0}^{t-1} \int_{s\in \mathbb{S}} \Psi^\top(s) \{A_j \bm{\Sigma}_1^{-1}(t) \Psi(S_j)\varepsilon_{j,1}-(1-A_j) \bm{\Sigma}_0^{-1}(t) \Psi(S_j) \varepsilon_{j,0}\}\mathbb{G}(ds)+o_p(t^{-1/2}),
\end{eqnarray*}
where $\bm{\Sigma}_a(t)=t^{-1}\sum_{j=0}^{t-1}\Mean \Psi(S_t) \mathbb{I}(A_t=a) \{\Psi(S_t)-\gamma \Psi(S_{t+1})\}$. 

In Part 3, we show that
\begin{eqnarray*}
\frac{\sqrt{t}\{\widehat{\tau}(t)-\tau_0\} }{\widehat{\sigma}(t)}\stackrel{d}{\to} N(0,1).
\end{eqnarray*}
Finally, in the last part, we prove Theorem \ref{thm1}. 

\textbf{Part 1: }A key observation is that, the proposed Q-estimator can be represented as the minimizer to the following least square loss over the sieve space,
\begin{eqnarray*}
\argmin_{\bar{Q}} \frac{1}{2t}\sum_{j=0}^{t-1}\{Y_j+\gamma \widehat{Q}_t(a;a,S_{j+1})-\bar{Q}(a;a,S_j)\}^2 \mathbb{I}(A_j=a).
\end{eqnarray*}

Let $Q^*(\widehat{Q}_t,\epsilon_t)=(1-\epsilon_t) \widehat{Q}_t+\epsilon_t \mu^*+\epsilon_t Q$ for a sequence of positive constants $\{\epsilon_t\}_t$ that satisfies $\epsilon_t=o(t^{-1/2})$. We will set $\mu^*$ to be either $\omega^*_t$ or $-\omega^*_t$. For any $\bar{Q}$, let $P_q \bar{Q}$ denote the projection of $\bar{Q}$ to the sieve space such that $\|P_q\bar{Q}-\bar{Q}\|_{\infty}$ achieves the smallest value. Under Condition C2(i), we have
\begin{eqnarray}\label{eqn:app}
\|P_q \mu^*-\mu^*\|_{\infty}=o(T^{-1/4})\,\,\hbox{and}\,\,\|P_q Q-Q\|_{\infty}=o(T^{-1/4}),
\end{eqnarray} 
for any $T_1\le t<T_K$. 
It follows that
\begin{eqnarray*}
P_q Q^*(\widehat{Q}_t,\epsilon_t)=(1-\epsilon_t) P_q \widehat{Q}_t+\epsilon_t P_q \mu^*+\epsilon_t P_q Q=(1-\epsilon_t) \widehat{Q}_t+\epsilon_t P_q \mu^*+\epsilon_t P_q Q,
\end{eqnarray*}
since $ \widehat{Q}_t$ belongs to the sieve space. By definition, we have
\begin{eqnarray*}
&&\frac{1}{2t}\sum_{j=0}^{t-1} \{Y_j+\gamma \widehat{Q}_t(a;a,S_{j+1})-\widehat{Q}_t(a;a,S_j)\}^2\mathbb{I}(A_j=a)\\ &\le& \frac{1}{2t}\sum_{j=0}^{t-1} \{Y_j+\gamma \widehat{Q}_t(a;a,S_{j+1})-P_q Q^*(\widehat{Q}_t,\epsilon_t)(a;a,S_j)\}^2\mathbb{I}(A_j=a).
\end{eqnarray*}
With some calculations, we can show that
\begin{eqnarray}\label{eqn:some3}
\begin{split}
	0\ge \frac{1}{t}\sum_{j=0}^{t-1} \{Y_j+\gamma \widehat{Q}_t(a;a,S_{j+1})-Q(a;a,S_j)\}\{P_q(\mu^*+Q)(a;a,S_j)-\widehat{Q}_t(a;a,S_j)\}\mathbb{I}(A_j=a)\\
	+\frac{1}{t}\sum_{j=0}^{t-1} \{Q(a;a,S_j)-P_q Q^*(\widehat{Q}_t,\epsilon_t/2)(a;a,S_j)\} \{P_q(\mu^*+Q)(a;a,S_j)-\widehat{Q}_t(a;a,S_j)\}\mathbb{I}(A_j=a).
\end{split}	
\end{eqnarray}
We next show that the first term on the RHS is equal to
\begin{eqnarray*}
\frac{1}{t}\sum_{j=0}^{t-1} \{Y_j+\gamma \widehat{Q}_t(a;a,S_{j+1})-Q(a;a,S_j)\}\{(\mu^*+Q)(a;a,S_j)-\widehat{Q}_t(a;a,S_j)\}\mathbb{I}(A_j=a)+o_p(t^{-1/2}).
\end{eqnarray*}
It suffices to show
\begin{eqnarray}\label{eqn:some1}
\begin{split}
	\frac{1}{t}\sum_{j=0}^{t-1} \{Y_j+\gamma \widehat{Q}_t(a;a,S_{j+1})-Q(a;a,S_j)\}\{P_q(\mu^*+Q)(a;a,S_j)-(\mu^*+Q)(a;a,S_j)\}\mathbb{I}(A_j=a)\\=o_p(t^{-1/2}). 
\end{split}	
\end{eqnarray}
A key observation is that under MA and CMIA, the following quantity,
\begin{eqnarray*}
\frac{1}{t}\sum_{j=0}^{t-1} \{Y_j+\gamma Q(a;a,S_{j+1})-Q(a;a,S_j)\}\{P_q(\mu^*+Q)(a;a,S_j)-(\mu^*+Q)(a;a,S_j)\}\mathbb{I}(A_j=a),
\end{eqnarray*}
forms a martingale difference sequence and is of the order of magnitude $o_p(t^{-1/2})$ by Chebyshev's inequality and \eqref{eqn:app}. It remains to show
\begin{eqnarray}\label{eqn:some2}
\begin{split}
	\frac{1}{t}\sum_{j=0}^{t-1} \{ \widehat{Q}_t(a;a,S_{j+1})-Q(a;a,S_{j+1})\}\{P_q(\mu^*+Q)(a;a,S_j)-(\mu^*+Q)(a;a,S_j)\}\mathbb{I}(A_j=a)\\=o_p(t^{-1/2}). 
\end{split}	
\end{eqnarray}
The LHS of \eqref{eqn:some2} can be upper bounded by $t^{-1}|\sum_{j=0}^{t-1}\{\widehat{Q}_t(a;a,S_{j+1})-Q(a;a,S_{j+1})\}| \sup_s |P_q(\mu^*+Q)(a;a,s)-(\mu^*+Q)(a;a,s)|$. By \eqref{eqn:app}, it suffices to show $t^{-1} |\sum_{j=0}^{t-1}\widehat{Q}_t(a;a,S_{j+1})-Q(a;a,S_{j+1})|=o_p(t^{-1/4})$, or 
\begin{eqnarray}\label{eqn:some4}
\frac{1}{t} \left\|\sum_{j=0}^{t-1} \Mean \Psi(S_{j+1})\right\|_2 \|\widehat{\beta}_a(t)-\beta_a^*\|_2=o_p(t^{-1/4}),
\end{eqnarray}
by Condition C2(i) and Cauchy-Schwarz inequality. Here, $\widehat{\beta}_a(t)$ denotes the proposed estimator for $\beta_a^*$ based on the dataset $\{(S_j,A_j,Y_j)\}_{0\le j<t}$. Under C1(i), the probability density functions of $\{S_t\}_t$ are uniformly bounded away from infinity. It follows from \eqref{condother} that $t^{-1} \|\sum_{j=0}^{t-1} \Mean \Psi(S_{j+1})\|_2=O(1)$. It suffices to show $\|\widehat{\beta}_a(t)-\beta_a^*\|_2=o_p(t^{-1/4})$. We introduce the following lemma which states that
\begin{eqnarray}\label{matrixinverse}
\|\bm{\Sigma}^{-1}(t)\|_2=O(1),
\end{eqnarray}
where $\bm{\Sigma}(t)=\Mean \widehat{\bm{\Sigma}}(t)$. 
\begin{lemma}\label{lemmaA1}
Under the given conditions, we have \eqref{matrixinverse} holds.
\end{lemma}
By \eqref{matrixinverse}, using similar arguments in proving Equations (E.21) and (E.22) of \cite{shi2020statistical}, we can show that $\|\widehat{\beta}_a(t)-\beta_a^*\|_2=o_p(t^{-1/4})$ under the given conditions. This yields \eqref{eqn:some4}. 

Similarly, we can show that
\begin{eqnarray*}
\frac{1}{t}\sum_{j=0}^{t-1} \{Y_j+\gamma \widehat{Q}_t(a;a,S_{j+1})-Q(a;a,S_j)\}\{Q(a;a,S_j)-\widehat{Q}_t(a;a,S_j)\}\mathbb{I}(A_j=a)=o_p(t^{-1/2}). 
\end{eqnarray*}
As such, the first term on the RHS of \eqref{eqn:some3} equals
\begin{eqnarray*}
\frac{1}{t}\sum_{j=0}^{t-1} \{Y_j+\gamma \widehat{Q}_t(a;a,S_{j+1})-Q(a;A_j,S_j)\} \mu^*(a;A_j,S_j)\mathbb{I}(A_j=a)+o_p(t^{-1/2}). 
\end{eqnarray*}
Under C1(i), the denominator of $\omega_t$ is uniformly bounded away from zero, and its numerator is uniformly bounded away from infinity for any $t$. As such $\mu^*$ is uniformly bounded. 
Using similar arguments, the second line of \eqref{eqn:some3} equals
\begin{eqnarray*}
\frac{1}{t}\sum_{j=0}^{t-1} \{Q(a;a,S_j)-\widehat{Q}_t(a;a,S_j)\}\mu^*(a;a,S_j)\mathbb{I}(A_j=a)+o_p(t^{-1/2}). 
\end{eqnarray*}
It follows from \eqref{eqn:some3} that
\begin{eqnarray}\label{eqn:some5}
\begin{split}
	0\ge \frac{1}{t}\sum_{j=0}^{t-1} \{Y_j+\gamma \widehat{Q}_t(a;a,S_{j+1})-Q(a;a,S_j)\}\mu^*(a;a,S_j)\mathbb{I}(A_j=a)\\
	+\frac{1}{t}\sum_{j=0}^{t-1} \{Q(a;a,S_j)-\widehat{Q}_t(a;a,S_j)\} \mu^*(a;a,S_j)\mathbb{I}(A_j=a)+o_p(t^{-1/2}).
\end{split}	
\end{eqnarray}
We next show
\begin{eqnarray*}
\frac{1}{t}\sum_{j=0}^{t-1} \{Q(a;a,S_j)-\widehat{Q}_t(a;a,S_j)\} \mu^*(a;a,S_j)\{\mathbb{I}(A_j=a)-b(a|S_j)\}=o_p(t^{-1/2}).
\end{eqnarray*}
It suffices to show
\begin{eqnarray}\label{eqn:some6}
\frac{1}{t}\sum_{j=0}^{t-1} \{Q(a;a,S_j)-\Psi^\top(S_j) \beta_a^*\}\mu^*(a;a,S_j)\{\mathbb{I}(A_j=a)-b(a|S_j)\}=o_p(t^{-1/2}),\\ \label{eqn:some7}
\frac{1}{t}\sum_{j=0}^{t-1} \{\Psi^\top(S_j) \beta_a^*-\Psi^\top(S_j)\widehat{\beta}_a(t)\}\mu^*(a;a,S_j)\{\mathbb{I}(A_j=a)-b(a|S_j)\}=o_p(t^{-1/2}).
\end{eqnarray}
Since the rewards are uniformly bounded, so is the Q-function. Under C2(i), $|\Psi^\top(s) \beta_a^*|$ is uniformly bounded for any $a$ and $s$. The LHS of \eqref{eqn:some6} forms a sum of martingale difference sequence under D1. Since the difference $Q(a;a,s)-\Psi^\top(s) \beta_a^*$ converges to zero uniformly for any $a$ and $s$, it follows from the Chebyshev's inequality that \eqref{eqn:some6} holds. Similarly, we can show that equal element of 
\begin{eqnarray*}
\frac{1}{t}\sum_{j=0}^{t-1} \Psi^\top(S_j)\mu^*(a;a,S_j)\{\mathbb{I}(A_j=a)-b(a|S_j)\}
\end{eqnarray*}
is $O_p(t^{-1/2})$. This together with $\|\beta_a^*-\widehat{\beta}_a(t)\|_2=O_p(t^{1/4})$, Condition C2(ii) and that $q=o(t^{1/2})$ yields \eqref{eqn:some7}. 

Consequently, the second term on the RHS of \eqref{eqn:some5} equals
\begin{eqnarray*}
\frac{1}{t}\sum_{j=0}^{t-1} \{Q(a;a,S_j)-\widehat{Q}_t(a;a,S_j)\} \mu^*(a;a,S_j)b(a|S_j)+o_p(t^{-1/2}).
\end{eqnarray*}
The strict positivity of the transition density function $p$ implies that the limiting density function of the state vector is bounded away from zero. This together with the condition on the initial state distribution in the statement of Theorem \ref{thm1} and the geometric ergodicity assumption in C1(ii) implies that
\begin{eqnarray*}
\int_{s\in \mathbb{S}} M(s)\Pi(ds)+\int_{s\in \mathbb{S}} M(s)F_{S_0}(ds)<+\infty,
\end{eqnarray*}
where $F_{S_0}$ denotes the distribution function of the initial state vector. It follows from Lemma 1 of \cite{meitz2019subgeometric} that the Markov chain $\{S_j\}_{j\ge 0}$ is exponentially $\beta$-mixing. Since the difference $Q(a;a,s)-\Psi^\top(s) \beta_a^*$ converges to zero uniformly for any $a$ and $s$, it guarantees that the zero-mean sum
\begin{eqnarray*}
\frac{1}{t}\sum_{j=0}^{t-1} [\{Q(a;a,S_j)-\Psi^\top(S_j) \beta_a^*\} \mu^*(a;a,S_j)b(a|S_j)
- \Mean \{Q(a;a,S_j)-\Psi^\top(S_j) \beta_a^*\} \mu^*(a;a,S_j)b(a|S_j)], 
\end{eqnarray*}
is of the order $o_p(t^{1/2})$. See e.g., the proof of Lemma E.2 of \cite{shi2020statistical}. Similarly, we can show that
\begin{eqnarray*}
\frac{1}{t}\sum_{j=0}^{t-1} \{\Psi^\top(S_j) \mu^*(a;a,S_j)b(a|S_j)
- \Mean \Psi^\top(S_j) \mu^*(a;a,S_j)b(a|S_j)\} \{\widehat{\beta}_a(t)-\beta_a^*\} =o_p(t^{-1/2}). 
\end{eqnarray*}
It follows that
\begin{eqnarray}\label{eqn:some8}
\begin{split}
	&&\frac{1}{t}\sum_{j=0}^{t-1} \{Q(a;a,S_j)-\widehat{Q}_t(a;a,S_j)\} \mu^*(a;a,S_j)b(a|S_j)\\
	&=&\frac{1}{t}\sum_{j=0}^{t-1} [\Mean Q(a;a,S_j) \mu^*(a;a,S_j)b(a|S_j)-\{\Mean \Psi^\top(S_j)\mu^*(a;a,S_j)b(a|S_j)\}^\top \widehat{\beta}_a(t)].
\end{split}	
\end{eqnarray}
This together with \eqref{eqn:some5}-\eqref{eqn:some7} yields that
\begin{eqnarray*}
\begin{split}
	0\ge \frac{1}{t}\sum_{j=0}^{t-1} \{Y_j+\gamma \widehat{Q}_t(a;a,S_{j+1})-Q(a;a,S_j)\}\mu^*(a;a,S_j)\mathbb{I}(A_j=a)\\
	+\frac{1}{t}\sum_{j=0}^{t-1} \Mean Q(a;a,S_j) \mu^*(a;a,S_j)\mathbb{I}(A_j=a)-\{\Mean \Psi^\top(S_j)\mu^*(a;a,S_j)\mathbb{I}(A_j=a)\}^\top \widehat{\beta}_a(t)\\+o_p(t^{-1/2}).
\end{split}	
\end{eqnarray*}
By setting $\mu^*$ to be $\omega_t$ and $-\omega_t$, we obtain that
\begin{eqnarray}\label{eqn:some9}
\begin{split}
	\frac{1}{t}\sum_{j=0}^{t-1} \Mean Q(a;a,S_j) \omega_t(a;a,S_j)\mathbb{I}(A_j=a)-\{\Mean \Psi^\top(S_j)\omega_t(a;a,S_j)\mathbb{I}(A_j=a)\}^\top \widehat{\beta}_a(t)\\
	+\frac{1}{t}\sum_{j=0}^{t-1} \{Y_j+\gamma \widehat{Q}_t(a;a,S_{j+1})-Q(a;a,S_j)\}\omega_t(a;a,S_j)\mathbb{I}(A_j=a)=o_p(t^{-1/2}). 
\end{split}	
\end{eqnarray}
Similar to \eqref{eqn:some8}, we can show that
\begin{eqnarray}\label{eqn:some10}
\begin{split}
	\frac{1}{t}\sum_{j=0}^{t-1} \{Q(a;a,S_{j+1})-\widehat{Q}_t(a;a,S_{j+1})\}\omega_t(a;a,S_j)\mathbb{I}(A_j=a)\\
	=\frac{1}{t}\sum_{j=0}^{t-1} [\Mean Q(a;a,S_{j+1})\omega_t(a;a,S_j)\mathbb{I}(A_j=a)-\{\Mean \Psi^\top(S_{j+1})\omega_t(a;a,S_j)\mathbb{I}(A_j=a)\}^\top \widehat{\beta}_a(t)].
\end{split}	
\end{eqnarray}
The RHS can be represented as
\begin{eqnarray*}
\Mean Q(a;a,S')\omega_t(a;a,S)\mathbb{I}(A=a)-\{\Mean \Psi^\top(S')\omega_t(a;a,S)\mathbb{I}(A=a)\}^\top \widehat{\beta}_a(t),
\end{eqnarray*}
where $(S,A,S')$ denotes a triplet uniformly sampled from the set $\{(S_j,A_j,S_{j+1})\}_{0\le j<t}$. By definition of $\omega_t$, we have
\begin{eqnarray}\label{eqn:some10.5}
\begin{split}
	\frac{\mathbb{I}(A'=a)}{b(A'|S')}\Mean \{\omega_t(a;a,S)\mathbb{I}(A=a)|S'\}=\frac{1}{\gamma}\left\{\omega_t(a;A',S')-(1-\gamma)\frac{\mathbb{I}(A'=a)g(S')}{b(A'|S')p_b(S')}\right\}\\+O(T^{-1}),
\end{split}	
\end{eqnarray}
where $g$ denotes the density function of $\mathbb{G}$, $p_b(S')$ denotes the density function of $S'$, and $A'$ denotes the action assigned after observing $S'$. As such, the RHS of \eqref{eqn:some10} equals
\begin{eqnarray*}
\frac{1}{t\gamma }\sum_{j=0}^{t-1} \left[\Mean Q(a;A_j,S_j)\left\{\omega_t(a;A_j,S_j)-(1-\gamma)\frac{\mathbb{I}(A_j=a)g(S_j)}{b(A_j|S_j)p_j(b;S_j)}\right\}\right.\\
\left.-\Mean \left\{\omega_t(a;A_j,S_j)-(1-\gamma)\frac{\mathbb{I}(A_j=a)g(S_j)}{b(A_j|S_j)p_j(b;S_j)}\right\}\Psi^\top(S_{j+1}) \widehat{\beta}_a(t)\right]+O(T^{-1}).
\end{eqnarray*}
This together with \eqref{eqn:some9} and \eqref{eqn:some10} yields \eqref{proofthm1step0}. The proof is hence completed. 

\textbf{Part 2:} It suffices to show
\begin{eqnarray}\label{eqn:some11}
\begin{split}
	\frac{1}{(1-\gamma)t}\sum_{j=0}^{t-1}\omega_t(a;A_j,S_j)\varepsilon_{j,a}=\frac{1}{t}\sum_{j=0}^{t-1}\left\{\int_{s\in \mathbb{S}}\Psi(s)\mathbb{G}(ds)\right\}^\top\mathbb{I}(A_j=a)\bm{\Sigma}_a^{-1}(t)\Psi(S_j)\varepsilon_{j,a}\\+o_p(t^{-1/2}).
\end{split}	
\end{eqnarray}
Under C2(i), the difference between $\omega_t(a;A_j,S_j)\mathbb{I}(A_j=a)$ and $\Psi(S_j)^\top \theta_{a,t}^{*}\mathbb{I}(A_j=a)$ converges to zero uniformly for any $j$. In addition,
\begin{eqnarray*}
\frac{1}{(1-\gamma)t}\sum_{j=0}^{t-1} \{\omega_t(a;a,S_j)\mathbb{I}(A_j=a)-\Psi(S_j)^\top \theta_{a,t}^{*}\mathbb{I}(A_j=a)\}\varepsilon_{j,a},
\end{eqnarray*}
corresponds to a sum of martingale difference sequence. As such, the LHS of \eqref{eqn:some11} equals
\begin{eqnarray}\label{eqn:some12}
\frac{1}{(1-\gamma)t}\sum_{j=0}^{t-1} \Psi(S_j)^\top \theta_{a,t}^{*}\mathbb{I}(A_j=a)\varepsilon_{j,a}+o_p(t^{-1/2}).
\end{eqnarray}
Next, it follows from \eqref{eqn:some10.5} and Condition C2(i) that $\theta_{a,t}^*$ satisfies the following,
\begin{eqnarray*}
\left\|\Mean \left\{\Psi^\top(S)\theta_{a,t}^*\mathbb{I}(A=a)\Psi^{\top}(S')-\frac{1}{\gamma}\Psi^\top(S)\theta_{a,t}^*\mathbb{I}(A=a)\Psi^\top(S)\right\}-\frac{1-\gamma}{\gamma} \int_{\mathbb{S}} \Psi(s)\mathbb{G}(ds)\right\|_2\\=o(t^{-1/4}).
\end{eqnarray*}
In Lemma \ref{lemmaA1}, we have shown that
\begin{eqnarray*}
\|\Mean^{-1} [\Psi(S)\{\Psi(S)-\gamma \Psi(S')\}^{\top}\mathbb{I}(A=a)]\|_2=O(1).
\end{eqnarray*}
It follows from the definition of $\bm{\Sigma}_a$ that
\begin{eqnarray*}
\theta_{a,t}^*=(1-\gamma) \left\{\int_{s\in \mathbb{S}}\Psi(s)\mathbb{G}(ds)\right\}^\top\mathbb{I}(A_j=a)\bm{\Sigma}_a^{-1}(t)+o(t^{-1/4}).
\end{eqnarray*}
Using similar arguments in \eqref{eqn:some12}, we can show that the RHS of \eqref{eqn:some11} equals
\begin{eqnarray*}
\frac{1}{(1-\gamma)t}\sum_{j=0}^{t-1} \Psi(S_j)^\top \theta_{a,t}^{*}\mathbb{I}(A_j=a)\varepsilon_{j,a}+o_p(t^{-1/2}),
\end{eqnarray*}
as well. The proof is hence completed.

\textbf{Part 3: }
Define $\xi_{j,a}=\Psi(S_j)\mathbb{I}(A_j=a)$ and
\begin{eqnarray*}
\bm{\Omega}(t)=\Mean\left\{\frac{1}{t} \sum_{j=0}^{t-1}\left(\begin{array}{ll}
	\xi_{j,0} \varepsilon_{j,0}\\
	\xi_{j,1} \varepsilon_{j,1}
\end{array}\right) \left(\begin{array}{ll}
	\xi_{j,0} \varepsilon_{j,0}\\
	\xi_{j,1} \varepsilon_{j,1}
\end{array}\right)^\top \right\}.
\end{eqnarray*}
Based on the results in Part 2, the asymptotic variance of $\sqrt{t}\{\widehat{\tau}(t)-\tau_0\}$ is given by $\sigma^2(t)=\bm{U}^\top \bm{\Sigma}^{-1}(t) \bm{\Omega}(t) \{\bm{\Sigma}^{-1}(t)\}^\top \bm{U}$. We begin by providing a lower bound for $\sigma^2(t)$. Notice that
\begin{eqnarray}\label{sigmalowerbound}
\sigma^2(t)\ge \lambda_{\min}\{\bm{\Omega}(t)\} \|\bm{U}^\top \bm{\Sigma}^{-1}(t)\|_2^2\ge \lambda_{\min}\{\bm{\Omega}(t)\} \lambda_{\min}[\bm{\Sigma}^{-1}(t) \{\bm{\Sigma}^{-1}(t)\}^{\top}] \|\bm{U}\|_2^2.
\end{eqnarray}
Under C1(iii), we have $\liminf_q \|\bm{U}\|_2^2>0$. In addition, notice that $\bm{\Sigma}^{-1}(t) \{\bm{\Sigma}^{-1}(t)\}^{\top}$ is positive semi-definite. It follows that $\lambda_{\min}[\bm{\Sigma}^{-1}(t) \{\bm{\Sigma}^{-1}(t)\}^{\top}]=1/\lambda_{\max}[\bm{\Sigma}(t) \{\bm{\Sigma}(t)\}]$. Similar to the proof of Lemma \ref{lemmaA1}, we can show $\sup_{t\ge 1}\|\bm{\Sigma}(t)\|_2=O(1)$ under C2(ii) and hence $\sup_{t\ge 1}\lambda_{\max}[\bm{\Sigma}(t) \{\bm{\Sigma}(t)\}]=O(1)$. This further yields 
\begin{eqnarray*}
\inf_{t\ge 1} \lambda_{\min}[\bm{\Sigma}^{-1}(t) \{\bm{\Sigma}^{-1}(t)\}^{-1}]>0. 
\end{eqnarray*}
Suppose $\bm{\Omega}(t)$ satisfies 
\begin{eqnarray}\label{Omegatlowerbound}
\liminf_t \lambda_{\min}\{\bm{\Omega}(t)\}>0.
\end{eqnarray} 
It follows that $\sigma^2(t)$ is bounded away from zero, for sufficiently large $t$. 
It follows that
\begin{eqnarray}\label{lemma2keystep2}
\frac{\sqrt{t}\{\widehat{\tau}(t)-\tau_0\}}{\sigma(t)}=\frac{\sqrt{t}\bm{U}^\top \zeta(t)}{\sigma(t)}+o(1),
\end{eqnarray}
where
\begin{eqnarray*}
\zeta(t)=\frac{1}{t}\sum_{j<t}\bm{\Sigma}^{-1}(t)\left(\begin{array}{ll}
	\xi_{j,0} \varepsilon_{j,0}\\
	\xi_{j,1} \varepsilon_{j,1}
\end{array}\right).
\end{eqnarray*}
%
Similar to the proof of Lemma \ref{lemma2}, we can show for any $j\ge 0$, $a\in \{0,1\}$, 
\begin{eqnarray*}
\Mean (\xi_{j,a} \varepsilon_{j,a}|\{S_i,A_i,Y_i\}_{i<j} )=0.
\end{eqnarray*}
By the definition of $\zeta(t)$, $\sqrt{t}\bm{U}^\top\zeta(t)/ \sigma(t)$ forms a martingle with respect to the filtration $\sigma(\{S_j,A_j,Y_j\}_{j<t})$, i.e. the $\sigma$-algebra generated by $\{S_j,A_j,Y_j\}_{j<t}$.  
By the martingale central limit theorem, we can show $\sqrt{t}\bm{U}^\top \zeta(t)/\sigma(t)\stackrel{d}{\to} N(0,1)$. 

To complete the proof of Part 2, we need to show \eqref{Omegatlowerbound} holds and that $\widehat{\sigma}(t)/\sigma(t)\stackrel{P}{\to} 1$. The assertion $\widehat{\sigma}(t)/\sigma(t)\stackrel{P}{\to} 1$ can be similarly proven using arguments from Step 3 of the proof of Theorem 1, \cite{shi2020}. 
We show \eqref{Omegatlowerbound} holds in the following lemma. 
This completes the proof of this part. 
\begin{lemma}\label{lemmaA2}
Under the given conditions, we have \eqref{Omegatlowerbound} holds. 
\end{lemma}

\textbf{Part 4: }Results in Part 2 yield that
$\sqrt{T_k}\{ \widehat{\tau}(T_k)-\tau_0\}/\sigma(T_k) \stackrel{d}{\to} N(0,1)$ for each $1\le k\le K$.  In addition, for any $K$-dimensional vector $\bm{a}=(a_1,\cdots,a_K)^\top$, it follows from \eqref{lemma2keystep2} that
\begin{eqnarray*}
\sum_{k=1}^K \frac{a_k \sqrt{T_k}\{ \widehat{\tau}(T_k)-\tau_0\}}{\sigma(T_k)}=\sum_{k=1}^K \frac{a_k \sqrt{T_k}\bm{U}^\top \zeta_1(T_k)}{\sigma(T_k)}+o_p(1).
\end{eqnarray*}
The leading term on the RHS can be rewritten as a weighted sum of $\{\xi_{j,0}\varepsilon_{j,0},\xi_{j,1}\varepsilon_{j,1} \}_{0\le j<t}$. Similar to the proof in Part 3, we can show it forms a martingale with respect to the filtration $\sigma(\{S_j,A_j,Y_j\}_{j<t})$. We now derive its asymptotic normality for any $\bm{a}$, using the martingale central limit theorem for triangular arrays. 

By Corollary 2 of \cite{McLeish1974}, we need to verify the following two conditions:

\noindent (a) $\max_{0\le j< t} |\sum_{k=1}^K a_k T_k^{-1/2} \bm{U}^\top \bm{\Sigma}^{-1}(T_k) (\xi_{j,0}^\top \varepsilon_{j,0}, \xi_{j,1}^\top \varepsilon_{j,1})^\top\{\sigma(T_k)\}^{-1}\mathbb{I}(j<T_k)|\stackrel{P}{\to} 0$;\\
(b) $\sum_{j=0}^{T-1} |\sum_{k=1}^K a_k T_k^{-1/2}\bm{U}^\top \bm{\Sigma}^{-1}(T_k) (\xi_{j,0}^\top \varepsilon_{j,0}, \xi_{j,1}^\top \varepsilon_{j,1})^\top\{\sigma(T_k)\}^{-1}\mathbb{I}(j<T_k)|^2$ converges to some constant in probability. 

Since $K$ is fixed, to verify (a), it suffices to show $$\max_{1\le j<t,1\le k\le K} T_k^{-1/2}|\bm{U}^\top \bm{\Sigma}^{-1}(T_k) (\xi_{j,0}^\top \varepsilon_{j,0}, \xi_{j,1}^\top \varepsilon_{j,1})^\top\{\sigma(T_k)\}^{-1}|\stackrel{P}{\to} 0.$$ In Lemma \ref{lemmaA1}, we have shown $\|\bm{\Sigma}^{-1}(t)\|=O(1)$. Under the assumption that the rewards are uniformly bounded, so are $|\varepsilon_{j,a}|$'s as well. Using similar arguments in Part 3 of the proof, we can show that $\sigma(t)/\|\bm{U}\|_2$ is bounded away from zero. Therefore, it suffices to show $T_k^{-1/2} \max_{0\le j<t}\|\xi_{j,a}\|_2\stackrel{P}{\to} 0$. Under Condition C2(ii), we have $\sup_{s} \|\Psi(s)\|_2=O(q^{1/2})$ and hence $\max_{0\le j<t}\|\xi_{j,a}\|_2=O(q^{1/2})$. The assertion thus follows by noting that $T_k/T\to c_k$ for some strictly positive constant $c_k>0$ and that $q=o(T)$.

Using similar arguments in the proof of Lemma E.2 of \cite{shi2020}, we can show 
\begin{eqnarray}\label{step3eq}
\left\|\frac{1}{t} \sum_{j=0}^{t-1} (\xi_{j,0}^\top \varepsilon_{j,0}, \xi_{j,1}^\top \varepsilon_{j,1})^\top (\xi_{j,0}^\top \varepsilon_{j,0}, \xi_{j,1}^\top \varepsilon_{j,1})-\bm{\Omega}(t)\right\|_2\stackrel{P}{\to} 0,
\end{eqnarray}
as $t\to \infty$. This together with the facts $\|\bm{\Sigma}^{-1}(t)\|=O(1)$ and $\sigma(t)/\|\bm{U}\|_2$ is bounded away from zero implies that
\begin{eqnarray*}
\left|\frac{a_{k_1}a_{k_2}}{\sqrt{T_{k_1}T_{k_2}} \sigma^2(T_{k_1}\wedge T_{k_2})}\sum_{j=0}^{T_{k_1} \wedge T_{k_2}} \bm{U}^\top \bm{\Sigma}^{-1}(T_{k_1}) (\xi_{j,0}^\top \varepsilon_{j,0}, \xi_{j,1}^\top \varepsilon_{j,1})^\top (\xi_{j,0}^\top \varepsilon_{j,0}, \xi_{j,1}^\top \varepsilon_{j,1}) \{\bm{\Sigma}^{-1}(T_{k_2})\}^\top \bm{U}\right.\\
\left.-\frac{a_{k_1}a_{k_2} (T_{k_1}\wedge T_{k_2})}{\sqrt{T_{k_1}T_{k_2}}\sigma^2(T_{k_1}\wedge T_{k_2})} \bm{U}^\top \bm{\Sigma}^{-1}(T_{k_1}) \bm{\Omega}(T_{k_1}\wedge T_{k_2}) \{\bm{\Sigma}^{-1}(T_{k_2})\}^\top \bm{U}\right\|_2\\
\le \frac{a_{k_1} a_{k_2}}{\sigma^2(T_{k_1}\wedge T_{k_2})} \|\bm{U}\|_2^2 \max_k \|\bm{\Sigma}^{-1}(T_k)\|_2^2  \left\|\frac{1}{T_{k_1}\wedge T_{k_2}} \sum_{j=0}^{T_{k_1}\wedge T_{k_2}-1} (\xi_{j,0}^\top \varepsilon_{j,0}, \xi_{j,1}^\top \varepsilon_{j,1})^\top (\xi_{j,0}^\top \varepsilon_{j,0}, \xi_{j,1}^\top \varepsilon_{j,1})-\bm{\Omega}(t)\right|\\\stackrel{P}{\to} 0,
\end{eqnarray*}
where $a\wedge b=\min(a,b)$. It follows that
\begin{eqnarray}\label{someinequality1}
\begin{split}
	&\left|\sum_{j=0}^{T-1} \left|\sum_{k=1}^K a_k T_k^{-1/2}\bm{U}^\top \bm{\Sigma}^{-1}(T_k) (\xi_{j,0}^\top \varepsilon_{j,0}, \xi_{j,1}^\top \varepsilon_{j,1})^\top\{\sigma(T_k)\}^{-1}\mathbb{I}(j<T_k)\right|^2\right.\\
	-&\left.\sum_{k_1\neq k_2} \frac{a_{k_1}a_{k_2} (T_{k_1}\wedge T_{k_2})}{\sqrt{T_{k_1}T_{k_2}}\sigma^2(T_{k_1}\wedge T_{k_2})} \bm{U}^\top \bm{\Sigma}^{-1}(T_{k_1}) \bm{\Omega}(T_{k_1}\wedge T_{k_2}) \{\bm{\Sigma}^{-1}(T_{k_2})\}^\top \bm{U}\right|=o_p(1). 
\end{split}
\end{eqnarray}
\begin{lemma}\label{lemmaergodic}
Under the given conditions, we have $\|\bm{\Sigma}(t)-\bm{\Sigma}^*\|_2=O(t^{-1/2})$ and $\|\bm{\Omega}(t)-\bm{\Omega}^*\|_2=O(t^{-1/2})$ for some matrices $\bm{\Sigma}^*$ and $\bm{\Omega}^*$ that are invariant to $t$.
\end{lemma}
Combining Lemma \ref{lemmaergodic} with \eqref{matrixinverse}, 
we can show that $\|\bm{\Sigma}^{-1}(t)-\bm{\Sigma}^{*-1}\|_2=O(t^{-1/2})$. 
It follows from \eqref{condother} that $\|\bm{U}\|_2=O(q^{1/2})$. This together with the condition $q=o(\sqrt{T}/\log T)$ yields that $\sigma^2(t)\stackrel{P}{\to} \sigma^{*2}$ where $\sigma^{*2}=\bm{U}^\top \bm{\Sigma}^{*-1} \bm{\Omega}^* (\bm{\Sigma}^{* -1})^\top \bm{U}$. Similar to \eqref{someinequality1}, we have
\begin{eqnarray}\label{someinequality2}
\begin{split}
	&\sum_{k_1\neq k_2} \frac{a_{k_1}a_{k_2} (T_{k_1}\wedge T_{k_2})}{\sqrt{T_{k_1}T_{k_2}}\sigma^2(T_{k_1}\wedge T_{k_2})} \bm{U}^\top \bm{\Sigma}^{-1}(T_{k_1}) \bm{\Omega}(T_{k_1}\wedge T_{k_2}) \{\bm{\Sigma}^{-1}(T_{k_2})\}^\top \bm{U}\\
	\stackrel{P}{\to}&\sum_{k_1\neq k_2} \frac{a_{k_1}a_{k_2} (T_{k_1}\wedge T_{k_2})}{\sqrt{T_{k_1}T_{k_2}}(\sigma^{(0)*})^2} \bm{U}^\top \bm{\Sigma}^{*-1} \bm{\Omega}^{*} (\bm{\Sigma}^{*-1})^\top \bm{U}\to \sum_{k_1\neq k_2} \frac{a_{k_1} a_{k_2} (c_{k_1}\wedge c_{k_2})}{\sqrt{c_{k_1} c_{k_2}}},
\end{split}
\end{eqnarray}
where $c_k$'s are defined in Section \ref{asytest}. 
This together with \eqref{someinequality1} yields that
\begin{eqnarray*}
\sum_{j=0}^{T-1} \left|\sum_{k=1}^K a_k T_k^{-1/2}\bm{U}^\top \bm{\Sigma}^{-1}(T_k) (\xi_{j,0}^\top \varepsilon_{j,0}, \xi_{j,1}^\top \varepsilon_{j,1})^\top\{\sigma(T_k)\}^{-1}\mathbb{I}(j<T_k)\right|^2\stackrel{P}{\to} \frac{a_{k_1} a_{k_2} (c_{k_1}\wedge c_{k_2})}{\sqrt{c_{k_1} c_{k_2}}}.
\end{eqnarray*}
Conditions (a) and (b) are thus verified. Using similar arguments in Step 3 of the proof of Theorem 1 in \cite{shi2020statistical}, we can show
\begin{eqnarray*}
\sum_{k=1}^K \frac{a_k \sqrt{T_k}\{ \widehat{\tau}(T_k)-\tau_0\}}{\sigma(T_k)}=\sum_{k=1}^K \frac{a_k \sqrt{T_k}\{ \widehat{\tau}(T_k)-\tau_0\}}{\widehat{\sigma}(T_k)}+o_p(1),
\end{eqnarray*}
for any $(a_1,\cdots,a_K)$. This yields the joint asymptotic normality of our test statistics. 

By \eqref{someinequality2}, its covariance matrix is given by $\bm{\Xi}_0$ whose $(k_1,k_2)$-th entry is equal to $(c_{k_1}c_{k_2})^{-1/2} c_{k_1} \wedge c_{k_2}$. Similarly, we can show $\widehat{\bm{\Xi}}$ is a consistent estimator for $\bm{\Xi}_0$. This completes the proof of Theorem \ref{thm1} under D1.


\subsubsection{Proof under D2}
The proof is very similar to that under D1. Using similar arguments in the proof of Lemma \ref{lemmaA1}, we can show that under D2, $\|\{\bm{\Sigma}(t)\}^{-1}\|_2=O(1)$,  $\|\bm{\Sigma}(t)-\bm{\Sigma}^*\|_2=O(t^{-1/2})$ and  $\|\bm{\Omega}(t)-\bm{\Omega}^*\|_2=O(t^{-1/2})$ for some time-invariant matrices $\bm{\Sigma}^*$ and $\bm{\Omega}^*$ with $\|(\bm{\Sigma}^*)^{-1}\|_2=O(1)$. 

Notice that the marginalized density ratio $\omega_t$ is well-defined under D2 for any $T_1\le t\le T_K$. 
Using similar arguments in the proof under D1, we can show
\begin{eqnarray*}
\frac{\sqrt{t} \{\widehat{\tau}(t)-\tau_0\} }{\sigma(t)}=\frac{\sqrt{t}\bm{U}^\top \zeta(t)}{\sigma(t)}+o_p(1). 
\end{eqnarray*}
It follows that for any $K$-dimensional vector $\bm{a}=(a_1,\cdots,a_K)^\top$,
\begin{eqnarray*}
\sum_{k=1}^K \frac{a_k\sqrt{T_k} \{\widehat{\tau}(T_k)-\tau_0\} }{\sigma(T_k)}=\sum_{k=1}^K \frac{a_k\sqrt{T_k}\bm{U}^\top \zeta_1(T_k)}{\sigma(T_k)}+o_p(1). 
\end{eqnarray*}
%
Finally, using similar arguments in the proof of Lemma \ref{lemmaA2}, we can show \eqref{step3eq} holds under D2 as well. Now, the joint asymptotic normality of our test statistics follow using arguments from Part 3 of the proof under D1. Similarly, we can show $\widehat{\bm{\Xi}}$ is consistent. This completes the proof under D2. 

\subsubsection{Proof under D3}
The proof under D1 implies that
\begin{eqnarray}\label{firststage}
\frac{\sqrt{T_1}\{\widehat{\tau}(T_1)-\tau_0\}}{\sigma(T_1)}=\frac{\sqrt{T_1}\bm{U}^\top \zeta_1(T_1)}{\sigma(T_1)}+o_p(1). 
\end{eqnarray}
The rest of the proof is divided into two parts. In the first part, we show for $k=2,\cdots,K$,
\begin{eqnarray}\label{laterstage}
\frac{\sqrt{T_k}\{\widehat{\tau}(T_k)-\tau_0\}}{\sigma^*(T_k)}=\frac{\sqrt{T_k}\bm{U}^\top \zeta_1^*(T_k)}{\sigma^*(T_k)}+o_p(1),
\end{eqnarray}
for some $\zeta_1^*(T_k)$ and $\sigma^*(T_k)$ defined below. In the second part, we show the assertion in Theorem \ref{thm1} holds under D3. 

\textbf{Part 1: }For any $1\le k\le K$, consider the matrices 
\begin{eqnarray*}
\bm{\Sigma}^{(k)}=\frac{1}{T_k-T_{k-1}} \sum_{j=T_{k-1}}^{T_k-1} \Mean [\bm{\Sigma}_j|\{(S_t,A_t,Y_t)\}_{0\le t<T_{k-1}}]\,\,\,\,\hbox{and}\,\,\,\,\widehat{\bm{\Sigma}}^{(k)}=\frac{1}{T_k-T_{k-1}} \sum_{j=T_{k-1}}^{T_k-1} \bm{\Sigma}_j.
\end{eqnarray*}
We show in Lemma \ref{lemmaA4} below that for $k=2,\cdots,K$,
\begin{eqnarray}\label{proofD3eq1}
\|\bm{\Sigma}^{(k)}-\widehat{\bm{\Sigma}}^{(k)}\|_2=o_p(q^{-1/2}),
\end{eqnarray}
and
\begin{eqnarray}\label{proofD3eq2}
\|\{\overline{\bm{\Sigma}}^{(k)}\}^{-1}\|_2=O_p(1).
\end{eqnarray}
where $\overline{\bm{\Sigma}}^{(k)}=T_k^{-1} \sum_{i=1}^k (T_i-T_{i-1}) \bm{\Sigma}^{(i)}$. 
\begin{lemma}\label{lemmaA4}
Under the given conditions, we have \eqref{proofD3eq1} and \eqref{proofD3eq2} hold. 
\end{lemma}

Based on these results, using similar arguments in the proof of Theorem 2 of \cite{shi2020statistical}, we can show that $\|\widehat{\beta}_a(t)-\beta_a^*\|_2$ converges at a rate of $o_p(t^{-1/4})$. Next,  using similar arguments in Part 1 of the proof under D1, we can show that
\begin{eqnarray}\label{proofD3eq4}
\sqrt{T_k}\{\widehat{\tau}(T_k)-\tau_0\}=\sqrt{T_k}\bm{U}^\top \zeta_1^*(T_k)+o_p(1),\,\,\,\,\forall k\in \{2,\cdots,K\},
\end{eqnarray} 
where
\begin{eqnarray*}
\zeta_1^*(T_k)=\frac{1}{T_k} \sum_{j=1}^{T_k}(\overline{\bm{\Sigma}}^{(k)})^{-1} \left(\begin{array}{c}
	\xi_{j,0}\varepsilon_{j,0}\\
	\xi_{j,1}\varepsilon_{j,1}
\end{array}
\right).
\end{eqnarray*}
For $1\le k\le K$, define
\begin{eqnarray*}
\bm{\Omega}^{(k)}=\frac{1}{T_k-T_{k-1}}\sum_{j=T_{k-1}}^{T_k-1} \Mean [(\xi_{j,0}^\top \varepsilon_{j,0}, \xi_j^\top\varepsilon_{j,1})^\top (\xi_{j,0}^\top \varepsilon_{j,0}, \xi_j^\top\varepsilon_{j,1})|\{(S_t,A_t,Y_t)\}_{0\le t<T_{k-1}}],
\end{eqnarray*}
and $\overline{\bm{\Omega}}^{(k)}=T_k^{-1} \sum_{i=1}^k (T_i-T_{i-1}) \bm{\Sigma}^{(i)}$. For any $2\le k\le K$, we have $\lambda_{\min}(\overline{\bm{\Omega}}^{(k)})\ge \lambda_{\min}(T_k^{-1} T_1 \bm{\Omega}^{(1)})$. Since $T_k^{-1} T_1\to c_k^{-1} c_1>0$ and $ \lambda_{\min}(\bm{\Omega}^{(1)})=\lambda_{\min}(\bm{\Omega}(T_1))$ is bounded away from zero, $\lambda_{\min}(\overline{\bm{\Omega}}^{(k)})$ is bounded away from zero for $k=2,\cdots,K$ as well. Define
\begin{eqnarray*}
\{\sigma^*(T_k)\}^2=\bm{U}^\top (\overline{\bm{\Sigma}}^{(k)})^{-1} \overline{\bm{\Omega}}^{(k)} \{(\overline{\bm{\Sigma}}^{(k)})^{-1}\}^\top \bm{U}.
\end{eqnarray*}
It can be shown that $\sigma^*(T_k)/\|\bm{U}\|_2$ is bounded away from zero, for $k=2,\cdots,K$. Using similar arguments in Part 2 of the proof under D1, 
we can show \eqref{laterstage} holds. This completes the proof for Part 1. 

\textbf{Part 2: }Let $\sigma^*(T_1)=\sigma(T_1)$. By \eqref{firststage} and \eqref{laterstage}, we have for any $K$-dimensional vector $\bm{a}=(a_1,\cdots,a_K)^\top$ that
\begin{eqnarray}\label{D3part2eq}
\sum_{k=1}^K \frac{a_k \sqrt{T_k} \{\widehat{\tau}(T_k)-\tau_0\}}{\sigma^*(T_k)}=\sum_{k=1}^K \frac{a_k \sqrt{T_k} \bm{U}^\top \zeta_1(T_k)}{\sigma^*(T_k)}+o_p(1).
\end{eqnarray}
In the following, we show the leading term on the RHS of \eqref{D3part2eq} is asymptotically normal. Similar to the proof under D1, it suffices to verify the following conditions:

\noindent (a) $\max_{0\le j< T} |\sum_{k=1}^K a_k T_k^{-1/2} \bm{U}^\top (\overline{\bm{\Sigma}}^{(k)})^{-1} (\xi_{j,0}^\top \varepsilon_{j,0}, \xi_{j,1}^\top \varepsilon_{j,1})^\top\mathbb{I}(j<T_k)|\stackrel{P}{\to} 0$;\\
(b) $\sum_{j=0}^{T-1} |\sum_{k=1}^K a_k T_k^{-1/2}\bm{U}^\top (\overline{\bm{\Sigma}}^{(k)})^{-1} (\xi_{j,0}^\top \varepsilon_{j,0}, \xi_{j,1}^\top \varepsilon_{j,1})^\top\{\sigma^*(T_k)\}^{-1}\mathbb{I}(j<T_k)|^2$ converges to some constant in probability. 

Condition (a) can be proven in a similar manner as in Part 3 of the proof under D1. 
Notice that for $k=2,\cdots,K$, $\overline{\bm{\Sigma}}^{(k)}$, $\overline{\bm{\Omega}}^{(k)}$ and $\sigma^*(T_k)$ are random variables and depend on the observed data history. In the proof of Lemma \ref{lemmaA4}, we show $\|(\overline{\bm{\Sigma}}^{(k)})^{-1}-(\bm{\bm{\Sigma}^{**}})^{-1}\|_2=O_p(T^{-1/2})$ for some deterministic matrix $\bm{\Sigma}^*$ and all $k\in \{2,\cdots,K\}$. Similarly, we can show $\|\overline{\bm{\Omega}}^{(k)}-\bm{\Omega}^{**}\|_2=O_p(T^{-1/2})$ and $\|\{\sigma^{*}(T_k)\}^2-(\sigma^{**})^2\|_2=O_p(T^{-1/2})$ for some  $\bm{\Sigma}^*$, $\sigma^{**}$ and all $k\in \{2,\cdots,K\}$. Moreover, using similar arguments in the proof of Lemma \ref{lemmaA4}, we can show
\begin{eqnarray*}
\left\|\frac{1}{T_k-T_{k-1}} \sum_{j=T_{k-1}}^{T_k-1}  \left(\begin{array}{cc}
	\xi_{j,0} \varepsilon_{j,0}\\
	\xi_{j,1} \varepsilon_{j,1}
\end{array}\right)\left(\begin{array}{cc}
	\xi_{j,0} \varepsilon_{j,0}\\
	\xi_{j,1} \varepsilon_{j,1}
\end{array}\right)^\top-\bm{\Omega}^{(k)}\right\|_2=o_p(q^{-1/2}),\,\,\,\,\,\,\,\,\forall k=2,\cdots,K.
\end{eqnarray*}
This further implies that
\begin{eqnarray*}
\left\|\frac{1}{T_k} \sum_{j=0}^{T_k-1} \left(\begin{array}{cc}
	\xi_{j,0} \varepsilon_{j,0}\\
	\xi_{j,1} \varepsilon_{j,1}
\end{array}\right)\left(\begin{array}{cc}
	\xi_{j,0} \varepsilon_{j,0}\\
	\xi_{j,1} \varepsilon_{j,1}
\end{array}\right)^\top-\bm{\Omega}^{**} \right\|_2=o_p(q^{-1/2}),\,\,\,\,\,\,\,\,\forall k=2,\cdots,K.
\end{eqnarray*}
Based on these results, using similar arguments in Part 3 of the proof of Lemma \ref{lemmaA1}, we obtain (b). The joint asymptotic normality of $\sqrt{T_1}\{\widehat{\tau}(T_1)-\tau_0\}/\sigma^*(T_1),\cdots,\sqrt{T_1}\{\widehat{\tau}(T_1)-\tau_0\}/\sigma^*(T_K)$ thus follows. 

Consistency of $\widehat{\bm{\Xi}}$ can be similarly proven. We omit the details for brevity. 





\subsection{Proof of Theorem \ref{thm2}}
As discussed in Section \ref{secalphaspend}, $(Z_1^*,Z_2^*,\cdots,Z_K^*)^\top $ is jointly normal with mean zero and covariance matrix $\widehat{\bm{\Xi}}$, conditional on the observed data. By Theorem 1, we have $\widehat{\bm{\Xi}}\stackrel{P}{\to} \bm{\Xi}_0$ where $\bm{\Xi}_0$ is the asymptotic covariance matrix of $(Z_1,Z_2,\cdots,Z_K)^\top$. Let $\alpha^*(t)=\alpha(tT)$ for any $0\le t\le 1$, we have $\alpha(T_k)\to \alpha^*(c_k)$ for any $1\le k\le K$. Notice that $\{\widehat{b}_k\}_{1\le k\le K}$ is a continuous function of $\widehat{\bm{\Xi}}$ and $\{\alpha(T_k)\}_{1\le k\le K}$, it follows that $\widehat{b}_k\stackrel{P}{\to} b_{k,0}$ for $1\le k\le K$, where $\{b_{k,0}\}_{1\le k\le K}$ are recursively defined as follows:
\begin{eqnarray*}
\hbox{Pr}\left\{ \max_{1\le j< k} (Z_{j,0}-b_{j,0})\le 0, Z_{k,0}>b_{k,0}\right\}=\alpha^*(c_k)-\alpha^*(c_{k-1}),
\end{eqnarray*}
where $(Z_{1,0},Z_{2,0},\cdots,Z_{K,0})^\top$ is asymptotically normal with mean zero and covariance matrix $\bm{\Xi}_0$. 

Theorem \ref{thm1} implies that $\{Z_1-\sqrt{T_1} \tau_0/\widehat{\sigma}(T_1),Z_2-\sqrt{T_2}\tau_0/\widehat{\sigma}(T_2),\cdots,Z_K-\sqrt{T_K}\tau_0/\widehat{\sigma}(T_K)\}^\top\stackrel{d}{\to} (Z_{1,0},Z_{2,0},\cdots,Z_{K,0})^\top$. 
It follows that
\begin{eqnarray}\label{limitingdistribution}
\begin{split}
	\prob\left(\bigcup_{j=1}^k\{Z_j> \widehat{b}_j\}\right)\le \prob\left(\bigcup_{j=1}^k\{Z_j-\sqrt{T_j} \tau_0/\widehat{\sigma}(T_j)> \widehat{b}_j\}\right)\\\to \prob\left(\bigcup_{j=1}^k \{Z_{j,0}> b_{j,0}\}\right)=\alpha^*(c_k). 
\end{split}
\end{eqnarray}
The proof is hence completed by noting that $\alpha(T_k)\to \alpha^*(c_k)$. When $\tau_0=0$, the first inequality in \eqref{limitingdistribution} becomes an equality. The rejection probability thus converges to the nominal level. 

\subsection{Proof of Theorem \ref{thm3}}
Suppose $\tau_0=T^{-1/2} h$ for some $h>0$. Based on the proof of Theorem \ref{thm1}, we can show $\widehat{\sigma}(T_k)\stackrel{P}{\to} \sigma_k^*$ for some $\sigma_k^*>0$. It follows from \eqref{limitingdistribution} that
\begin{eqnarray*}
\prob\left(\bigcup_{j=1}^k\{Z_j> \widehat{b}_j\}\right)= \prob\left(\bigcup_{j=1}^k\{Z_j-\sqrt{T_j} \tau_0/\widehat{\sigma}(T_j)> \widehat{b}_j-h/\widehat{\sigma}(T_j)\}\right)\\\to \prob\left(\bigcup_{j=1}^k \{Z_{j,0}> b_{j,0}-h/\sigma_j^*\}\right)>\alpha^*(c_k). 
\end{eqnarray*} 
The second assertion in Theorem \ref{thm3} thus holds by noting that $\alpha(T_k)\to \alpha^*(c_k)$. 

Let $h\to \infty$, we obtain
\begin{eqnarray*}
\prob\left(\bigcup_{j=1}^k\{Z_j> \widehat{b}_j\}\right)= \prob\left(\bigcup_{j=1}^k \{Z_{j,0}> b_{j,0}-h/\sigma_j^*\}\right)+o(1)\to 1. 
\end{eqnarray*} 
The proof is hence completed.

\subsection{Proof of Lemma \ref{lemmaA1}}\label{seclemmaA1}
Notice that the matrix $\bm{\Sigma}(t)$ can be rewritten as
\begin{eqnarray*}
\bm{\Sigma}(t)=\left[\begin{array}{cc}
	\bm{\Sigma}_0(t) & \\
	& \bm{\Sigma}_1(t)
\end{array}
\right].
\end{eqnarray*}
It suffices to show
\begin{eqnarray*}
\|\{\bm{\Sigma}_a(t)\}^{-1}\|_2=O(1),
\end{eqnarray*}
for $a\in \{0,1\}$. 
Using similar arguments in Part 1 of the proof of Lemma E.2, \cite{shi2020}, it suffices to show
\begin{eqnarray*}
\bm{a}^\top \bm{\Sigma}_a(t) \bm{a}\ge \bar{c}_1 \|\bm{a}\|_2^2,\,\,\,\,\,\,\,\,\forall \bm{a},
\end{eqnarray*}
for some $\bar{c}_1>0$ and sufficiently large $t$. By definition, we have
\begin{eqnarray*}
\bm{\Sigma}_a(t)=\frac{1}{t}\sum_{j=0}^t \int_{s,s'} \Psi(s) \{\Psi(s)-\gamma \Psi(s') \}^\top b(a|s)p_j(b;s) p(s'|a,s) dsds'.
\end{eqnarray*}
Under D1, $b$ is strictly positive. Since $p$ is strictly positive, so is $p_j(b;\cdot)$ for any $j\ge 2$. It suffices to show
\begin{eqnarray}\label{prooflemmaA1eq2.1}
\bm{a}^\top \int_{s,s'\in \mathbb{S}} \Psi(s)\{\Psi(s)-\gamma \Psi(s') \}^\top dsds' \bm{a}\ge \bar{c}_2 \|\bm{a}\|_2^2,\,\,\,\,\,\,\,\,\forall \bm{a},
\end{eqnarray}
for some $\bar{c}_2>0$. Notice that LHS of \eqref{prooflemmaA1eq2.1} is smaller than
\begin{eqnarray*}
\lambda(\mathbb{S})\int_{s\in \mathbb{S}} \{\bm{a}^\top \Psi(s)\}^2 ds-\gamma \int_{s,s'\in \mathbb{S}}|\bm{a}^\top \Psi(s)| |\bm{a}^\top \Psi(s')|dsds',
\end{eqnarray*}
where $\lambda(\mathbb{S})$ is the Lebesgue measure of $\mathbb{S}$. Since $\mathbb{S}$ is compact, we have $\lambda(\mathbb{S})<+\infty$. 
By Cauchy-Schwarz inequality, LHS of \eqref{prooflemmaA1eq2.1} is greater than or equal to
\begin{eqnarray*}
\lambda(\mathbb{S})\int_{s\in \mathbb{S}} \{\bm{a}^\top \Psi(s)\}^2 ds-\lambda(\mathbb{S})\int_{s\in \mathbb{S}}\frac{\gamma}{2}\{\bm{a}^\top \Psi(s)\}^2 ds-\lambda(\mathbb{S})\int_{s\in \mathbb{S}}\frac{\gamma}{2}\{\bm{a}^\top \Psi(s')\}ds'
\\\ge (1-\gamma)\lambda(\mathbb{S}) \int_{s\in \mathbb{S}}\{\bm{a}^\top \Psi(s)\}^2 ds.
\end{eqnarray*}
This is directly implied by Condition C2(ii). The proof is hence completed. 

\subsection{Proof of Lemma \ref{lemmaA2}}\label{seclemmaA2}
We focus on \eqref{Omegatlowerbound}. 
Notice that $\bm{\Omega}(t)$ is a block-diagonal matrix formed by
\begin{eqnarray*}
\bm{\Omega}(t)=\left[\begin{array}{cc}
	\bm{\Omega}_0(t) & \\
	& \bm{\Omega}_1(t)
\end{array}
\right],
\end{eqnarray*}
where
\begin{eqnarray*}
\bm{\Omega}_a(t)=\frac{1}{t}\sum_{j<t}\int_{s\in \mathbb{S}} \Mean \{(\xi_{0,a}\xi_{0,a}^\top) \varepsilon_{0,a}^2|S_0=s\} p_j(b;s)ds.
\end{eqnarray*}
It suffices to show $\lambda_{\min}(\bm{\Omega}_{a,t})$ is bounded away from zero for sufficiently large $t$. Under the assumption that $p$ is strictly positive, $\{p_j(b;\cdot)\}_{j\ge 2}$ are uniformly bounded away from zero. It suffices to show 
\begin{eqnarray*}
\lambda_{\min}\left[\int_{s\in \mathbb{S}} \Mean \{(\xi_{0,a}\xi_{0,a}^\top) \varepsilon_{0,a}^2|S_0=s\}ds \right]
\end{eqnarray*}
is bounded away from zero. However, this is directly implied by Conditions C2(ii), C3 and the fact that $b$ is strictly bounded away from zero. The proof is thus completed.

\subsection{Proof of Lemma \ref{lemmaergodic}}
For any random variable $\mathbb{Z}$ that satisfies $\prob(|\mathbb{Z}|\le \mathbb{N})=1$ and any integer $J$, we have
\begin{eqnarray}\label{prooflemma3eq12}
\begin{split}
	&\left|\Mean \mathbb{Z}- \sum_{j=-J}^J \frac{\mathbb{N}j}{J}\hbox{Pr}\left(\frac{\mathbb{N} j}{J}\le \mathbb{Z}< \frac{\mathbb{N}(j+1)}{J}\right)\right|\\ 
	\le& \sum_{j=-J}^{J} \Mean \left|\mathbb{Z}-\frac{\mathbb{N}j}{J}\right|\mathbb{I}\left(\frac{\mathbb{N} j}{J}\le \mathbb{Z}< \frac{\mathbb{N}(j+1)}{J}\right)\le \frac{\mathbb{N}}{J}.
\end{split}
\end{eqnarray}
Under D1, $p_j(b;\cdot)$ will converge to the stationary distribution $\mu(\cdot)$. Define
\begin{eqnarray*}
\bm{\Sigma}_a^*=\int \Psi(s)\{\Psi(s)-\gamma \Psi(s')\}^\top p(s'|a,s)b(a|s)\mu(s)dsds'.
\end{eqnarray*}
For any $q$-dimensional vectors $\nu_1$ and $\nu_2$ with unit $\ell_2$ norm, we define the function 
\begin{eqnarray*}
f(\nu_1,\nu_2,s)=\int \nu_1^\top \Psi(s')\{\Psi(s')-\gamma \Psi(s^{''})\}^\top \nu_2 p(s^{''}|a',s')b(a'|s') \left\{\sum_{a} b(a|s)p(a'|a,s)\right\} ds^{''}ds'.
\end{eqnarray*}
Under Condition (C2)(ii), $f$ is a bounded function of $s$. Notice that
\begin{eqnarray*}
\nu_1^\top \bm{\Sigma}_a(t) \mu_2=\frac{1}{t} \mu_1^\top \Mean \Psi(S_0)\{\Psi(S_0)-\gamma \Psi(S_1)\}^\top \mu_2 \mathbb{I}(A_0=a)+\frac{1}{t}\sum_{j=0}^{t-2} \Mean f(\nu_1,\nu_2,S_j).
\end{eqnarray*}
For any $T_1\le t\le T_K$, the absolute value of the first term on the RHS is of the order $O(q/T)=O(T^{-1/2})$ under C2(ii) and that $q=O(\sqrt{T})$. By \eqref{prooflemma3eq12}, the second term on the RHS can be approximated by
\begin{eqnarray}\label{eqnsomeapproximation}
\frac{1}{t-1}\sum_{k=0}^{t-2}\sum_{j=-J}^J \frac{\mathbb{N}j}{J}\prob\left(\frac{\mathbb{N}j}{J}\le f(\nu_1,\nu_2,S_k)\le \frac{\mathbb{N}(j+1)}{J}\right),
\end{eqnarray} 
with the approximation error bounded by $O(T^{-1}+J^{-1})$. Under the geometric ergodicity assumption in C1(ii), the probability 
\begin{eqnarray*}
\prob\left(\frac{\mathbb{N}j}{J}\le f(\nu_1,\nu_2,S_k)\le \frac{\mathbb{N}(j+1)}{J}\right)
\end{eqnarray*}
can be approximated by $\Mean_{S\sim \mu} \mathbb{I}(J^{-1}\mathbb{N}j\le f(\nu_1,\nu_2,S)\le J^{-1}\mathbb{N}(j+1))$ with the approximation error bounded by $O(\rho^{k})$. As such, \eqref{eqnsomeapproximation} can be approximated by $\Mean_{S\sim \mu} f(\nu_1,\nu_2,S)=\bm{\Sigma}_a^*$ with the approximation error bounded by $O(T^{-1 }J)$. Consequently, we obtain
\begin{eqnarray*}
\sup_{\|\nu_1\|_2=1,\|\nu_2\|_2=1}| \nu_1^\top \{\bm{\Sigma}_a^* -\bm{\Sigma}_a(t)\} \nu_2|=O(T^{-1} J)+O(J^{-1}). 
\end{eqnarray*}
By setting $J=\sqrt{T}$, the RHS becomes $O(T^{-1/2})=O(t^{-1/2})$ for any $T_1\le t\le T_k$. This implies that $\|\bm{\Sigma}_a^* -\bm{\Sigma}_a(t)\|_2=O(t^{-1/2})$. Similarly, we can show $\|\bm{\Sigma}(t)-\bm{\Sigma}^*\|_2=O(t^{-1/2})$ and $\|\bm{\Omega}(t)-\bm{\Omega}^*\|_2=O(t^{-1/2})$ for some $\bm{\Sigma}^*$ and $\bm{\Omega}^*$. The proof is hence completed. 

\subsection{Proof of Lemma \ref{lemmaA4}}
Under C1(iv), we have \eqref{remarkeq1} holds. Similar to \eqref{remarkeq2}, we can show $\Pi^{(k)}$ has a probability density function $\mu^{(k)}$ given by
\begin{eqnarray}\label{muks}
\mu^{(k)}(s')=\sum_{a\in\{0,1\}} \int_{s\in \mathbb{S}} b^{(k)}(a|s) p(s';a,s)\Pi^{(k)}(ds).
\end{eqnarray}
For $a'\in\{0,1\}$, define
\begin{eqnarray*}
\bm{\Sigma}_a^{(k)*}=\int_{s,s'\in \mathbb{S}} \Psi(s)\{\Psi(s)-\gamma \Psi(s')\}^\top \mu^{(k)}(s) b^{(k)}(a|s) p(s';a,s)dsds'.
\end{eqnarray*}
Condition on $\{(S_j,A_j,Y_j)\}_{1\le j< T_{k-1}}$, the matrix $\bm{\Sigma}_a^{(k)*}$ is deterministic. Let $\bm{\Sigma}^{(k)}$ be the block-diagonal matrix $\diag[\bm{\Sigma}_0^{(k)*},\bm{\Sigma}_1^{(k)*}]$ created by aligning $\bm{\Sigma}_0^{(k)*}$ and $\bm{\Sigma}_1^{(k)*}$ along the diagonal of $\bm{\Sigma}^{(k)}$. Similar to the proof of Lemma \ref{lemmaergodic}, we can show $\|\bm{\Sigma}^{(k)*}-\bm{\Sigma}^{(k)}\|_2=o(1)$, conditional on $\{(S_j,A_j,Y_j)\}_{1\le j< T_{k-1}}$, with probability tending to $1$. This implies for any sufficiently small $\epsilon>0$, 
\begin{eqnarray*}
\prob( \|\bm{\Sigma}^{(k)*}-\bm{\Sigma}^{(k)}\|_2>\epsilon|\{(S_j,A_j,Y_j)\}_{1\le j<T_{k-1}} )\stackrel{P}{\to} 0.
\end{eqnarray*}
The above conditional probability is bounded between $0$ and $1$. Using bounded convergence theorem, we have
\begin{eqnarray}\label{lemmaA4eq1}
\prob(\|\bm{\Sigma}^{(k)*}-\bm{\Sigma}^{(k)}\|_2>\epsilon)=o(1),
\end{eqnarray}
and hence $\|\bm{\Sigma}^{(k)*}-\bm{\Sigma}^{(k)}\|_2=o_p(1)$. 

Notice that $\sup_s |b^{(k)}(a|s)-b^*(a|s)|\stackrel{P}{\to} 0$ and $\|\Pi^{(k)}-\Pi^*\|_{\textrm{TV}}\stackrel{P}{\to} 0$. Define
\begin{eqnarray*}
\mu^*(s)=\sum_{a\in\{0,1\}} \int_{s\in \mathbb{S}} b^*(a|s) p(s';a,s)\Pi^{*}(ds).
\end{eqnarray*}
It follows that
\begin{eqnarray*}
|\mu^{(k)}(s')-\mu^*(s')|\le \sum_{a\in\{0,1\}} \int_{s\in \mathbb{S}} |b^{(k)}(a|s)-b^*(a|s)| p(s';a,s)\Pi^{(k)}(ds)\\
+\sum_{a\in\{0,1\}}\int_{s\in \mathbb{S}} b^*(a|s)p(s';a,s) |\Pi^{(k)}(ds)-\Pi^*(ds)|,
\end{eqnarray*}
and hence $\sup_{s} |\mu^{(k)}(s)-\mu^*(s)|\stackrel{P}{\to} 0$. Under Condition C2(ii), we can show $\|\bm{\Sigma}_a^{(k)*}-\bm{\Sigma}_a^{**}\|_2\stackrel{P}{\to} 0$ where
\begin{eqnarray*}
\bm{\Sigma}_a^{**}=\int_{s,s'\in \mathbb{S}} \Psi(s)\{\Psi(s)-\gamma \Psi(s')\}^\top \mu^{*}(s) b^*(a|s) p(s';a,s)dsds'.
\end{eqnarray*}
Let $\bm{\Sigma}^{**}$ be the block-diagonal matrix $\diag[\bm{\Sigma}_0^{**},\bm{\Sigma}_1^{**}]$, we obtain $\|\bm{\Sigma}^{(k)*}-\bm{\Sigma}^{**}\|_2=o_p(1)$ for any $k\ge 2$. Combining this together with \eqref{lemmaA4eq1}, we obtain $\|\bm{\Sigma}^{(k)}-\bm{\Sigma}^{**}\|_2=o_p(1)$. According to the proof under D1, $\|\bm{\Sigma}^{(1)}-\bm{\Sigma}^{*}\|_2=o(1)$. Thus, we have for any $2\le k\le K$ that
\begin{eqnarray}\label{lemmaA4eq2}
\|\overline{\bm{\Sigma}}^{(k)}-T_k^{-1} T_1 \bm{\Sigma}^{*}-T_k^{-1}(T_k-T_1)\bm{\Sigma}^{**}\|_2=o_p(1).
\end{eqnarray}

Similar to the proof of Lemma \ref{lemmaA1}, we can show $\mu^{(k)}$'s are uniformly bounded away from $0$ and $\infty$. It follows that $\mu^*$ is uniformly bounded away from $0$ and $\infty$ as well. Using similar arguments in Lemma \ref{lemmaA1},  we can show $\|\{T_k^{-1} T_1 \bm{\Sigma}^{*}+T_k^{-1} (T_k-T_1)\bm{\Sigma}^{**}\}^{-1}\|_2=O(1)$. Using similar arguments in Part 1 of the proof of Lemma E.2, \cite{shi2020}, we have by \eqref{lemmaA4eq2} that $\|(\overline{\bm{\Sigma}}^{(k)})^{-1}\|_2=O(1)$, with probability tending to $1$.  \eqref{proofD3eq2} is thus proven. 

Assertion \eqref{proofD3eq1} now follows using similar arguments in the proof of Lemma E.2, \cite{shi2020}.

\end{document}